\pdfoutput=1
\documentclass[a4paper,10pt]{article}
\usepackage{geometry}
\usepackage{fullpage}
\usepackage{graphicx}
\usepackage{amsmath}
\usepackage{url}
\usepackage{syntax}
\usepackage{listings}
\usepackage{hyperref}
\usepackage[strict]{changepage}
\usepackage{color}
\usepackage{textcomp}
\usepackage{subfigure}
\usepackage[titles, subfigure]{tocloft}
\usepackage{floatrow}

\hypersetup{
    colorlinks,
    citecolor=black,
    filecolor=black,
    linkcolor=black,
    urlcolor=black
}

\newcommand{\lbda}{\color[rgb]{0,.3,.7} \lambda}

\begin{document}

\title{Inducing Probabilistic Programs by Bayesian Program Merging}
\author{Irvin Hwang\\Stanford University  \and Andreas Stuhlm\"{u}ller\\MIT \and Noah D. Goodman\\Stanford University}
\date{}
\maketitle
\thispagestyle{empty}
\begin{abstract}

This report outlines an approach to learning generative models from data. We express models as probabilistic programs, which allows us to capture abstract patterns within the examples. By choosing our language for programs to be an extension of the algebraic data type of the examples, we can begin with a program that generates all and only the examples. We then introduce greater abstraction, and hence generalization, incrementally to the extent that it improves the posterior probability of the examples given the program. Motivated by previous approaches to model merging and program induction, we search for such explanatory abstractions using program transformations. We consider two types of transformation: {\em Abstraction} merges common subexpressions within a program into new functions (a form of anti-unification). {\em Deargumentation} simplifies functions by reducing the number of arguments. We demonstrate that this approach finds key patterns in the domain of nested lists, including parameterized sub-functions and stochastic recursion.

\end{abstract}

\tableofcontents

\section{Introduction}

\begin{figure}[b]
\begin{center}
\includegraphics[scale=.26]{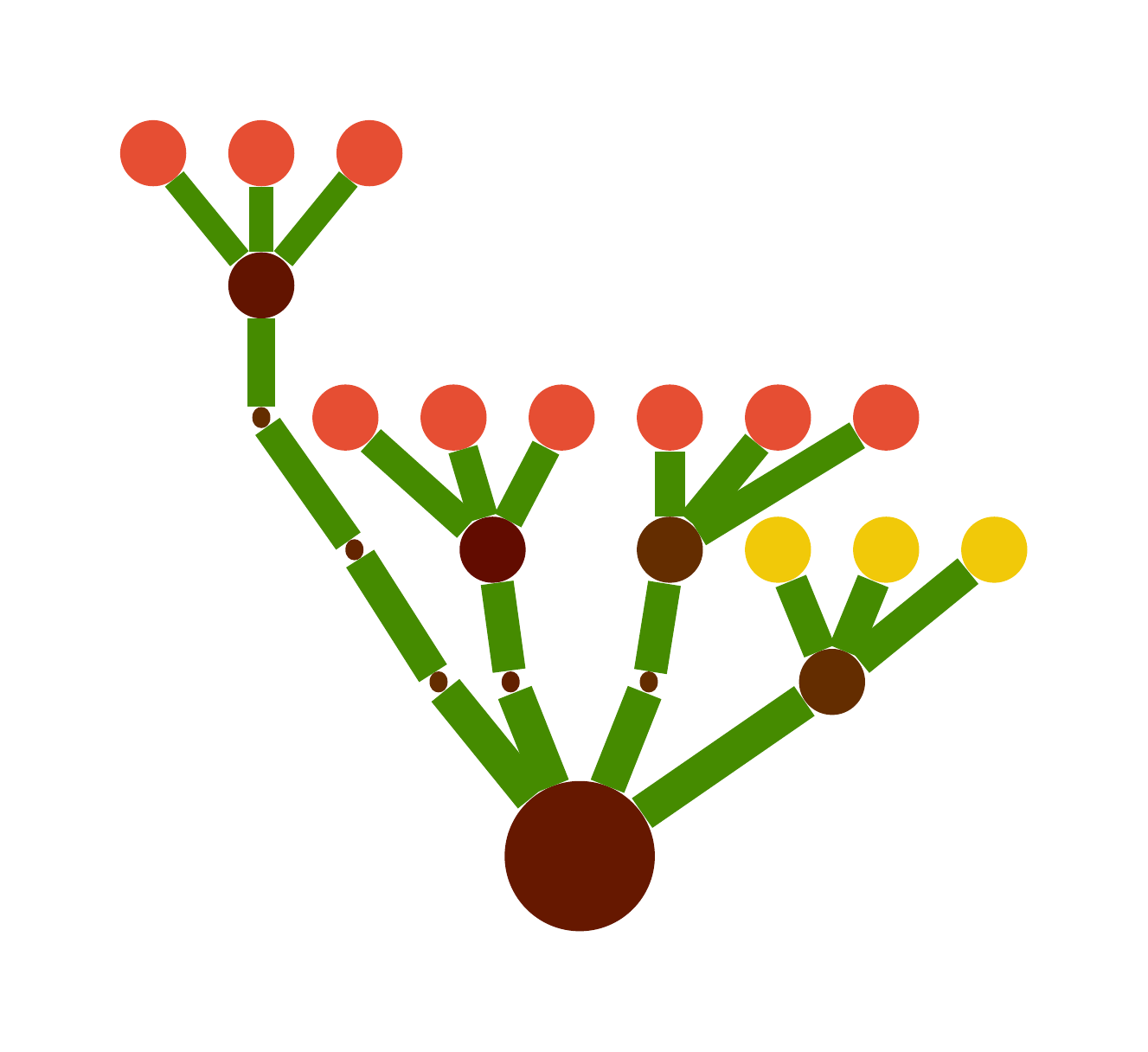}
\includegraphics[scale=.26]{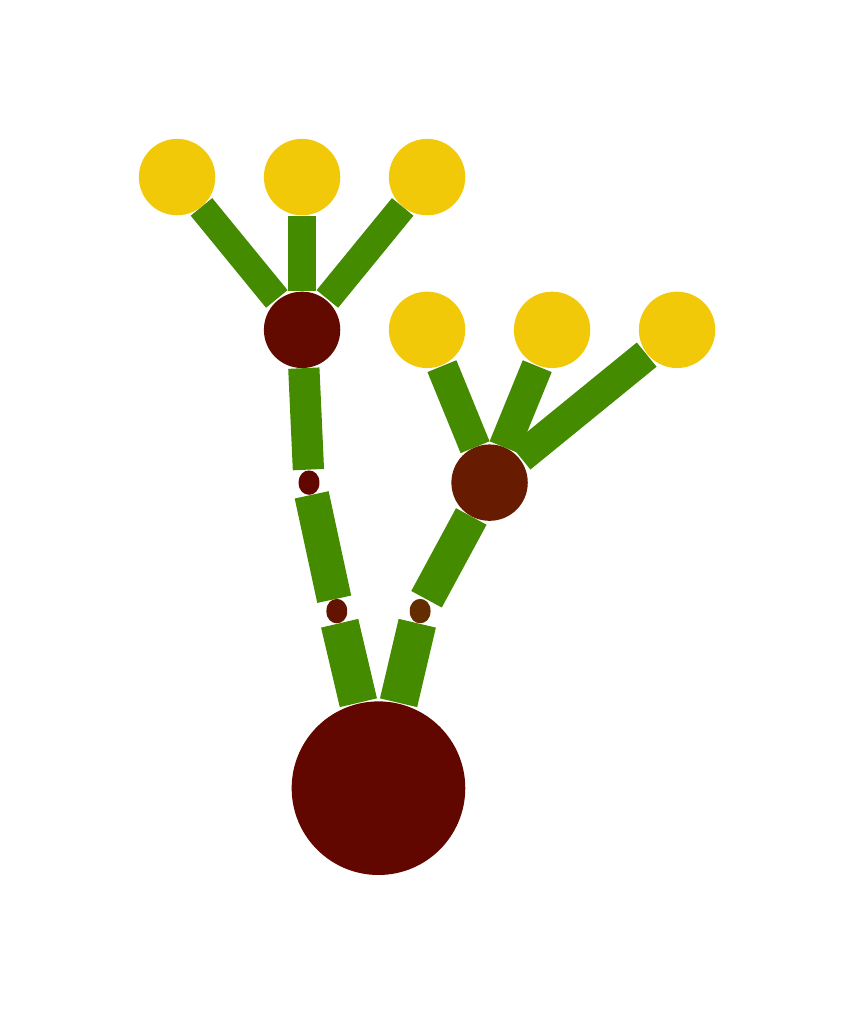}
\includegraphics[scale=.26]{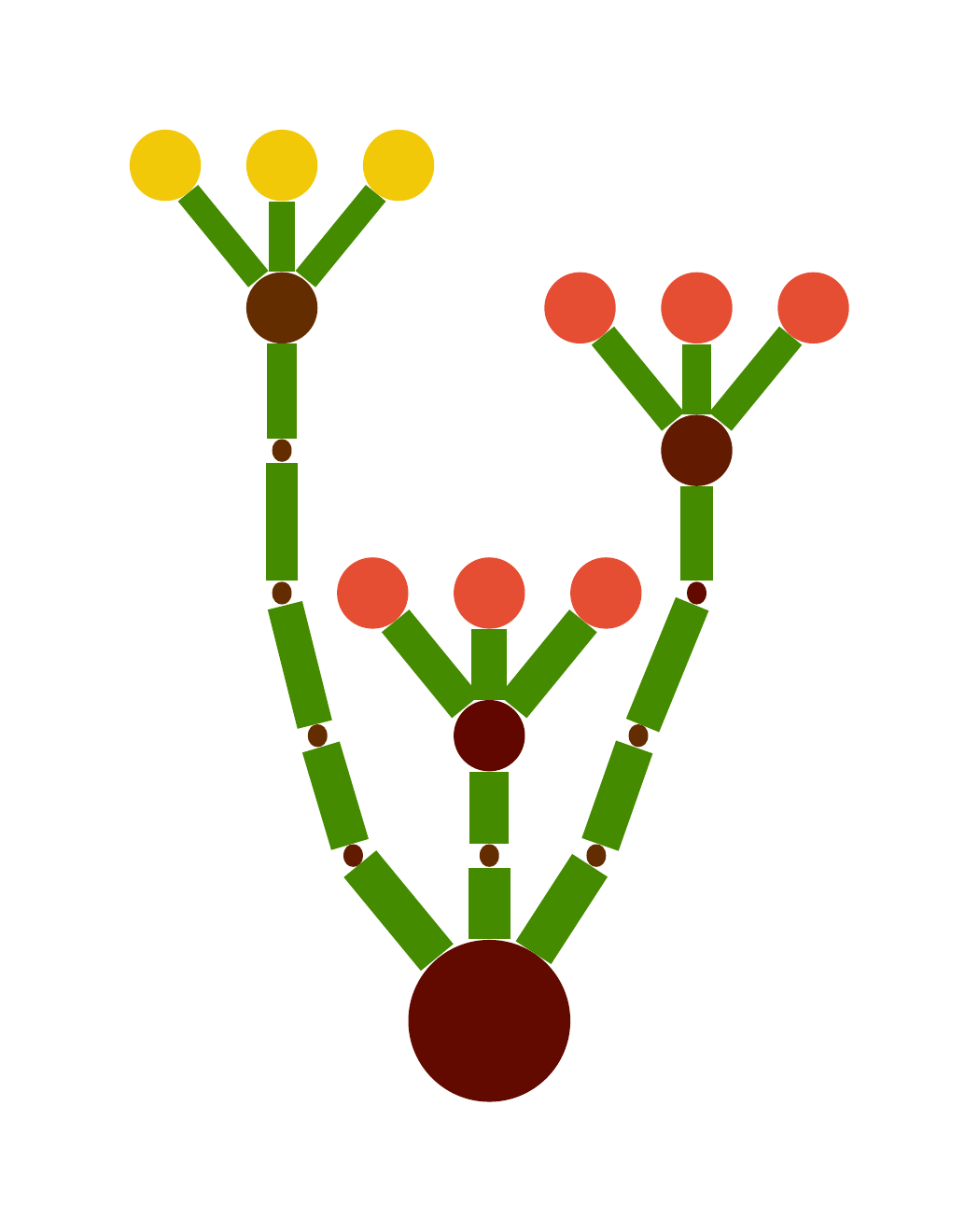}
\includegraphics[scale=.26]{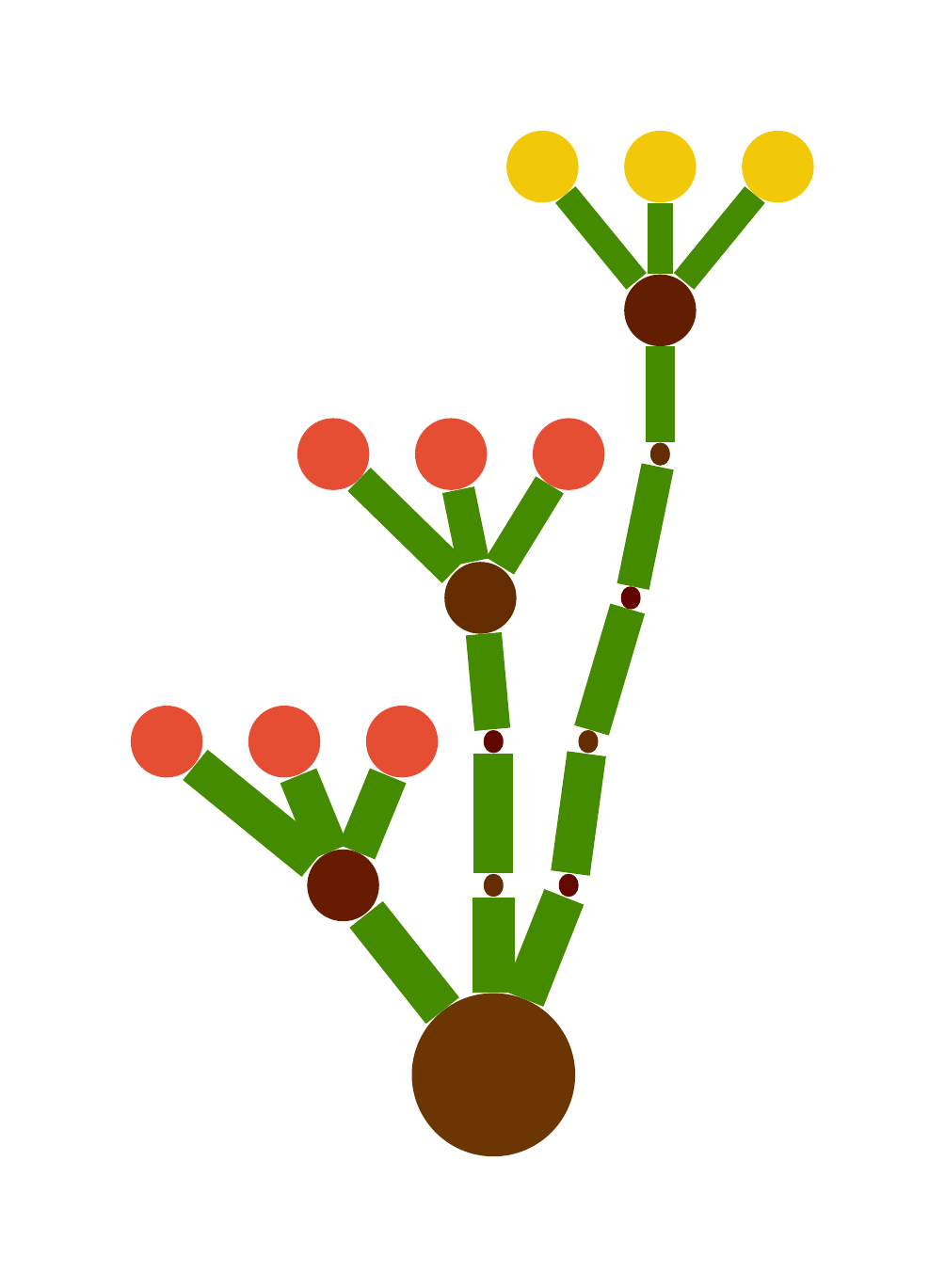}
\end{center}
\caption{Tree-like objects.}
\label{fig:plants}
\end{figure}

What patterns do you see when you look at figure \ref{fig:plants}? You might describe the image as a series of trees, where each tree has a large, brown base and a number of green branches of variable length, with each branch ending in a flower that is either yellow or red. Recognizing such patterns is an important aspect of intelligence, both human and machine.
One way to approach such pattern recognition is as the problem of learning generative models for observed examples. We wish to find a description of the process that gave rise to a set of examples, and we form this description in a rich enough language to capture the abstract patterns---in this report, a probabilistic programming language. 
We build on the representation in \cite{A.Stuhlmueller:2010:6d11a} and explore a family of algorithms for learning probabilistic programs from data.

Generative models play a prominent role in modern machine learning (e.g., Hidden Markov models and probabilistic context-free grammars) and have led to a wide variety of applications.  There is a trade-off between the variety of patterns a model class is able to capture and the feasibility of learning models in that class \cite{Russell2003}.  Much of machine learning has focused on studying classes of models with limited expressiveness in order to develop tractable algorithms for modeling large data sets.  Our investigation takes a different approach and explores how learning might proceed in an expressive class of models with a focus on identifying abstract patterns from small amounts of data.

We represent generative models as programs in a probabilistic programming language. A probabilistic program represents a probability distribution, and each evaluation of the program results in a sample from the distribution. We implement these programs in a subset of the probabilistic programming language Church \cite{N.D.Goodman:2008:f2a0d}.  
These programs can have parameterized functions and recursion, which allow for natural representation of ``long-range'' dependencies and recursive patterns.  We will frame searching this space of models in terms of Bayesian model merging \cite{Stolcke:1994:IPG:645515.658235} and demonstrate that this approach can find interesting patterns in a simple domain of colored trees.

The main components of our approach are as follows: We represent data in an algebraic data type and generative models probabilistic programming language that extends this data type, we guide search through program space using the Bayesian posterior probability. Our algorithm begins by transforming the data into a large program that generates all and only the examples, it then explores program space by identifying (approximately) repeated structure in the large program and transforming the program to make sharing explicit; these search moves are formalized in a family of useful program transformations.
The probabilistic programs learned in this manner can be understood as generative models and reasoning about such models can be formulated in terms of probabilistic inference. We illustrate these ideas on colored trees (i.e. list-structured data).

Before we proceed, a note about what this report is and is not: This report \emph{is} a status update on our working system, containing detailed code and illustrative examples. It documents some progress we have made that we believe can be useful more generally.
This report \emph{is not} a completed academic work. In particular, it does little to situate our work within the context of previous work (some of which has  directly inspired us), provides little high-level discussion, and aims for illustrative examples rather than compelling applications.

\subsection{Bayesian model merging}

Bayesian model merging \cite{Stolcke:1994:IPG:645515.658235} is a framework for searching through a space of generative models in order to find a model that accurately generates the observed data.  The main idea is to search model space through a series of ``merge'' transformations, using the posterior $P(M|D)$ of model $M$ given data $D$ as the criterion for selecting transformations.

We create an initial model by building a program that has a uniform distribution over the training set ({\em data incorporation}).  While this initial model has high likelihood $P(D|M)$, it never generates points outside the training set---it severely overfits the initial data. Therefore, we generate alternative model hypotheses using program transformations that collapse model structure and that often result in better generalizations.

This technique has been successfully applied to learning artificial probabilistic grammars represented as hidden Markov models, n-grams, and probabilistic context-free grammars \cite{Stolcke:1994:IPG:645515.658235}.

\subsection{Bayesian program merging}

We extend Bayesian model merging to models expressed in the rich class of probabilistic programs.  This allows us to represent complex patterns that go beyond context free grammars (for instance, parameterized sub-functions) and to use a wide range of transformations, including transformations that result in ``lossy'' compression of the input data. In the remainder of this document, we will describe the following parts required to implement Bayesian program merging:

\vskip .8em

\noindent {\bf Data representation and program language} \\
\noindent We represent data in terms of an algebraic data type, which gives us a way to form initial programs using the type constructors. Our language for programs consists of type constructors, lambda abstraction, and additional operators. This language is probabilistic, hence programs correspond to distributions on observations. Program structure corresponds to regularities in the observations.

\vskip .8em

\noindent {\bf Search objective: Posterior probability of probabilistic programs} \\
\noindent The objective for our search through program space is the posterior probability of a program given the observed data. This posterior decomposes into a prior based on program length and into a likelihood that we estimate using selective model averaging.

\vskip .8em

\noindent {\bf Search moves: Program transformations} \\
\noindent We describe two types of program transformation that we use as search moves: abstraction and deargumentation. Both transformations collapse program structure, which often increases prior probability of the program and improves generalization to unobserved data.

\vskip .8em

\noindent {\bf Search strategy: Beam search} \\
Given a space of programs, an objective function, and search moves between programs (transformations), we still need to specify a search strategy. We use beam search.

\newpage
\section{Data representation and program language}
We assume that we can represent our data using an algebraic data type. This assumption gives us a starting point for program induction, since any data can be directly translated into a program which is a derivation (sequence of data constructor operations) of the data from the type specification.

\subsection{Data and program representation in the list domain}
We can model the trees shown in figure \ref{fig:plants} in terms of nested lists (s-expressions).  Each tree consists of nodes, where each node has a size and color attributes along with a list of child nodes.
\begin{grammar}
<tree> ::= nil

<tree> ::= (node <data> <tree> <tree> ... )

<data> ::= (data <color> <size>)

<color> ::= (color [number])

<size> ::= (size [number])
\end{grammar}

Figure \ref{fig:tree-obs} shows two expressions in this langauge together with the corresponding visual representations.
We now have a way of representing data as rudimentary programs.  In order to capture interesting patterns, we need a more expressive language.  We use the following subset of Church for Bayesian program merging in the tree domain:
\begin{grammar}
<expr> ::= (begin <expr> <expr> ... ) 
\alt ($\lambda$ (<var> ...) <expr>)
\alt (<expr> <expr>)  
\alt (define <var> <expr>)
\alt (if (flip [number]) <expr> <expr>)
\alt <var> 
\alt <primitive>

<var> ::= V[number] | F[number] 

<primitive> ::= uniform-choice | list | node | data | color | size | [number]
\end{grammar}
To apply the Bayesian program merging techniques to data specified in a different data type the only change would be to replace list, etc., with the constructors of the new data type.

\subsection{Data incorporation}
The first step of Bayesian program merging is data incorporation.  Data incorporation is the creation of an initial model by going through each example in the training set and creating an expression that evaluates to this example (in terms of the algebraic data type constructors).  We combine these programs into a single expression that draws uniformly from this list. (That is, we assume that the observations are i.i.d.) 
\begin{lstlisting}[frame=trblsingle]
(define (incorporate-data trees)
  `($\lbda$ () (uniform-choice ,@(map tree->expression trees))))

(define (tree->expression tree)
  (if (null? tree)
      '()
      (pair 'node
            (pair (node-data->expression (first tree)) 
                  (map tree->expression (rest tree))))))

(define (node-data->expression lst)
  `(data (color (gaussian ,(first (second lst)) 25)) 
	 (size ,(first (third lst)))))
\end{lstlisting}

\begin{figure}[t]
  \begin{minipage}{.7\textwidth}
\begin{lstlisting}[boxpos=b]
(node (data (color 70) (size 0.7))
      (node (data (color 37) (size 0.3))
            (node (data (color 213) (size 0.3)))
            (node (data (color 207) (size 0.3)))
            (node (data (color 211) (size 0.3)))))
(node (data (color 43) (size 0.7))
      (node (data (color 47) (size 0.1))
            (node (data (color 33) (size 0.3))
                  (node (data (color 220) (size 0.3)))
                  (node (data (color 224) (size 0.3)))
                  (node (data (color 207) (size 0.3))))))
\end{lstlisting}
  \end{minipage}    
  \begin{minipage}{.25\textwidth}
    \includegraphics[scale=.26]{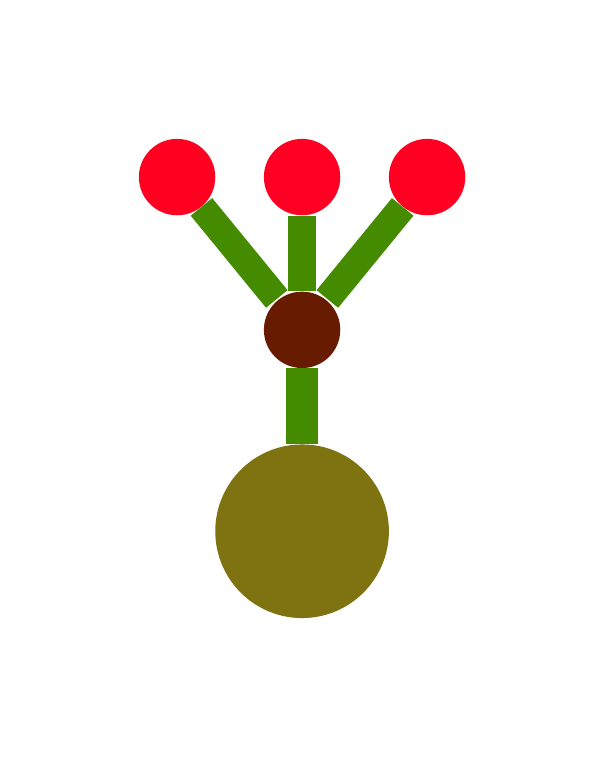}
    \includegraphics[scale=.26]{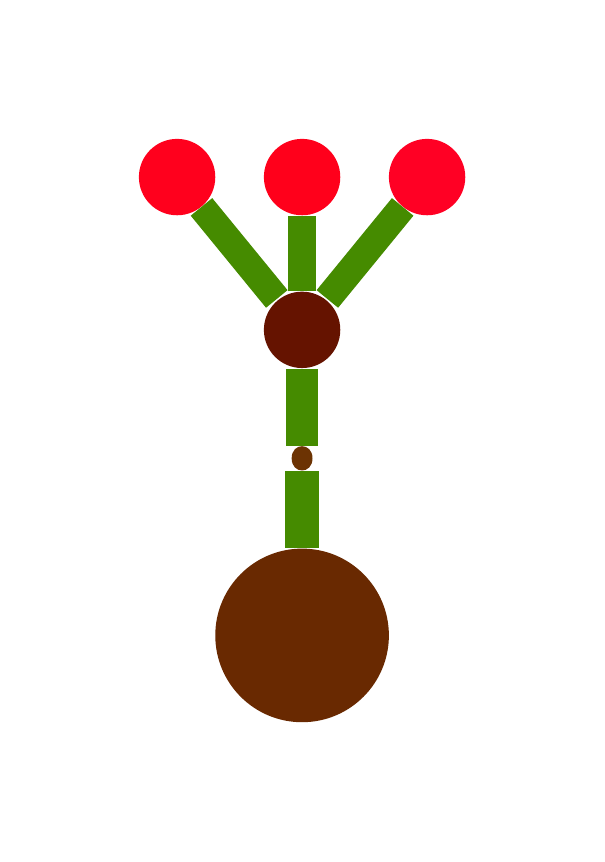}
  \end{minipage}        
\caption{Two tree observations.}
\label{fig:tree-obs}
\end{figure}

\begin{figure}[t]
  \centering
  \includegraphics[scale=.26]{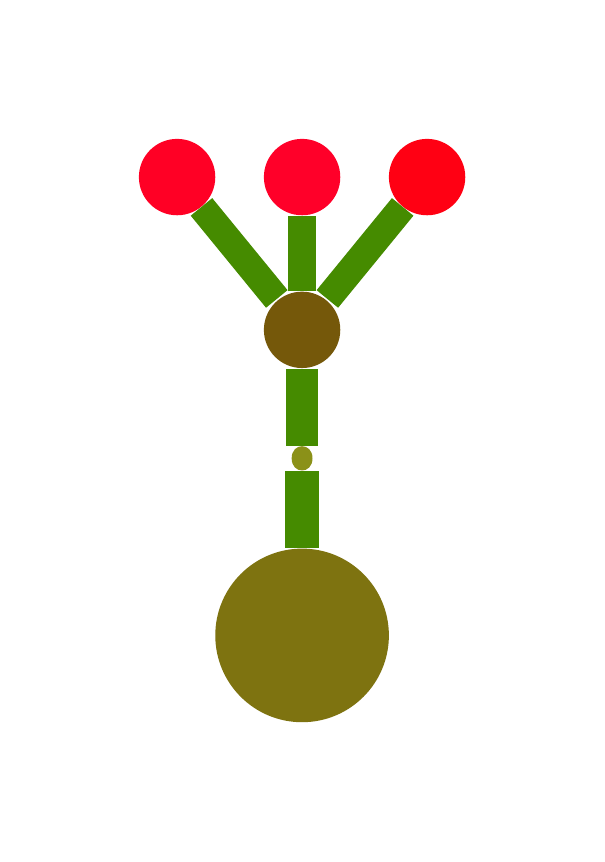}  
  \includegraphics[scale=.26]{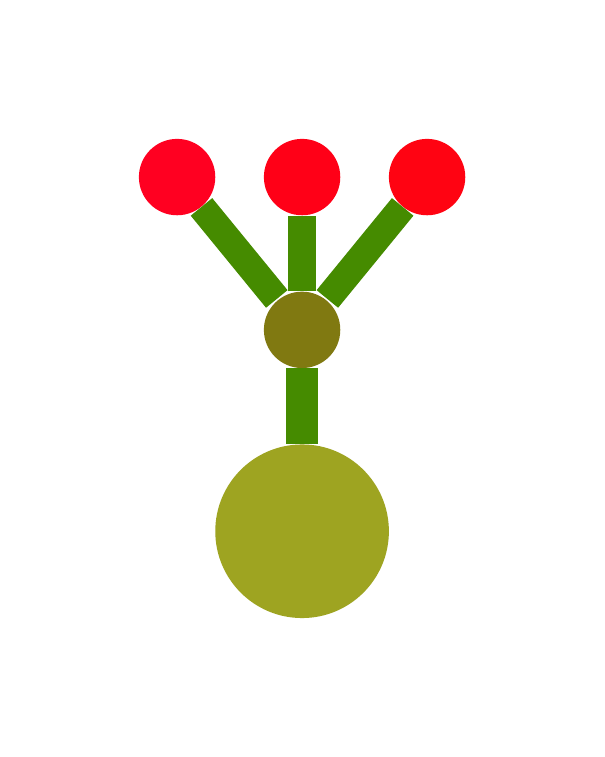}
  \includegraphics[scale=.26]{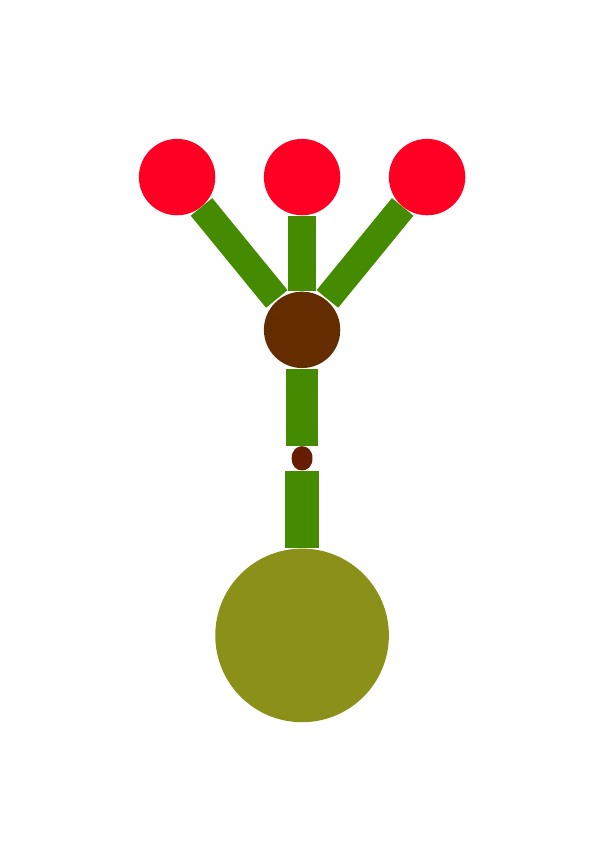}
  \includegraphics[scale=.26]{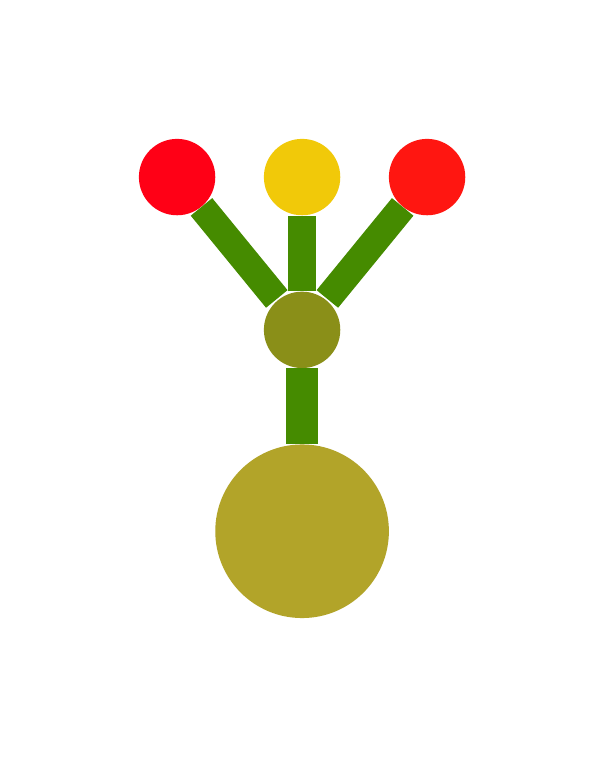}
  \includegraphics[scale=.26]{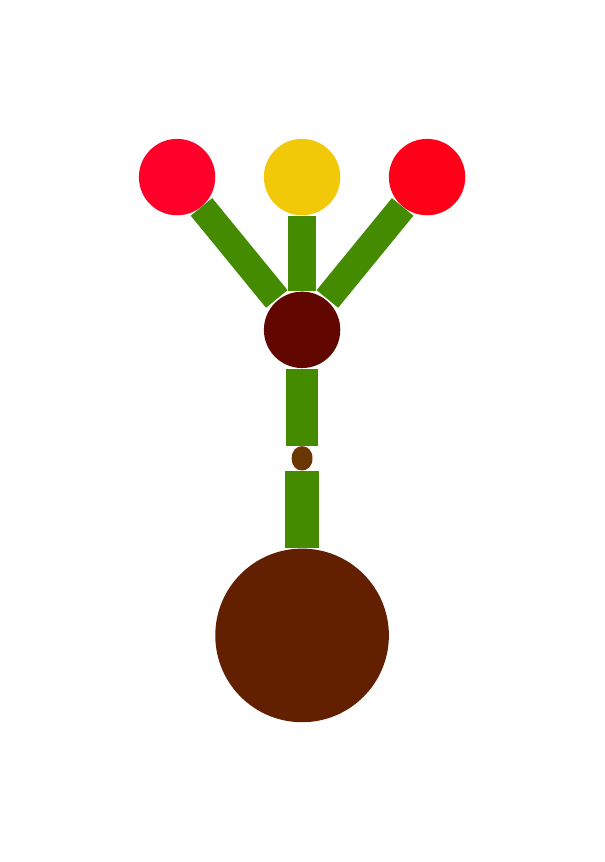}
  \includegraphics[scale=.26]{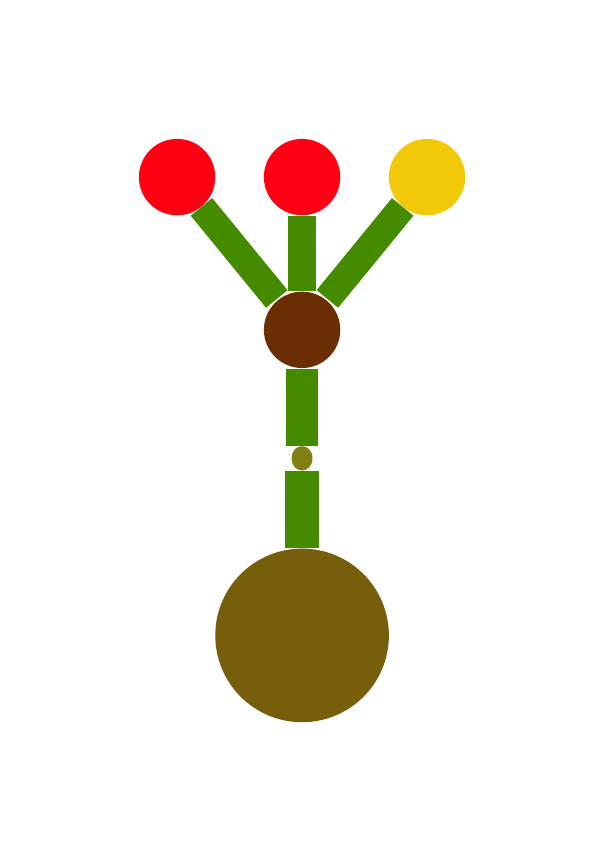}
  \includegraphics[scale=.26]{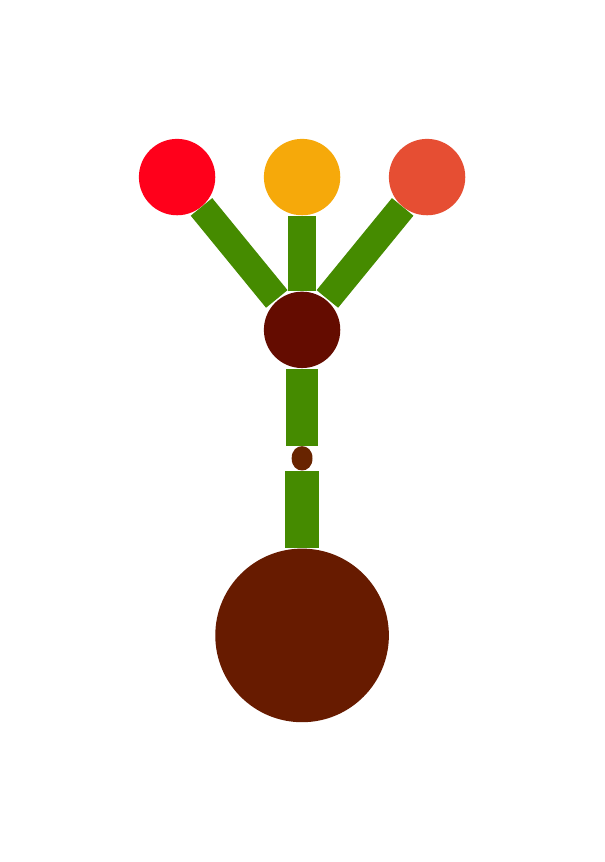}
  \includegraphics[scale=.26]{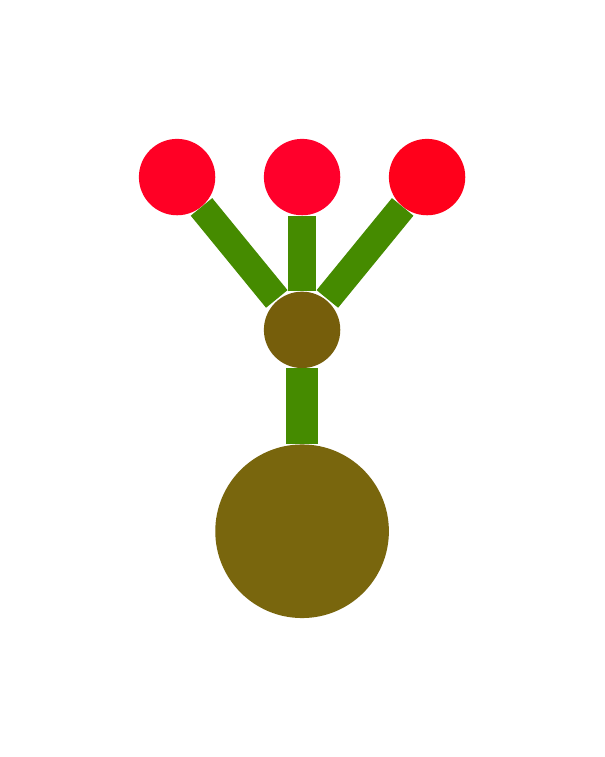}
  \includegraphics[scale=.26]{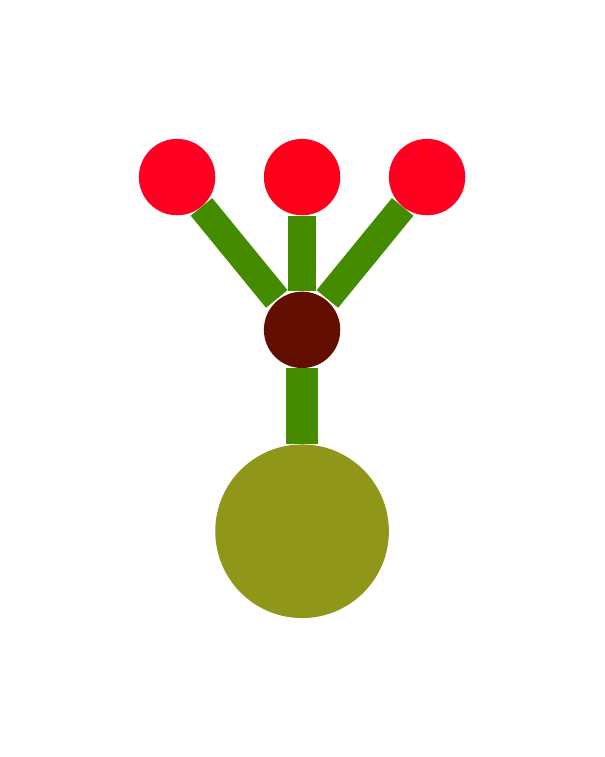}
  \caption{Data incorporation using the observations in figure \ref{fig:tree-obs} results in a program that generates samples such as the ones shown in this figure.}
  \label{fig:initprog}
\end{figure}

In our implementation, the data are unannotated s-expressions (e.g., \texttt{((1) (2))}), which can easily be converted into expressions in terms of the tree type constructors (the tree expression for the above example is \texttt{(node (data (color 1) (size 2)))}).  In this report, we assume for readability that all data is in tree expression form. An important question we leave for future work is how to perform data incorporation when the data is less structured and, for example, given in terms of feature vectors. Calling \texttt{incorporate-data} on the program shown in figure \ref{fig:tree-obs} results in the following program:
\begin{lstlisting}[mathescape=true]
($\lbda$ ()
  (uniform-choice
   (node (data (color (gaussian 70 25)) (size .7))
         (node (data (color (gaussian 37 25)) (size 0.3))
               (node (data (color (gaussian 213 25)) (size 0.3)))
               (node (data (color (gaussian 207 25)) (size 0.3)))
               (node (data (color (gaussian 211 25)) (size 0.3)))))
   (node (data (color (gaussian  43)) (size .7))
         (node (data (color (gaussian 47 25)) (size 0.1))
               (node (data (color (gaussian 33 25)) (size 0.3))
                     (node (data (color (gaussian 220 25)) (size 0.3)))
                     (node (data (color (gaussian 224 25)) (size 0.3)))
                     (node (data (color (gaussian 207 25)) (size 0.3))))))))
\end{lstlisting}
Samples generated from this program are shown in figure \ref{fig:initprog}.

\newpage
\section{Search objective: Posterior probability of probabilistic programs}
A generative model is a joint probability distribution over latent states and observed data. We can represent such a joint distribution as a program in a probabilistic programming language like Church \cite{N.D.Goodman:2008:f2a0d}. The structure of the program, i.e., the decomposition into functions and the flow of control, can capture regularities in the data. We illustrate this idea using a probabilistic program that generates the images in figure \ref{fig:plants}. Different parts of the program correspond to the different patterns we described earlier (to improve readability, we use semantically meaningful function and variable names instead of \texttt{F[number]} and \texttt{V[number]} as specified in our grammar).  
\begin{lstlisting}[mathescape=true]
(define (tree)
  (uniform-choice
   (node (body) (branch))
   (node (body) (branch) (branch))
   (node (body) (branch) (branch) (branch))
   (node (body) (branch) (branch) (branch) (branch))))

(define (body) 
  (data (color (gaussian 30 10)) (size .7)))

(define (branch)
  (if (flip .2) 
      (flower (if (flip .5) 150 255))
      (node (branch-info) (branch))))

(define (branch-info)
  (data (color (gaussian 0 25)) (size .1)))

(define (flower shade)
  (node (data (color (gaussian 0 25)) (size .3))
        (petal shade)
        (petal shade)
        (petal shade)))

(define (petal shade)
  (node (data (color shade) (size .3))))
\end{lstlisting}
Our programs are a combination of data constructors, control flow operations, and lambda abstractions.  When we call \texttt{(tree)}, the program shown above first determines the number of branches the tree will have and then creates a large body node that connects these branches.  Each branch function recursively connects a series of small nodes together and ends in a call to the flower function, passing it one of two colors.  The flower function creates a node with three identically colored ``petal'' nodes.  The program structure reflects patterns such as a flower having three petals and the body having a large base.  The compositionality of the language captures relational patterns such as the fact that branches end in flowers. Given that programs can represent such structures, our goal is to transform the initial program generated by data incorporation into such a more structured form.

\subsection{Abstract syntax utilities}
Some program transformations generate new functions. We call such functions {\em abstractions}. We represent them using a name, argument variables, and a pattern, i.e., the s-expression that makes up the body of the function. In the following code snippet, we use the \texttt{sym} function that creates readable unique symbols for function and variable names.
\begin{lstlisting}[frame=trbl]
(define (make-abstraction body variables)
  (make-named-abstraction (sym FUNC-SYMBOL) body variables))
(define (make-named-abstraction name body variables)
  (list 'abstraction name variables body))
(define abstraction->name second)
(define abstraction->vars third)
(define abstraction->body fourth)
\end{lstlisting}

{\em Programs} represent the generative models we search over. They consist of a list of abstractions and a body.
\begin{lstlisting}[frame=trbl]
(define (make-program abstractions body)
  (list 'program abstractions body))
(define program->abstractions second)
(define program->body third)
\end{lstlisting}  

We wrap programs into another data type to keep track of additional information during search. We motivate this in the section on search in more detail. The basic idea is to avoid recomputation of a program's likelihood (an expensive computation) when transformations do not affect a program's semantics.
\begin{lstlisting}[frame=trbl]
(define (make-program+ program posterior log-likelihood log-prior semantics-preserved)
  (list 'program+ program posterior log-likelihood log-prior 
        semantics-preserved))
(define program+->program second)
(define program+->posterior third)
(define program+->log-likelihood fourth)
(define program+->log-prior fifth)
(define program+->semantics-preserved sixth)
(define (program+->program-transform semantics-preserved program+ new-program)
  (make-program+ new-program 
                 (program+->posterior program+) 
                 (program+->log-likelihood program+) 
                 (program+->log-prior program+) 
                 semantics-preserved))
\end{lstlisting}

\subsection{Posterior probability}

Probabilistic programs correspond to probability distributions on observed data, i.e., we can compute the likelihood of a given observation under a program. Given a process for generating programs (a prior), we use Bayes theorem to compute the posterior probability of a program:
\begin{equation}P(M|D)\propto P(D|M)P(M)\end{equation}
Here, $P(D|M)$ is the probability that program $M$ generates data $D$, the likelihood, and $P(M)$ is the prior probability of program $M$. We use a prior based on program length:
\begin{equation}P(M)\propto e^{-\alpha \, \mathrm{size}(M)}\end{equation}
\begin{lstlisting}[frame=trbl]
(define (log-prior program)
  (- (* alpha (program-size program))))
\end{lstlisting}
This prior biases the search towards smaller programs. Increasing the constant $\alpha$ gives the prior more weight when calculating the posterior, which means that minimizing program size is a more important criterion. A program's size is the number of symbols in the function bodies as well as in the main body.
\begin{lstlisting}[frame=trbl]
(define (sexpr-size sexpr)
  (if (list? sexpr)
      (apply + (map sexpr-size sexpr))
      1))

(define (abstraction-size abstraction)
  (sexpr-size (abstraction->body abstraction)))
  
(define (program-size program)
  (let* ([abstraction-sizes (apply + (map abstraction-size (program->abstractions program)))]
         [body-size (sexpr-size (program->body program))])
    (+ abstraction-sizes body-size)))
\end{lstlisting}
Computing the likelihood is the difficult part of the posterior probability computation. Intuitively, we can think of the likelihood as tracking how good a particular program is at producing a set of target data.  This is important in search, since it gives precise, quantitative information on whether and how to adjust our hypothesis program. However, since there may be a large number of possible settings for the random choices of a program, this can make determining which choices lead to the observed data difficult. Furthermore, for any given data point, there could be multiple settings that generate this data point; to correctly compute the likelihood, we need to take into account all of them. Often, we cannot compute this quantity exactly due to limited computational resources.  In the following, we describe a stochastic approximation of this computation for list data. The fact that our likelihood estimation procedure is stochastic means that the final model learned with our implementation of Bayesian program merging may differ even for two runs on the same input data.

\subsection{Likelihood estimation in the list domain}

In the case of programs that generate list-structured data, we can estimate the likelihood by (1) generating samples using a sequential Monte Carlo method (\texttt{smc}) that generates the discrete structure of the examples, (2) extending each sample by forcing it to generate the continuous parameters, and (3) applying selective model averaging to these samples. We begin by factoring the problem of generating the data set into the problem of generating each of the data points:
\begin{eqnarray}
P(T|M) &=& \prod_{t \in M}P(t|M)
\end{eqnarray}
Here, $T$ is the observed set of trees, $t$ a single tree, and $M$ the generative model (program).
\begin{lstlisting}[frame=trbl]
(define (log-likelihood trees prog sample-size)
  (apply + (map ($\lbda$ (tree) (single-log-likelihood prog sample-size tree)) trees)))
\end{lstlisting}
We will estimate the likelihood of a single tree, $P(t|M)$, by evaluating the program many times, restricting the computation to result in the target tree each time. The product of the probabilities of all random choices made during a single evaluation corresponds to the probability of a single possible way of generating the tree. Since there may be multiple ways for a program to generate a given tree, we sum up the probability of each {\em distinct} parse to compute a lower bound on the true likelihood (selective model averaging).
\begin{lstlisting}[frame=trbl]
(define (single-log-likelihood program popsize tree)
  (let* ([program* (replace-gaussian (desugar program))]
         [model (eval (program->sexpr program*))]
         [topology-scores+tree-parameters (compute-topology-scores+evaluate model tree popsize)]
         [topology-scores (first topology-scores+tree-parameters)]
         [trees-with-parameters (second topology-scores+tree-parameters)]
         [data-scores (map ($\lbda$ (tree-with-parameters) (compute-data-score tree tree-with-parameters)) trees-with-parameters)]
         [scores (delete -inf.0 (map + topology-scores data-scores))]
         [score (if (null? scores)
                     -inf.0
                     (apply log-sum-exp scores))])
    score))
\end{lstlisting}
We take advantage of the fact we can directly compute the probability of a sample from a Gaussian given the parameters.  We therefore modify the Gaussian functions in the program to output the mean and variance for a particular node instead of sampling a value from the distribution.  The code below also changes the \texttt{uniform-choice} syntactic construct into a \texttt{uniform-draw}, which current Church implementations provide.
\begin{lstlisting}[frame=trbl]
(define (replace-gaussian program)
  (define (gaussian? sexpr)
    (tagged-list? sexpr 'gaussian))
  (define (return-parameters sexpr)
    `(list 'gaussian-parameters ,(second sexpr) ,(third sexpr)))
  (define (replace-in-abstraction abstraction)
    (make-named-abstraction (abstraction->name abstraction) (transform-sexp gaussian? return-parameters (abstraction->pattern abstraction)) (abstraction->vars abstraction)))
  (let* ([converted-abstractions (map replace-in-abstraction (program->abstractions program))]
	 [converted-body (transform-sexp gaussian? return-parameters (program->body program))])
    (make-program converted-abstractions converted-body)))

(define (desugar program)
  (define (uniform-choice? sexpr)
    (tagged-list? sexpr 'uniform-choice))
  (define (uniform-draw-conversion sexpr)
    `((uniform-draw (list ,@(map thunkify (rest sexpr))))))
  (define tests+replacements (zip (list uniform-choice?) (list uniform-draw-conversion)))
  (define (apply-transforms sexpr)
    (fold ($\lbda$ (test+replacement expr)
	    (transform-sexp (first test+replacement) (second test+replacement) expr))
	  sexpr
	  tests+replacements))
  (define (desugar-abstraction abstraction)
    (make-named-abstraction (abstraction->name abstraction) (apply-transforms (abstraction->pattern abstraction)) (abstraction->vars abstraction)))
  (let* ([converted-abstractions (map  desugar-abstraction (program->abstractions program))]
	 [converted-body (apply-transforms (program->body program))])
    (make-program converted-abstractions converted-body)))

(define (thunkify sexpr) `($\lbda$ () ,sexpr))
\end{lstlisting}
In the code below, \texttt{smc-core} forces the program to generate the desired data.  A detailed description of \texttt{smc-core} is beyond the scope of this report. In short, the method is an incremental forward sampler (with re-sampling); we separate out the continuous choices since forward sampling would have probability $0$ of generating the observed real values.
\begin{lstlisting}[frame=trbl]
(define (compute-topology-scores+evaluate model tree popsize)
  (let* ([smc-core-arguments (create-smc-core-args model tree popsize)]
         [samples (apply smc-core smc-core-arguments)]
         [repeat-symbol (find-repeat-symbol samples)]
         [unique-samples
          (fold ($\lbda$ (s a)
                  (if (member (mcmc-state->addrval s repeat-symbol)
                              (map ($\lbda$ (x) (mcmc-state->addrval x repeat-symbol)) a))
                      a (pair s a))) '() samples)]
         [topology-scores (map mcmc-state->score unique-samples)]
         [generated-trees (map mcmc-state->query-value unique-samples)])
    (list topology-scores generated-trees)))
\end{lstlisting}
We use the Gaussian parameters for each node's color to determine the probability of the observed color values of a tree:
\begin{lstlisting}[frame=trbl]
(define (compute-data-score tree tree-with-parameters)
    (if (null? tree)
        0
        (+ (single-data-score (node->data tree) (node->data tree-with-parameters)) (apply + (map compute-data-score (node->children tree) (node->children tree-with-parameters))))))

(define (single-data-score original-data parameterized-data)
  (let* ([color-score (score-attribute (data->color original-data) (data->color parameterized-data))]
         [size-score (score-attribute (data->size original-data) (data->size parameterized-data))])
    (+ color-score size-score)))

(define (score-attribute original-attribute parameterized-attribute)
  (if (tagged-list? parameterized-attribute 'gaussian-parameters)
      (log (normal-pdf (first original-attribute) (gaussian->mean parameterized-attribute) (gaussian->variance parameterized-attribute)))
      (if (= (first original-attribute) (first parameterized-attribute)) 0 -inf.0)))

(define gaussian->mean second)
(define gaussian->variance third)
\end{lstlisting}

\newpage
\section{Search moves: Program transformations}

We start with a program constructed by data incorporation and we want to propose changes such that its posterior probability improves. We will achieve this  using program transformations that isolate patterns and compress the program, thereby increasing its prior probability. In the following, we describe two such transformations, abstraction and deargumentation.

\subsection{Abstraction}

Abstraction aims to create new functions based on syntactic patterns in a program and replace these patterns with calls to the newly created functions.
This removes duplication in the code and acts as a proxy for recognizing repeated computation.
In terms of Bayesian model merging, this transformation merges the structure of the model, which potentially leads to models with better generalization properties \cite{Stolcke:1994:IPG:645515.658235}. We can also interpret this process as finding partial symmetries as in the work of Bokeloh et al. \cite{DBLP:journals/tog/BokelohWS10}, which was successful in the domain of inverse-procedural modeling.

The following code fragment implements this procedure. The function \texttt{compressions} finds all (lambda) abstractions that can be formed by anti-unifying (partially matching) pairs of subexpressions in a condensed form of the program (only the bodies of the functions and the body of the program). We filter out duplicate abstractions and then create compressed programs by replacing occurrences of the abstraction bodies in the original program (unification).

\begin{lstlisting}[frame=trbl]
(define (compressions program . nofilter)
  (let* ([condensed-program (condense-program program)]
         [abstractions (possible-abstractions condensed-program)]
         [compressed-programs (map (curry compress-program program) abstractions)]
         [prog-size (program-size program)]
         [valid-compressed-programs
          (if (not (null? nofilter))
              compressed-programs
              (filter ($\lbda$ (cp)
                        (<= (program-size cp)
                            (+ prog-size 1)))
                      compressed-programs))])
    valid-compressed-programs))

(define (condense-program program)
  `(,@(map abstraction->body (program->abstractions program))
    ,(program->body program)))

(define (possible-abstractions expr)
  (let* ([subexpr-pairs (list-unique-commutative-pairs (all-subexprs expr))]
         [abstractions (map-apply (curry anti-unify-abstraction expr) subexpr-pairs)])
    (filter-abstractions abstractions)))
\end{lstlisting}

The following example illustrates the abstraction transformation on a language slightly simpler than the tree example (no sizes or continuous properties).
\begin{lstlisting}[mathescape=true]
(uniform-choice 
 (node a (node a (node b) (node b)))
 (node a (node a (node c) (node c))))
\end{lstlisting}
A transformed version of this program looks like this:
\begin{lstlisting}[mathescape=true]
(begin
  (define (F1 V1 V2)
    (node a (node a (node V1) (node V2))))
  (uniform-choice (F1 b b) (F1 c c)))
\end{lstlisting}
Both programs have the same behavior, i.e., this transformation preserves semantics. Both programs return \texttt{(a (a (b) (b)))} and \texttt{(a (a (c) (c)))} with equal probability.

This transformation refactors a program with subexpressions that partially match into a program with a function which contains the common parts of the matching subexpressions as its body.  In the example shown above, the subexpressions that partially match are \texttt{(node a (node a (node b) (node b)))} and \texttt{(node a (node a (node c) (node c)))}.  The common subexpression is \texttt{(node a (node a (node x) (node y)))}. The function created using this common subexpression is \texttt{F1} and the original subexpressions are replaced with \texttt{(F1 b b)} and \texttt{(F1 c c)}. (There is further potential for compression here, due to the repeated arguments, which we will address using the deargumentation transformation.)

\subsubsection{Anti-unification}

There exists an abstraction transformation for each pair of subexpressions that have a partial match.  In the case of \texttt{(+ (+ 2 2) (- 2 5))} the following pairs of subexpressions have a partial match: \texttt{[2,2], [(+ 2 2), (- 2 5)], [(+ 2 2), (+ (+ 2 2) (- 2 5))]}. The process of finding a partial match between two expressions is called anti-unification.

\begin{figure}[t]
\includegraphics[scale=.30]{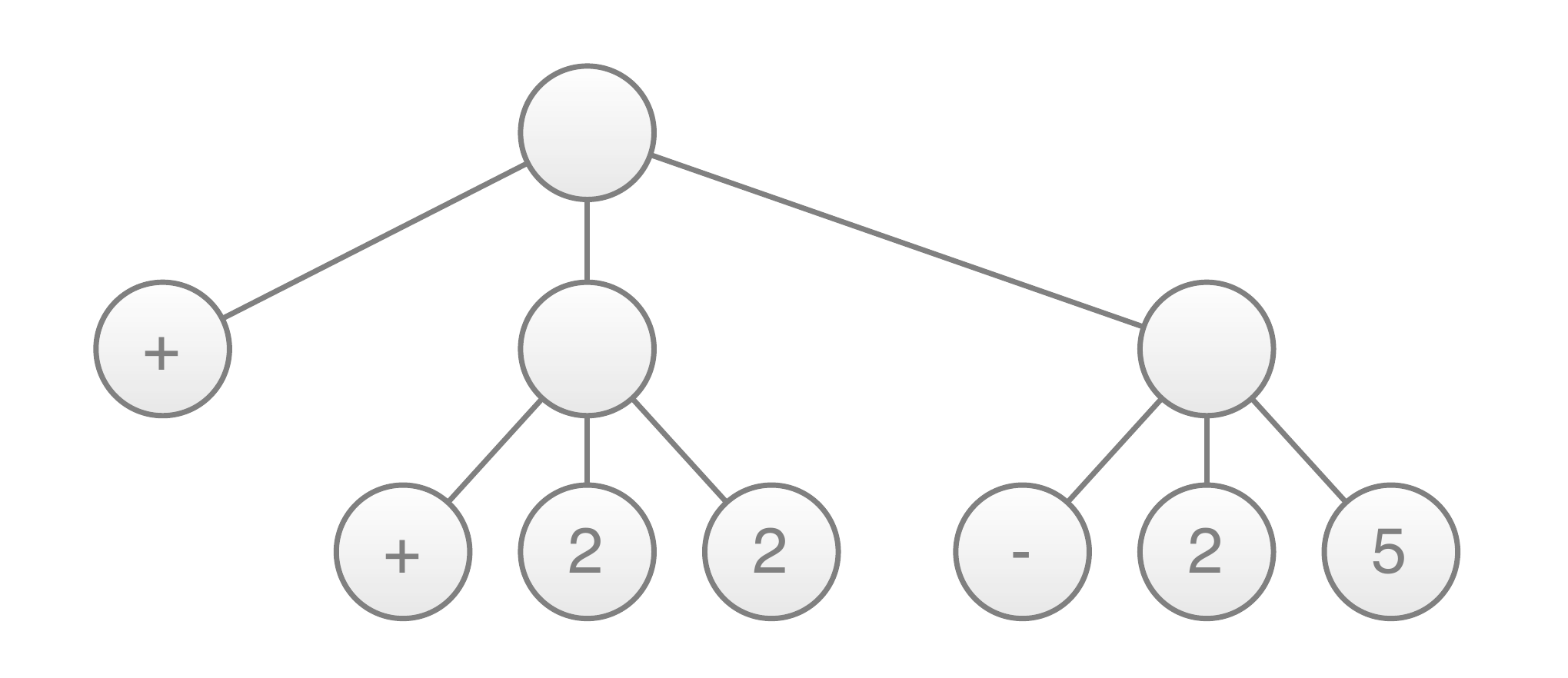}
  \caption{The expression \texttt{(+ (+ 2 2) (- 2 5))} represented as a tree. Anti-unification recursively finds common subtrees between two syntax trees. Matching this tree against itself, there are three partial matches: \texttt{[2,2]}, \texttt{[(+ 2 2), (- 2 5)]}, and \texttt{[(+ 2 2), (+ (+ 2 2) (- 2 5))]}.}
  \label{expressionTree}
\end{figure}

One way to understand the process is in terms of the syntax trees for the expressions.  Every s-expression is a tree where the lists and sublists of the s-expression make up the interior nodes and the primitive elements of the lists (e.g., symbols, numbers) are the leaves.  The tree in figure \ref{expressionTree} corresponds to the expression \texttt{(+ (+ 2 2) (- 2 5))}. We can find a partial match between two expressions by finding a common subtree between their tree representations.
\thisfloatsetup{capposition=beside,capbesideposition={top,outside},facing=no}

Anti-unification proceeds by recursively comparing two expressions, $A$ and $B$ (using the function \texttt{build-pattern}).  If $A$ and $B$ are the same primitive, this primitive is returned.  If $A$ and $B$ are lists $(a_1,\ldots,a_n)$ and $(b_1,\ldots,b_n)$ of the same length, a list $(c_1,\ldots,c_n)$ is returned, with each element $c_i$ being the anti-unification of $a_i$ and $b_i$.  Otherwise, a variable is returned.

\begin{lstlisting}[frame=trbl]
(define anti-unify
  ($\lbda$ (expr1 expr2)
    (begin
      (define variables '())

      (define (add-variable!)
        (set! variables (pair (sym (var-symbol)) variables))
        (first variables))
      
      (define (build-pattern expr1 expr2)
        (cond [(and (primitive? expr1) (primitive? expr2)) (if (equal? expr1 expr2) expr1 (add-variable!))]
              [(or (primitive? expr1) (primitive? expr2)) (add-variable!)]
              [(not (eqv? (length expr1) (length expr2))) (add-variable!)]
              [else
               (let ([unified-expr (map ($\lbda$ (subexpr1 subexpr2) (build-pattern subexpr1 subexpr2))
                                        expr1 expr2)])
                 unified-expr)]))
      (let ([pattern (build-pattern expr1 expr2)])
        (list pattern (reverse variables))))))
\end{lstlisting}

We now illustrate the process of anti-unification using the expressions \texttt{(+ (+ 2 2) (- 2 5))} and \texttt{(+ (- 2 3) 4)}:
\begin{enumerate}
\item We first compare the root of the trees and make sure that we have lists of the same size. In this case, they are both of size $3$, therefore we have matching roots. \\
  Partial match: \texttt{(* * *)}
\item We recursively attempt to match the three subexpressions: \texttt{+} against \texttt{+}, \texttt{(+ 2 2)} against \texttt{(- 2 3)}, and \texttt{(- 2 )} with \texttt{4}. Since \texttt{+} and \texttt{+} are identical primitives, they match. \\
  Partial match: \texttt{(+ * *)}
\item Comparing \texttt{(+ 2 2)} and \texttt{(- 2 3)}, we see that they are both lists of size 3 and therefore they match. \\
  Partial match: \texttt{(+ (* * *) *)}
\item Again, we recursively match subexpressions of \texttt{(+ 2 2)} and \texttt{(- 2 3)}, i.e., \texttt{+} against \texttt{-}, \texttt{2} against \texttt{2}, and \texttt{2} against \texttt{3}. Since \texttt{+} and \texttt{-} are primitives that do not match, we replace them with a variable. \\
  Partial match: \texttt{(+ (V1 * *) *)}
\item We compare \texttt{2} to \texttt{2} and \texttt{2} to \texttt{3}. \\
  Partial match: \texttt{(+ (V1 2 V2) *)}
\item We compare \texttt{(- 2 5)} and \texttt{4} and find that there is no match since \texttt{(- 2 5)} is a list and \texttt{4} is a primitive. \\
  Final match: \texttt{(+ (V1 2 V2) V3)}.
\end{enumerate}

\subsubsection{Refactoring programs}
Using the lambda abstractions we created from the partial matches between two subexpressions of some program $P$, we now attempt to refactor the program, possibly compressing it. For a given abstraction, we take the abstraction body and replace all subexpressions of $P$ that match this pattern by a function call.  We apply this replacement to all functions in $P$ as well as to the body of $P$ and insert the definition of our abstraction into the body of $P$ to generate a refactored program $P'$.

\begin{lstlisting}[frame=trbl]
(define (compress-program program abstraction)
  (let* ([compressed-abstractions (map (curry compress-abstraction abstraction) (program->abstractions program))]
         [compressed-body (replace-matches (program->body program) abstraction)])
    (make-program (pair abstraction compressed-abstractions)
                  compressed-body)))

(define (compress-abstraction compressor compressee)
  (make-named-abstraction (abstraction->name compressee)
                          (replace-matches (abstraction->body compressee) compressor)
                          (abstraction->vars compressee)))                           
\end{lstlisting}
We find and replace pattern matches of an abstraction $F$ in an expression $E$ by recursively matching the body of $F$ against $E$ using unification. If a match exists, then we return a function call to $F$. If there is no match, we return $E$ with matches replaced in each of its subexpressions. If $E$ is a non-matching primitive, we return $E$.

\begin{lstlisting}[frame=trbl]
(define (replace-matches s abstraction)
  (let ([unified-vars (unify s
                             (abstraction->body abstraction)
                             (abstraction->vars abstraction))])
    (if (false? unified-vars)
        (if (list? s)
            (map ($\lbda$ (si) (replace-matches si abstraction)) s)
            s)
        (pair (abstraction->name abstraction)
              (map ($\lbda$ (var) (replace-matches (rest (assq var unified-vars)) abstraction))
                   (abstraction->vars abstraction))))))
\end{lstlisting}
We illustrate refactoring using the expression \texttt{(+ (+ 2 2) (- 2 5))}. The partial match resulting from anti-unification between the subexpressions \texttt{[(+ (+ 2 2) (- 2 5)), (+ 2 2)]} is \texttt{(+ V1 V2)}.  We refactor the original expression \texttt{(+ (+ 2 2) (- 2 5))} in terms of \texttt{(+ V1 V2)} by creating a function \texttt{(define (F1 V1 V2) (+ V1 V2))} and replacing occurrences of its body in the original expression.  For example, one such replacement is \texttt{(F1 (+ 2 2) (- 2 5))}, another is \texttt{(+ (F1 2 2) (- 2 5))}.  In general, we apply the function wherever possible and get a refactored program such as:
\begin{lstlisting}
(+ (+ 2 2) (- 2 5))
=>
(begin
  (define (F1 V1 V2) (+ V1 V2))
  (F1 (F1 2 2) (- 2 5)))
\end{lstlisting}
The input to the refactoring procedure is a function $F$ created from anti-unification and an expression $E$ which will be refactored in terms of $F$.  In the example above, $F$ is \texttt{(define (F1 V1 V2) (+ V1 V2))}, $E$ is \texttt{(+ (+ 2 2) (- 2 5))}, and the result of refactoring is
\begin{lstlisting}
(begin
  (define (F1 V1 V2) (+ V1 V2))
  (F1 (F1 2 2) (- 2 5)))
\end{lstlisting}

In the example, \texttt{(+ (+ 2 2) (- 2 5))} matches \texttt{(+ V1 V2)}, therefore we return an application of \texttt{F1} to arguments \texttt{(+ 2 2)} and \texttt{(- 2 5)}, resulting in \texttt{(F1 (F1 2 2) (- 2 5))}\footnote{The function \texttt{F1} is equivalent to the addition function between two numbers, therefore refactoring the program in terms of this function does not compress the original expression.  An example where refactoring can compress an expression is the expression pair \texttt{[(+ (+ 2 2) (- 2 5)), (+ (+ 2 2) (- 2 6))]}.}.

\subsubsection{Unification}

The problem of determining whether there is a match between an expression $E$ and an abstraction $F$ is known as unification \cite{Robinson:1965:MLB:321250.321253}. We have described anti-unification, the process of creating an abstraction given a pair of expressions. Unification is the opposite of this process.  The return value of successful unification is a list of assignments for the arguments of $F$ such that $F$ applied to these arguments results in $E$.

The unification algorithm recursively checks whether the body of $F$ and $E$ are lists of the same size.  If they are, then unification returns a list containing the unification of each of the subexpressions.  If they are not the same size or only one of them is a list, unification returns \texttt{false}.  If both expressions are primitives, unification returns true if they are equal, false otherwise.  In the case where the function expression of the unification is a variable, we return an assignment, i.e., the variable along with the expression passed to unification. If any subunifications have returned \texttt{false}, we return \texttt{false}. If a variable that repeatedly occurs in $F$ is assigned to different values in different places, we also return \texttt{false}.
Otherwise, unification succeeds and we return the assignment of each unique variable of $F$.

\begin{lstlisting}[frame=trbl]
(define unify
  ($\lbda$ (s sv vars)
    (begin
      (define (variable? obj)
        (member obj vars))

      (define (check/remove-repeated unified-vars)
        (let* ([repeated-vars (filter more-than-one (map (curry all-assoc unified-vars) (map first unified-vars)))])
          (if (and (all (map all-equal? repeated-vars)) (not (any false? unified-vars)))
              (delete-duplicates unified-vars)
              #f)))
      
      (cond [(variable? sv) (if (eq? s '$\lbda$) #f (list (pair sv s)))]
            [(and (primitive? s) (primitive? sv)) (if (eqv? s sv) '() #f)]
            [(or (primitive? s) (primitive? sv)) #f]
            [(not (eqv? (length s) (length sv))) #f]
            [else
             (let ([assignments (map ($\lbda$ (si sj) (unify si sj vars)) s sv)])
               (if (any false? assignments)
                   #f
                   (check/remove-repeated (apply append assignments))))]))))
\end{lstlisting}

We illustrate unification using the abstraction \texttt{(define (F1 V1 V2) (+ V1 V2))} and the expression \texttt{(+ (+ 2 2) (- 2 5))}.

\begin{enumerate}
  \item Since \texttt{(+ (+ 2 2) (- 2 5))} and \texttt{(+ V1 V2)} are of the same length, we apply unification to the subexpression pairs \texttt{[+,+], [(+ 2 2), V1], [(- 2 5), V2]}.
  \item Unification between \texttt{+} and \texttt{+} returns the empty assignment list since neither is a variable and they match.
  \item Unification between \texttt{(+ 2 2)} and \texttt{V1} returns the assignment of \texttt{(+ 2 2)} to \texttt{V1} and likewise for \texttt{(- 2 5)} and \texttt{V2}.
  \item It follows that the function \texttt{F1} matches the expression \texttt{(+ (+ 2 2) (- 2 5))} with variable assignments \texttt{V1:=(+ 2 2)} and \texttt{V2:=(- 2 5)}.
\end{enumerate}

If the expression had been \texttt{(- (+ 2 2) (- 2 5))}, then unification between the outer \texttt{-} of the expression and the \texttt{+} of \texttt{F1} would have returned \texttt{false} and unification would have failed.

\subsubsection{Summary}

Abstraction is a program transformation that identifies repeated computation in a program by finding syntactic patterns. If the program has been generated using data incorporation, syntactic patterns directly correspond to patterns in the observed data. While this formalization of the notion of a pattern may seem limited at first, it is worth contemplating the central role of lambda abstraction in the lambda calculus and the expressiveness of this language.

The abstraction process has two steps: First, create abstractions from common subexpressions in a program using anti-unification. Second, compress the program using these abstractions by replacing instances of the abstractions with function calls via unification.
It is important to note that a given program will usually have many possible abstraction transformations, corresponding to the different repeated patterns.

We now illustrate abstraction using the tree example that we first used in the section on data incorporation (figure \ref{fig:tree-obs}).
For this program, anti-unification finds $17$ possible abstractions. As examples, we show both the abstraction that results in the best compression (smallest program) and the abstraction that results in the 5th smallest program:
\begin{lstlisting}
(abstraction F1 (V1 V2)
             (data (color (gaussian V1 25)) (size V2)))

(abstraction F1 (V1 V2 V3 V4)
             (node (data (color (gaussian V1 25)) (size 0.3))
                   (node (data (color (gaussian V2 25)) (size 0.3)))
                   (node (data (color (gaussian V3 25)) (size 0.3)))
                   (node (data (color (gaussian V4 25)) (size 0.3)))))
\end{lstlisting}
The programs compressed using these abstractions look like this (size $55$ and $66$):
\begin{lstlisting}
(program
 ((abstraction F1 (V1 V2)
               (data (color (gaussian V1 25)) (size V2))))
 (uniform-choice
  (node (F1 70 1)
        (node (F1 37 0.3) (node (F1 213 0.3))
              (node (F1 207 0.3)) (node (F1 211 0.3))))
  (node (F1 43 1)
        (node (F1 47 0.1)
              (node (F1 33 0.3) (node (F1 220 0.3))
                    (node (F1 224 0.3)) (node (F1 207 0.3)))))))

(program
 ((abstraction F1 (V1 V2 V3 V4)
               (node (data (color (gaussian V1 25)) (size 0.3))
                     (node (data (color (gaussian V2 25)) (size 0.3)))
                     (node (data (color (gaussian V3 25)) (size 0.3)))
                     (node (data (color (gaussian V4 25)) (size 0.3))))))
 (uniform-choice
  (node (data (color (gaussian 70 25)) (size 0.7))
        (F1 37 213 207 211))
  (node (data (color (gaussian 43 25)) (size 0.7))
        (node
         (data (color (gaussian 47 25)) (size 0.1))
         (F1 33 220 224 207)))))
\end{lstlisting}
The second abstraction corresponds to a ``flower''-like pattern. In this pattern, we could capture even more structure by replacing the variables for the petal colors with a fixed value, since they are similar. Instead of explaining the data as drawn from multiple Gaussians with slightly different means, we could explain the data as generated by a single Gaussian. Our second program transformation, deargumentation, addresses this issue.
\subsection{Deargumentation}
Deargumentation is a program transformation that takes a function $F$ in a program and changes its definition by removing one of the function arguments. Wherever this removed argument is used within $F$, we instead have an independent sample from a new constant or distribution. This new value depends on the values of the original argument in the overall program; depending on how we map these argument values to replacements (specified via \texttt{replacement-function}), we create different program transformations.

\begin{lstlisting}[frame=trbl]
(define (deargument replacement-function program abstraction variable)
  (let* ([abstraction* (remove-abstraction-variable replacement-function program abstraction variable)])
    (if (null? abstraction*)
        '()
        (let* ([program+abstraction* (update-abstraction program abstraction*)]
               [program* (remove-application-argument program+abstraction* abstraction variable)])
          program*))))
\end{lstlisting}
The abstraction whose variable is being removed keeps the same body, but the variable removed is now assigned a value within the body instead of having its value passed in as an argument.
\begin{lstlisting}[frame=trbl]
(define (remove-abstraction-variable replacement-function program abstraction variable)
  (let* ([variable-instances (find-variable-instances program abstraction variable)]
         [variable-definition (replacement-function program abstraction variable variable-instances)])
    (if (equal? variable-definition NO-REPLACEMENT)
        '()
        (let* ([new-body `(($\lbda$ (,variable) ,(abstraction->body abstraction)) ,variable-definition)]
               [new-variables (delete variable (abstraction->vars abstraction))])
          (make-named-abstraction (abstraction->name abstraction) new-body new-variables)))))

(define (program->abstraction-applications program target-abstraction)
  (define (target-abstraction-application? sexpr)
    (if (non-empty-list? sexpr)
        (if (equal? (first sexpr) (abstraction->name target-abstraction))
            #t
            #f)
        #f))
  (let* ([abstraction-bodies (map abstraction->body (program->abstractions program))]
         [possible-locations (pair (program->body program) abstraction-bodies)])
    (deep-find-all target-abstraction-application? possible-locations)))

(define (deep-find-all pred? sexp)
  (filter pred? (all-subexprs sexp)))

(define (all-subexprs t)
  (let loop ([t (list t)])
    (cond [(null? t) '()]
          [(primitive? (first t)) (loop (rest t))]
          [else (pair (first t) (loop (append (first t) (rest t))))])))

(define (find-variable-instances program abstraction variable)
  (let* ([abstraction-applications (program->abstraction-applications program abstraction)]
         [variable-position (abstraction->variable-position abstraction variable)]
         [variable-instances (map (curry ith-argument variable-position) abstraction-applications)])
    variable-instances))

(define (ith-argument i function-application)
  (list-ref function-application (+ i 1)))
\end{lstlisting}
After the abstraction for function $F$ has been adjusted, we change all applications of $F$ by removing the appropriate argument.
\begin{lstlisting}[frame=trbl]
(define (remove-application-argument program abstraction variable)
  (define (abstraction-application? sexpr)
    (if (non-empty-list? sexpr)
        (equal? (first sexpr) (abstraction->name abstraction))
        #f))
  (define (change-application variable-position application)
    (define (change-recursive-arguments argument)
      (if (abstraction-application? argument)
          (change-application variable-position argument)
          argument))
    (let* ([ith-removed (remove-ith-argument variable-position application)])
      (map change-recursive-arguments ith-removed)))
  (let* ([variable-position (abstraction->variable-position abstraction variable)]
         [program-sexpr (program->sexpr program)]
         [changed-sexpr (transform-sexp abstraction-application? (curry change-application variable-position) program-sexpr)]
         [program* (sexpr->program changed-sexpr)])
    program*))

(define (remove-ith-argument i function-application)
  (append (take function-application (+ i 1)) (drop function-application (+ i 2))))

(define (transform-sexp pred? func sexp)
  (if (pred? sexp)
      (func sexp)
      (if (list? sexp)
          (map (curry transform-sexp pred? func) sexp)
          sexp)))
\end{lstlisting}
In the following, we demonstrate that this transformation is useful for compactly representing continuous values that may have been distorted by noise, for identifying replicated arguments, and for inducing recursive structure.

\subsubsection{Compactly representing noisy data}

When we unify two expressions to create abstractions, all places where the expressions do not exactly match result in the creation of a variable. For example, given \texttt{(+ 2 1.99)} and \texttt{(+ 2 2.01)}, we unify to \texttt{(+ 2 V)}. If we know that the system we are modeling is noisy, we may want to treat these two expressions as essentially identical, i.e., unifying to \texttt{(+ 2 2)} instead. We achieve this effect by using the deargumentation transform with a \texttt{replacement-function} called \texttt{noisy-number-replacement}. This function replaces a variable within an abstraction with the mean of the values of all its instances.

\begin{lstlisting}[frame=trbl]
(define (noisy-number-replacement program abstraction variable variable-instances)
  (if (all (map number? variable-instances))
      (mean variable-instances)
      NO-REPLACEMENT))
\end{lstlisting}

\begin{figure}
  \subfigure[Samples from the original program.]{
    \includegraphics[scale=.26]{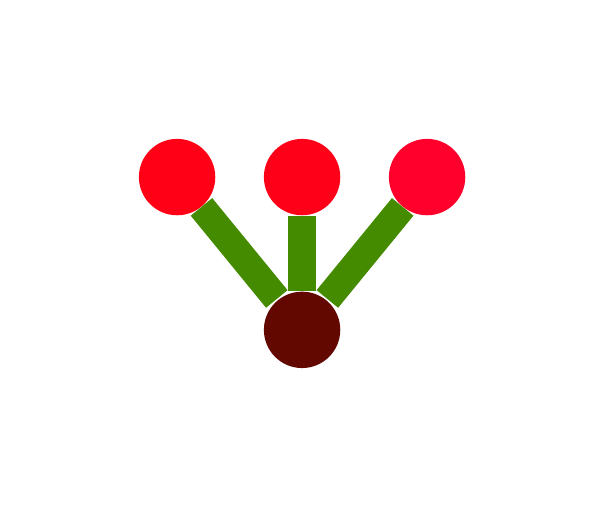}
    \includegraphics[scale=.26]{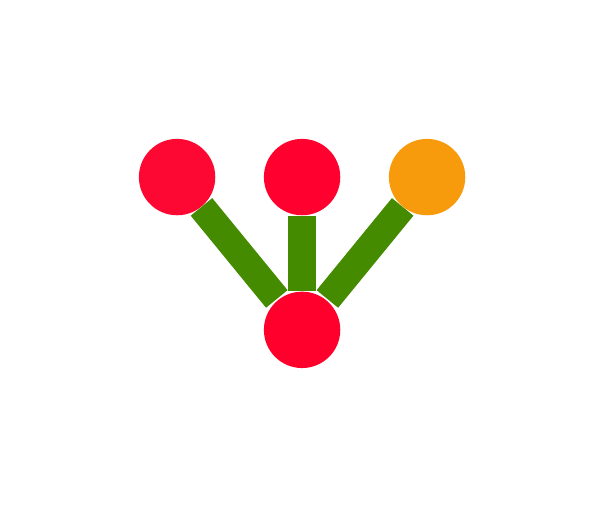}
    \includegraphics[scale=.26]{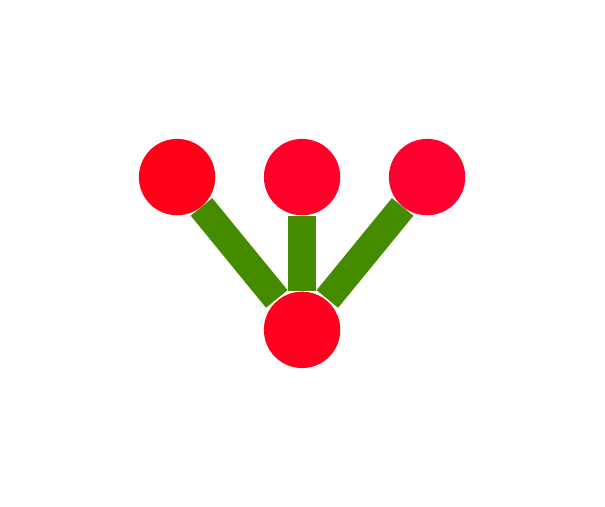}
    \includegraphics[scale=.26]{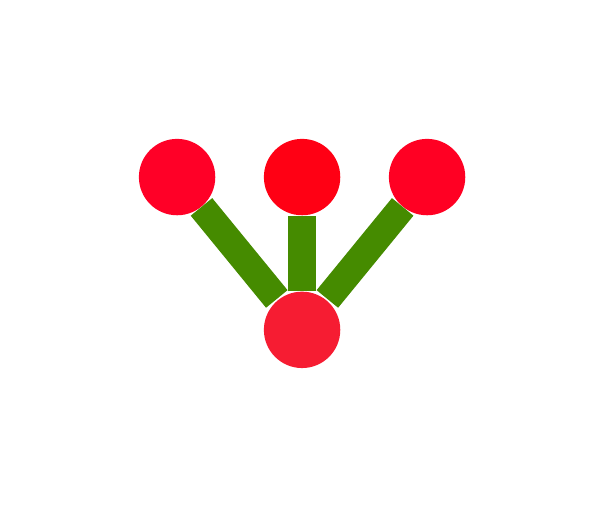}
    \includegraphics[scale=.26]{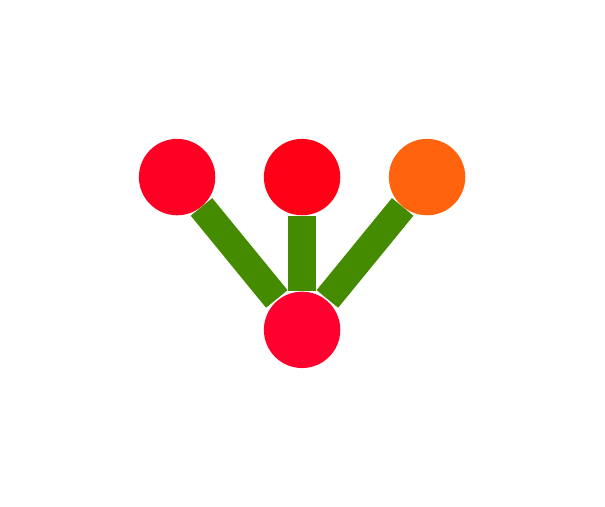}
    \includegraphics[scale=.26]{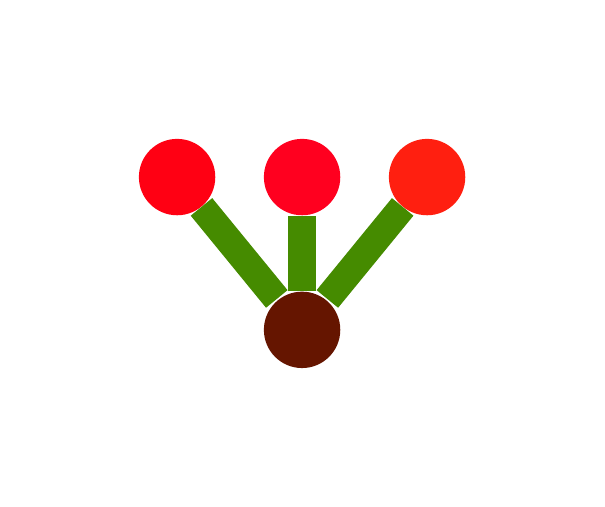}
    \includegraphics[scale=.26]{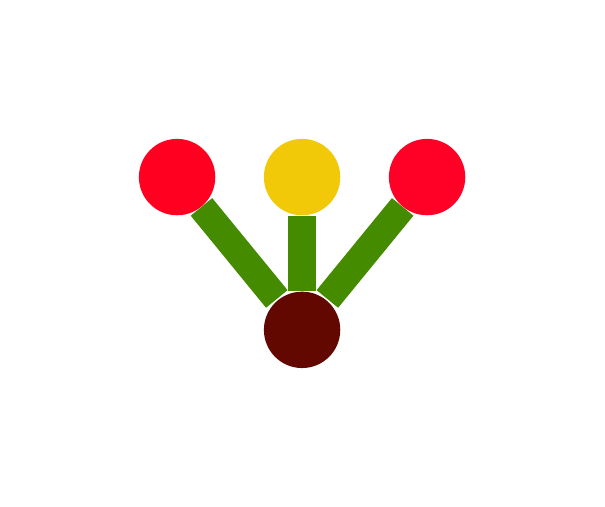}
    \includegraphics[scale=.26]{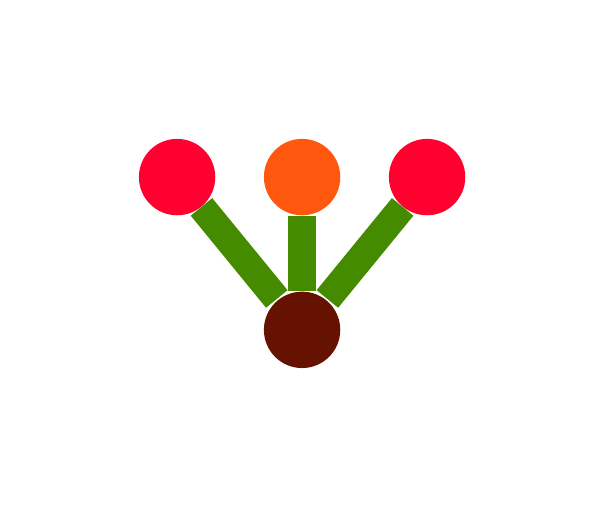}
    \includegraphics[scale=.26]{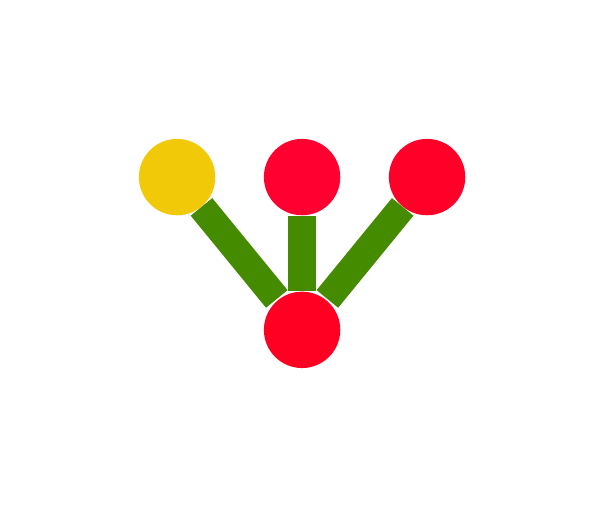}
    \label{fig:noisy-orig}
    }
  \subfigure[An argument with values of uniform color has been replaced with a stochastic choice.]{
    \includegraphics[scale=.26]{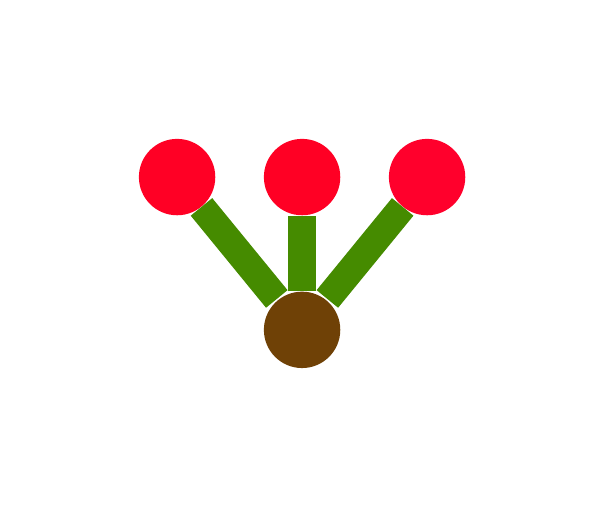}
    \includegraphics[scale=.26]{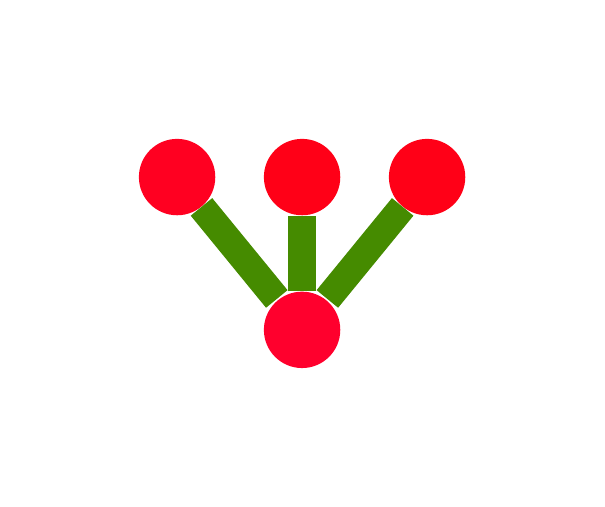}
    \includegraphics[scale=.26]{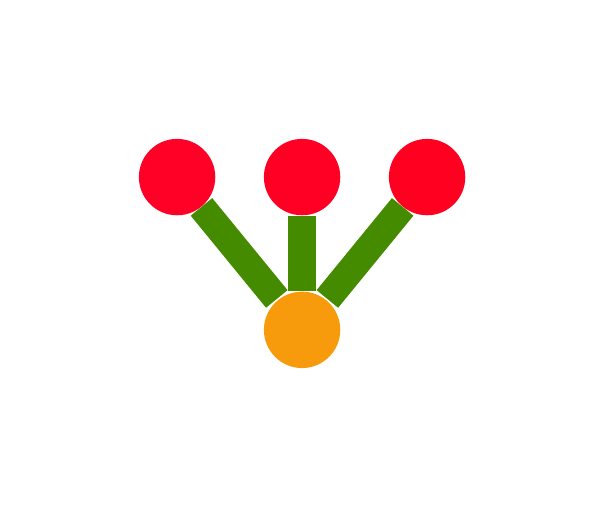}
    \includegraphics[scale=.26]{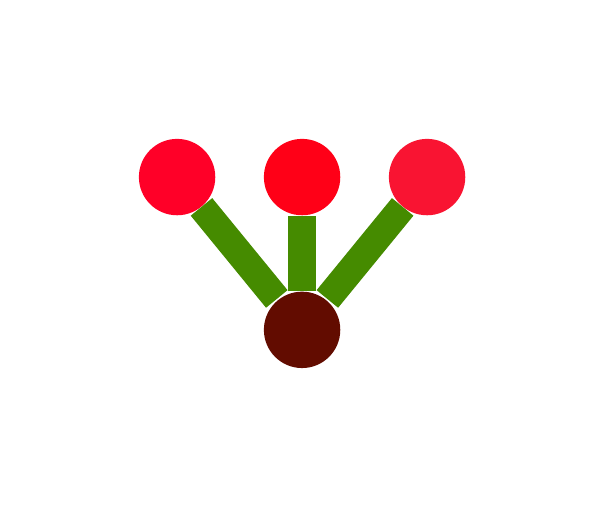}
    \includegraphics[scale=.26]{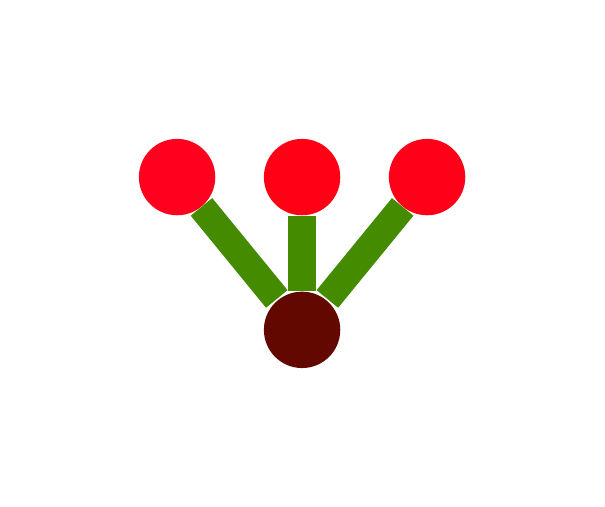}
    \includegraphics[scale=.26]{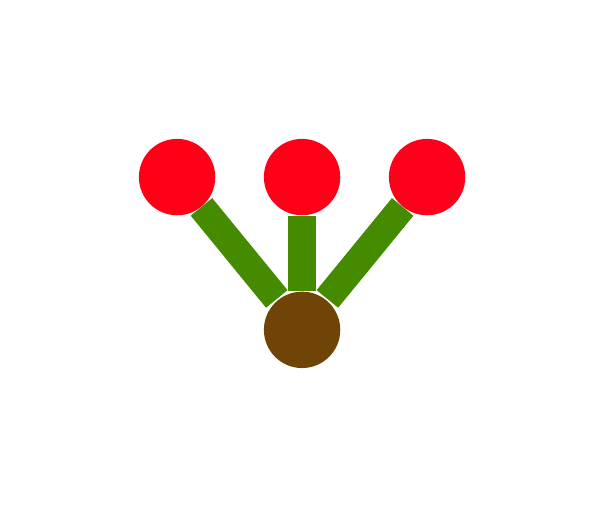}
    \includegraphics[scale=.26]{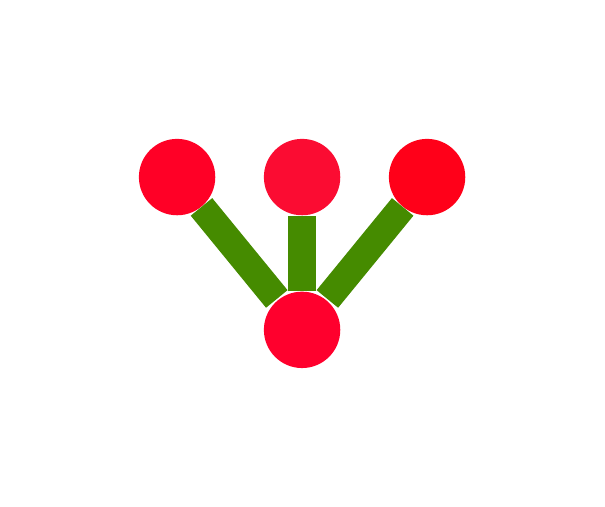}
    \includegraphics[scale=.26]{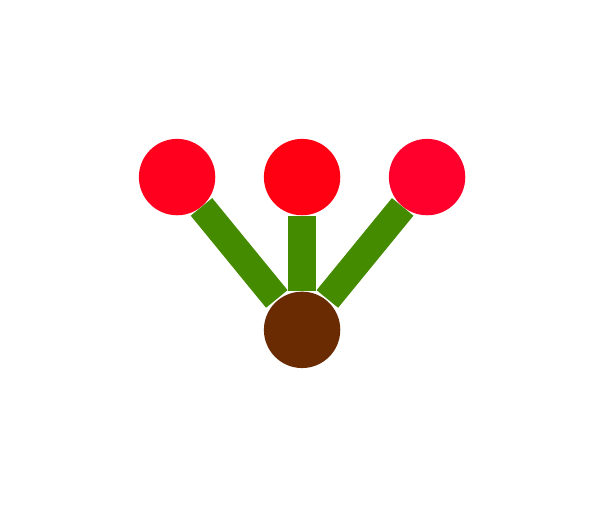}
    \includegraphics[scale=.26]{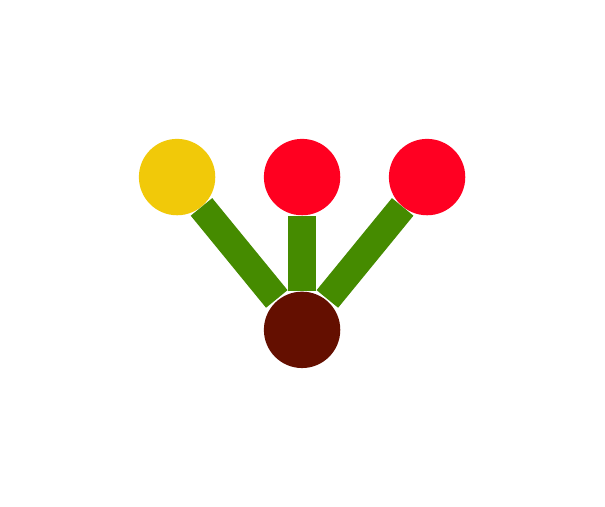}
    \label{fig:noisy-good}    
    }
  \subfigure[An argument with non-uniform values has been replaced with a stochastic choice.]{
    \includegraphics[scale=.26]{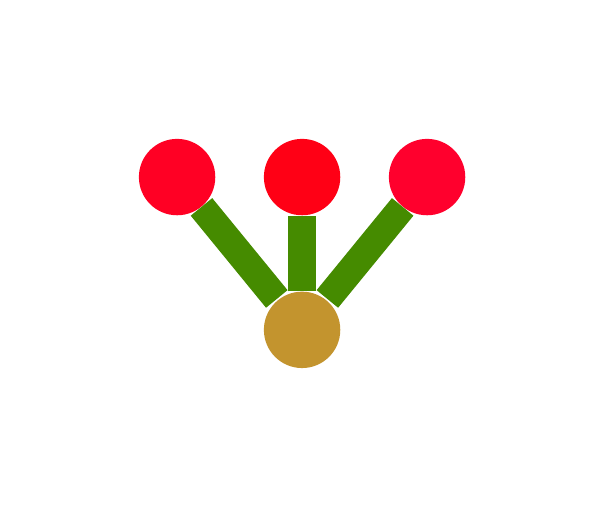}
    \includegraphics[scale=.26]{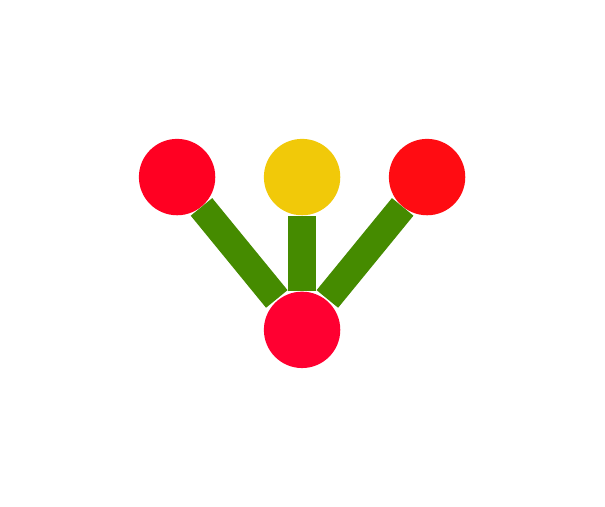}
    \includegraphics[scale=.26]{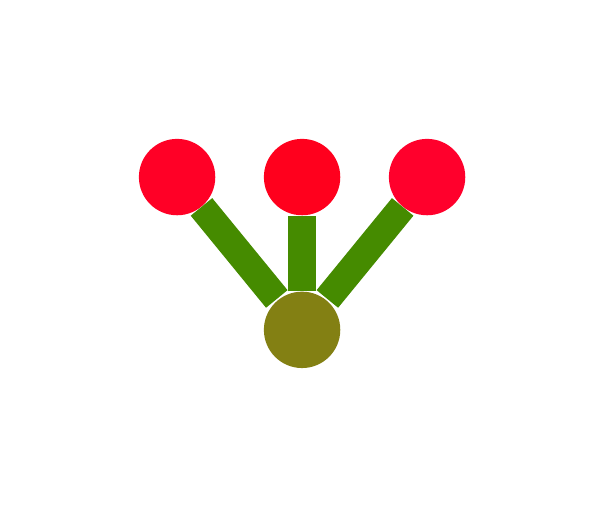}
    \includegraphics[scale=.26]{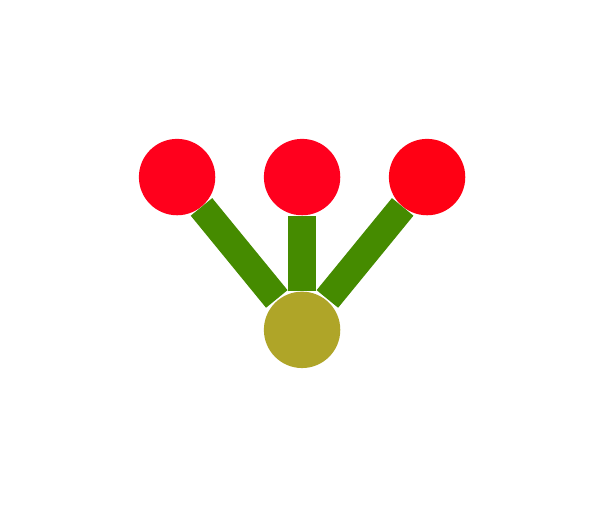}
    \includegraphics[scale=.26]{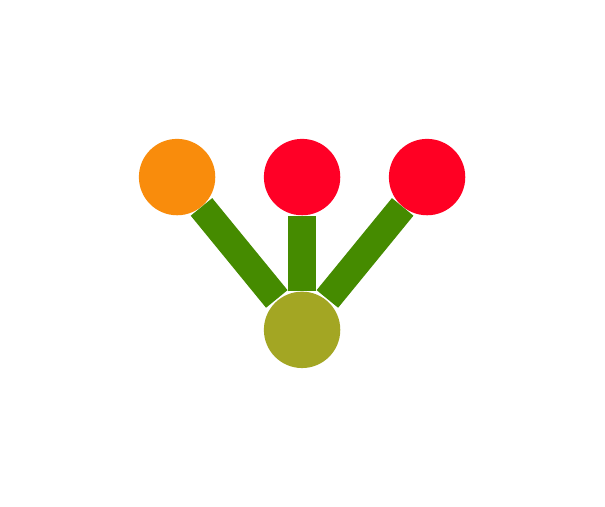}
    \includegraphics[scale=.26]{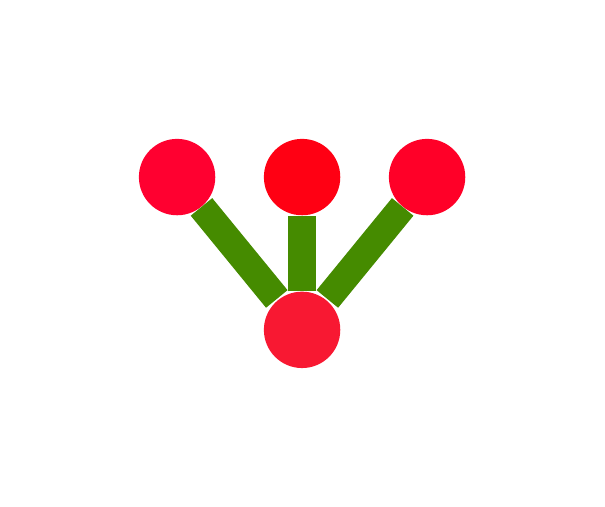}
    \includegraphics[scale=.26]{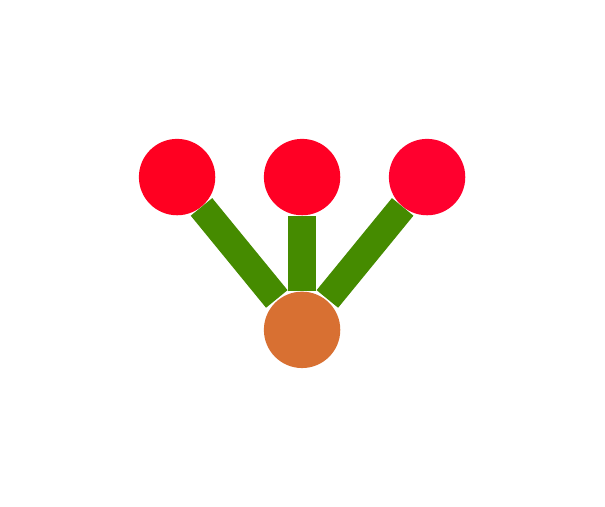}
    \includegraphics[scale=.26]{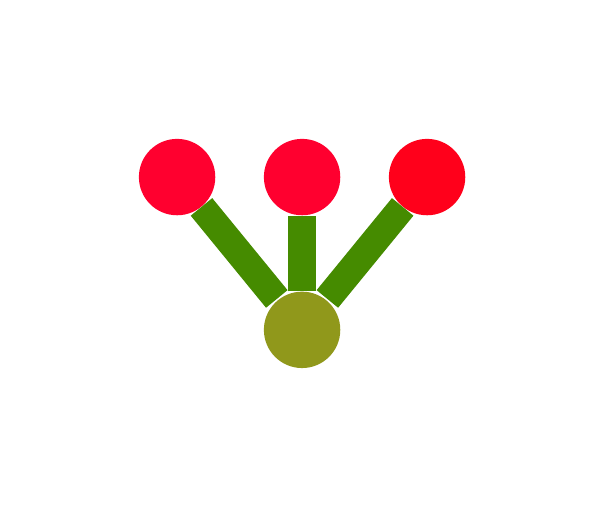}
    \includegraphics[scale=.26]{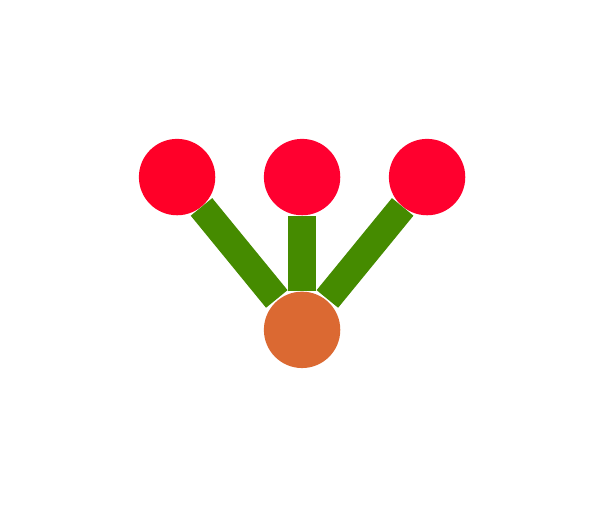}
    \label{fig:noisy-bad}        
    }
  \caption{{\bf Replacing arguments with stochastic choices.} When an argument takes on similar values in all its instantiations, replacing it with a stochastic choice can result in samples almost indistinguishable from the original program (b). When values are dissimilar, using a Gaussian centered on the mean of the values will result in samples that look unlike the samples of the original program (c).}
\end{figure}

Suppose that abstraction results in this program (samples shown in figure \ref{fig:noisy-orig}):
\begin{lstlisting}[mathescape=true]
(begin
  (define flower
    ($\lbda$ (V1 V2 V3 V4)
      (node (data (color (gaussian V1 25)) (size 0.3))
            (node (data (color (gaussian V2 25)) (size 0.3)))
            (node (data (color (gaussian V3 25)) (size 0.3)))
            (node (data (color (gaussian V4 25)) (size 0.3))))))
  (uniform-choice
   (flower 200 213 207 211)
   (flower 33 220 224 207)))
\end{lstlisting}
If we apply the deargumentation transform to the \texttt{flower} function and variable \texttt{V2}, we identify the two instances of \texttt{V2}, $213$ and $220$, and we are going to replace this variable by its mean, $216.5$.  The abstraction \texttt{flower} is then changed using \texttt{remove-abstraction-variable} to:
\begin{lstlisting}
(define flower
  ($\lbda$ (V1 V3 V4)
    (($\lbda$ (V2)
       (node (data (color (gaussian V1 25)) (size 0.3))
             (node (data (color (gaussian V2 25)) (size 0.3)))
             (node (data (color (gaussian V3 25)) (size 0.3)))
             (node (data (color (gaussian V4 25)) (size 0.3)))))
     216.5)))
\end{lstlisting}
The program is now adjusted to incorporate the new version of \texttt{flower} using \texttt{remove-application-argument} (samples in figure \ref{fig:noisy-good}):
\begin{lstlisting}[mathescape=true]
(begin
  (define flower
    ($\lbda$ (V1 V3 V4)
      (($\lbda$ (V2)
         (node (data (color (gaussian V1 25)) (size 0.3))
               (node (data (color (gaussian V2 25)) (size 0.3)))
               (node (data (color (gaussian V3 25)) (size 0.3)))
               (node (data (color (gaussian V4 25)) (size 0.3)))))
       216.5)))
  (uniform-choice
   (flower 200 207 211)
   (flower 33 224 207)))
\end{lstlisting}
The more applications of \texttt{flower} exist within the program, the bigger the impact of this simplification. This transform will affect the likelihood of the data under the program. Since we search based on posterior probability, incorporating both prior and likelihood, we trade off improvements in model complexity and fit to the data in a principled way. If we had applied the deargumentation transform to the \texttt{flower} argument \texttt{V1}, which is used with values that are less similar, we would get a program that gives lower likelihood to the data than the program generated by removing \texttt{V2} (samples in figure \ref{fig:noisy-bad}):
\begin{lstlisting}[mathescape=true]
(begin
  (define flower
    ($\lbda$ (V2 V3 V4)
      (($\lbda$ (V1)
         (node (data (color (gaussian V1 25)) (size 0.3))
               (node (data (color (gaussian V2 25)) (size 0.3)))
               (node (data (color (gaussian V3 25)) (size 0.3)))
               (node (data (color (gaussian V4 25)) (size 0.3)))))
       116.5)))
  (uniform-choice
   (flower 213 207 211)
   (flower 220 224 207)))
\end{lstlisting}

\subsubsection{Merging similar variables}
Abstraction creates functions whose arguments represent expression parts that differ between specific abstraction instances.  Arguments that take on similar values in each of their instantiations might have the same underlying generative process.  We take advantage of this potential redundancy by merging these arguments.  This is achieved by using the deargumentation transform with the following \texttt{replacement-function}:
\begin{lstlisting}[frame=trbl]
(define (same-variable-replacement program abstraction variable variable-instances)
  (let ([possible-match-variables (delete variable (abstraction->vars abstraction))])
    (find-matching-variable program abstraction variable-instances possible-match-variables)))

(define (find-matching-variable program abstraction variable-instances possible-match-variables)
  (define (equal? a b) 
    (if (and (pair? a) (pair? b)) 
        (and (equal? (car a) (car b)) (equal? (cdr a) (cdr b))) 
        (if (and (number? a) (number? b))
            #t
            (eq? a b)))) 
  (if (null? possible-match-variables)
      NO-REPLACEMENT
      (let* ([hypothesis-variable (uniform-draw possible-match-variables)]
             [hypothesis-instances (find-variable-instances program abstraction hypothesis-variable)])
        (if (equal? hypothesis-instances variable-instances)
            hypothesis-variable
            (find-matching-variable program abstraction variable-instances (delete hypothesis-variable possible-match-variables))))))
\end{lstlisting}

If two variables take on similar values in all of their applications, we consider them to be the same.  Above, we define two items to be similar if they have the same list structure or if they are both numbers.  We still consider two objects as similar even if the numbers have different values and allow the resulting likelihood to filter out models that do not explain the data.

To detect matching variables, we first choose an abstraction, choose a variable for this abstraction, and then check whether any of the other variables of this abstraction have a matching set of instance values. If a variable matches, we return it as the new definition of the variable that is being removed\footnote{Alternatively, we could directly replace the variable name in the abstraction body and get slightly better compression. We chose to go the redefinition route in order to keep the interface within the class of deargumentation transforms.}.

\begin{figure}
  \subfigure[Samples from the original program.]{
    \includegraphics[scale=.26]{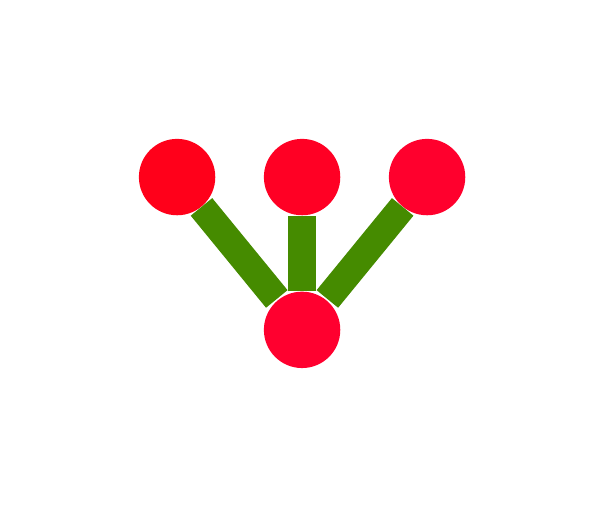}
    \includegraphics[scale=.26]{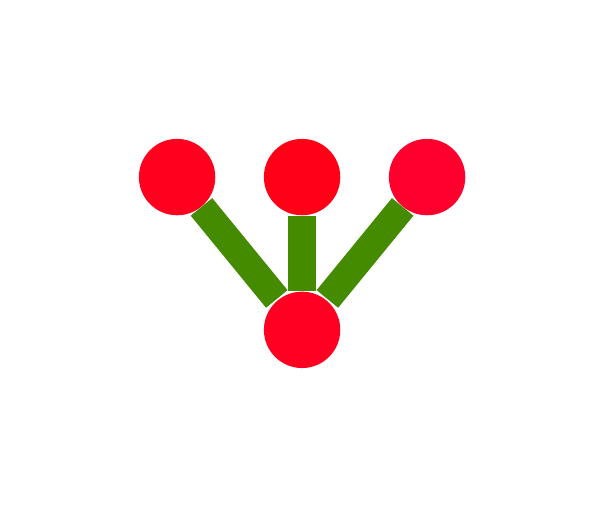}
    \includegraphics[scale=.26]{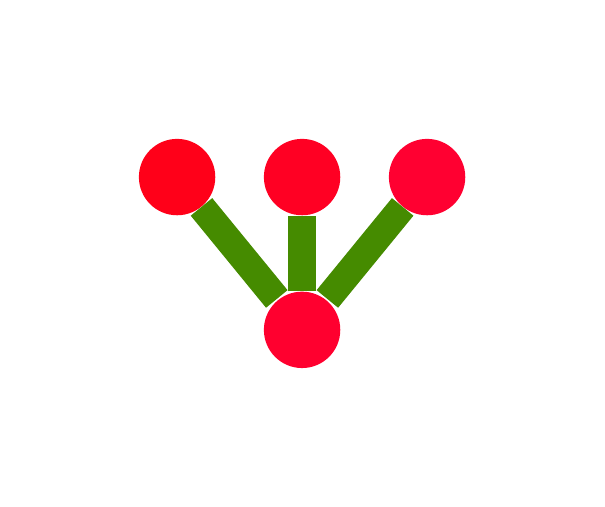}
    \includegraphics[scale=.26]{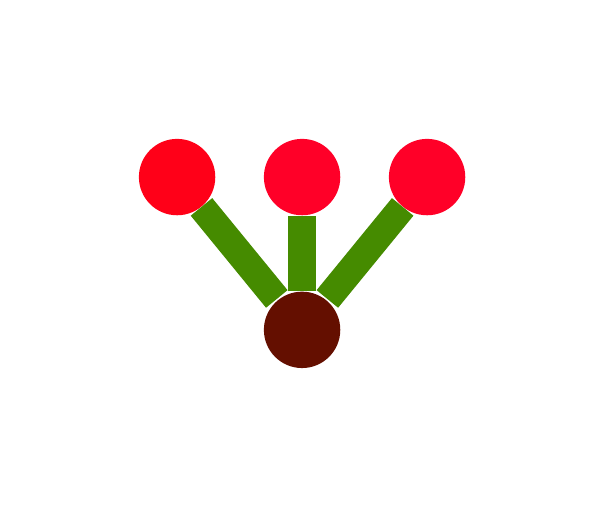}
    \includegraphics[scale=.26]{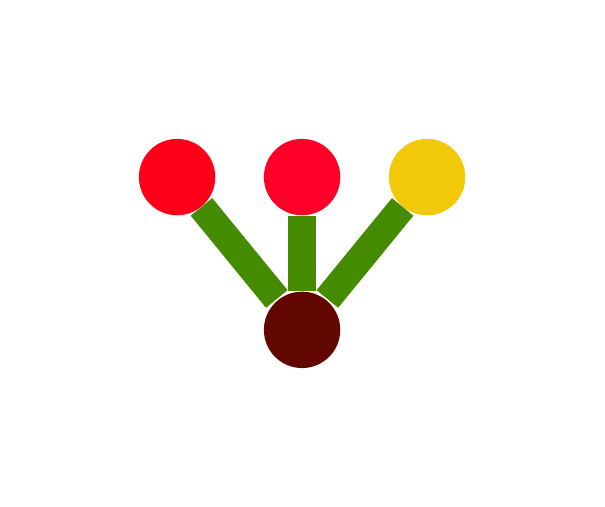}
    \includegraphics[scale=.26]{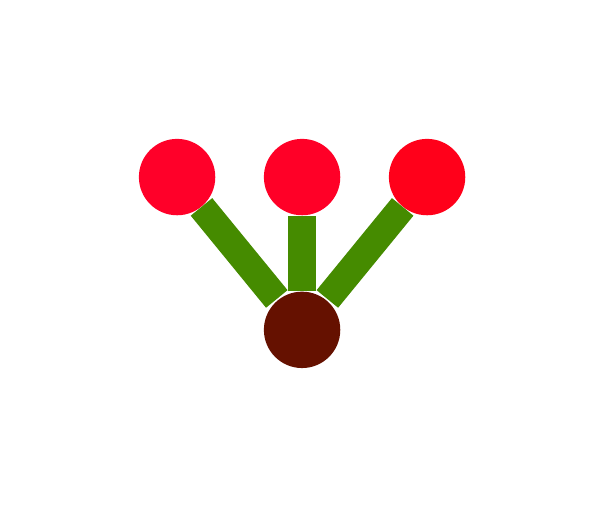}
    \includegraphics[scale=.26]{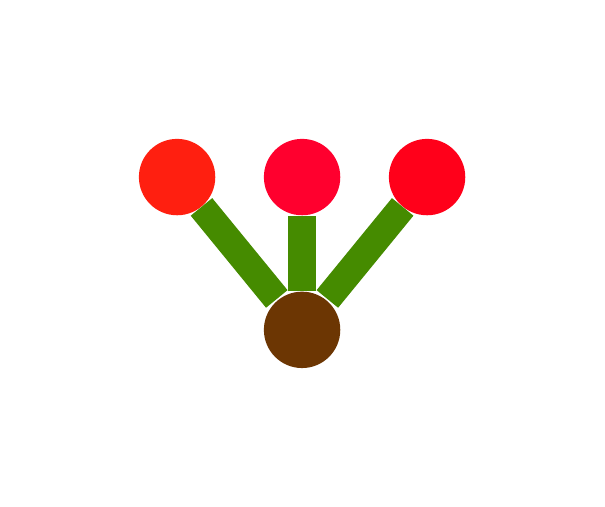}
    \includegraphics[scale=.26]{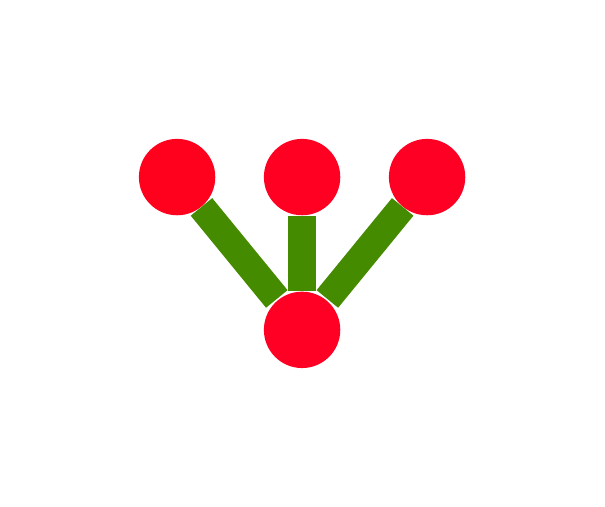}
    \includegraphics[scale=.26]{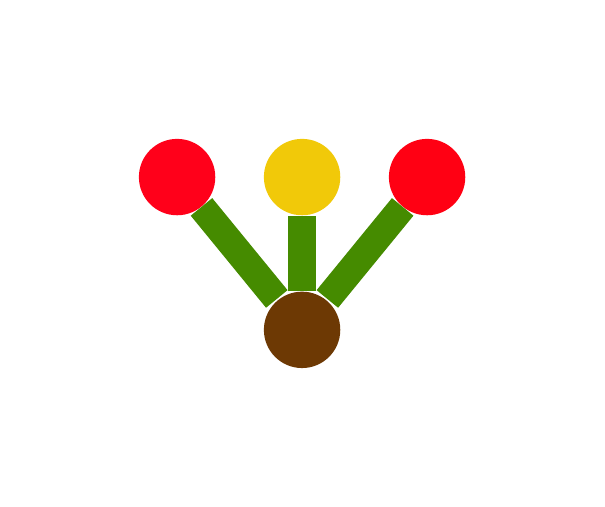}
    \label{fig:merge-orig}
    }
  \subfigure[Two arguments with similar instantiations have been merged.]{
    \includegraphics[scale=.26]{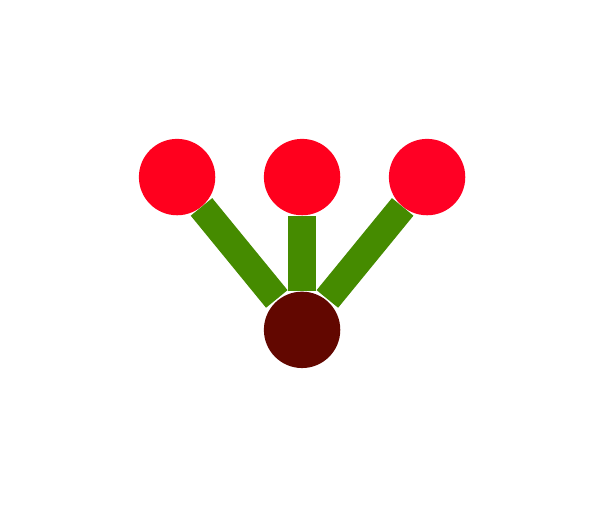}
    \includegraphics[scale=.26]{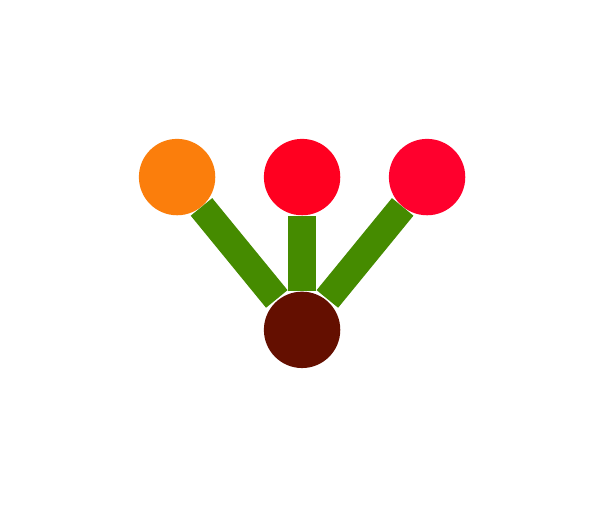}
    \includegraphics[scale=.26]{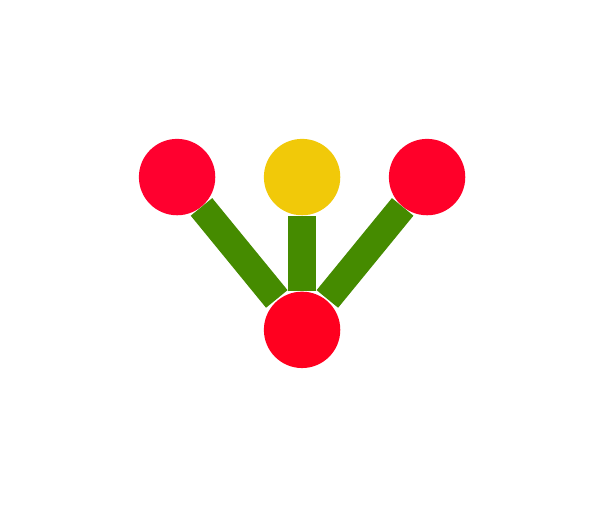}
    \includegraphics[scale=.26]{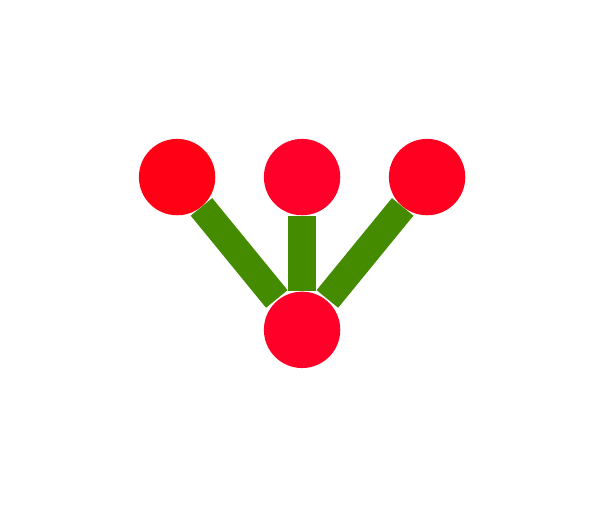}
    \includegraphics[scale=.26]{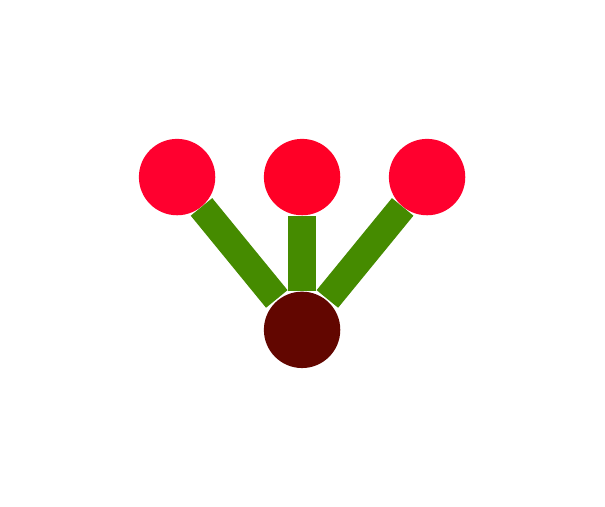}
    \includegraphics[scale=.26]{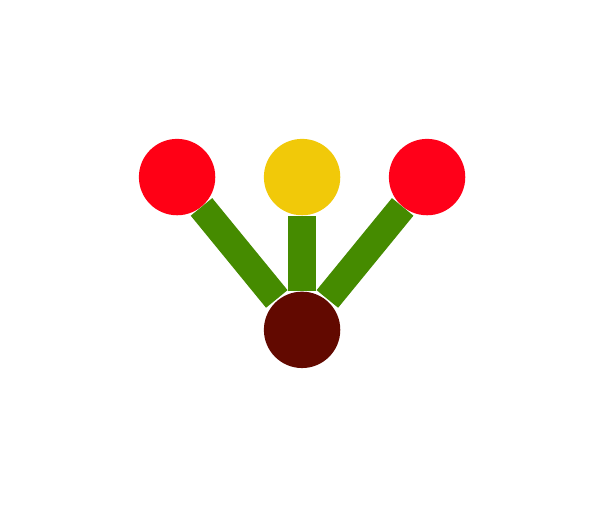}
    \includegraphics[scale=.26]{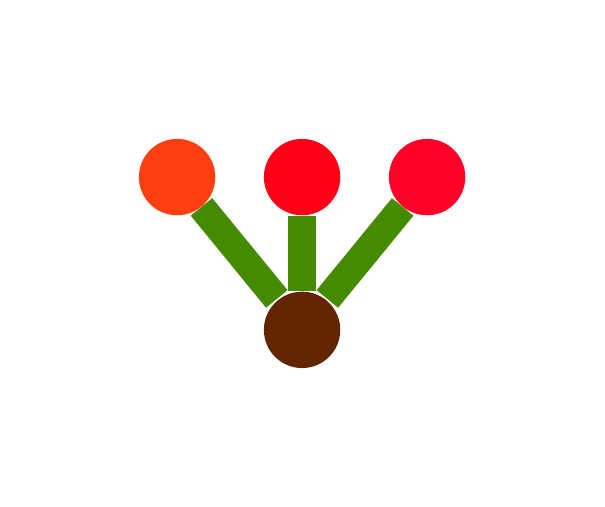}
    \includegraphics[scale=.26]{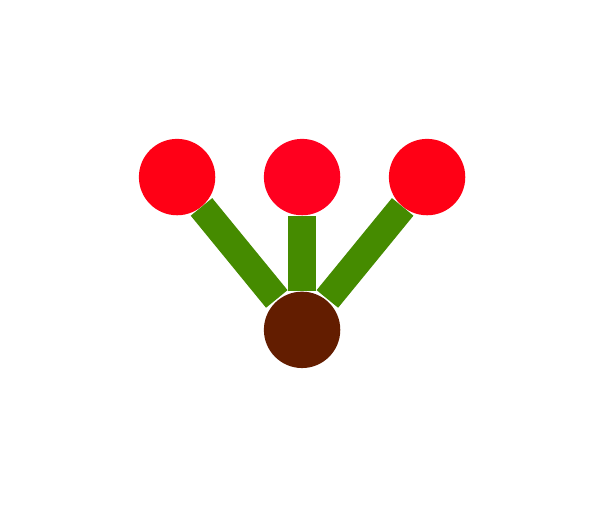}
    \includegraphics[scale=.26]{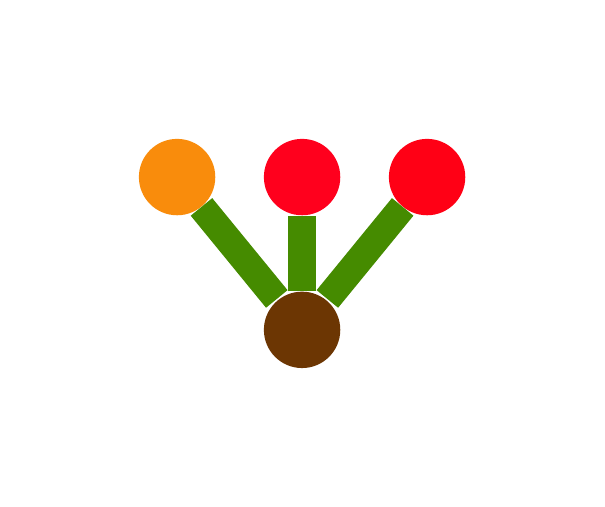}
    \label{fig:merge-good}    
    }
  \subfigure[Two arguments with dissimilar instantiations have been merged.]{
    \includegraphics[scale=.26]{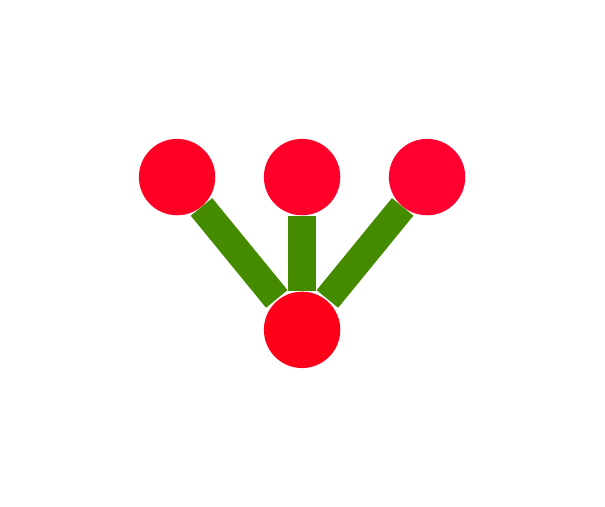}
    \includegraphics[scale=.26]{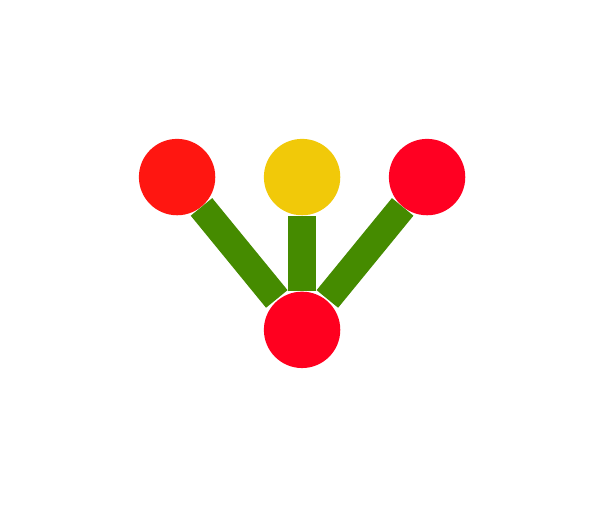}
    \includegraphics[scale=.26]{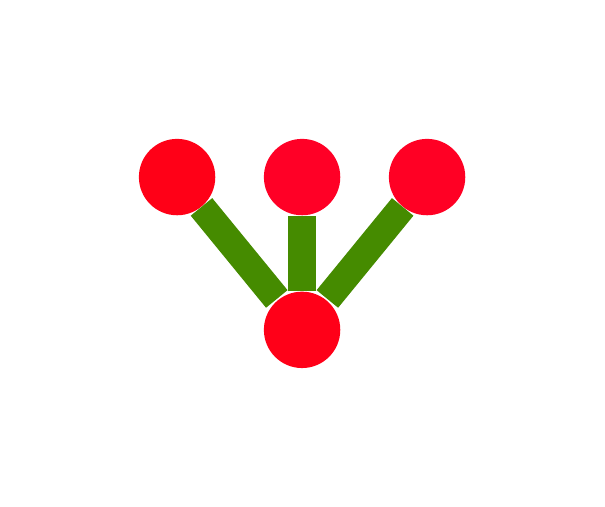}
    \includegraphics[scale=.26]{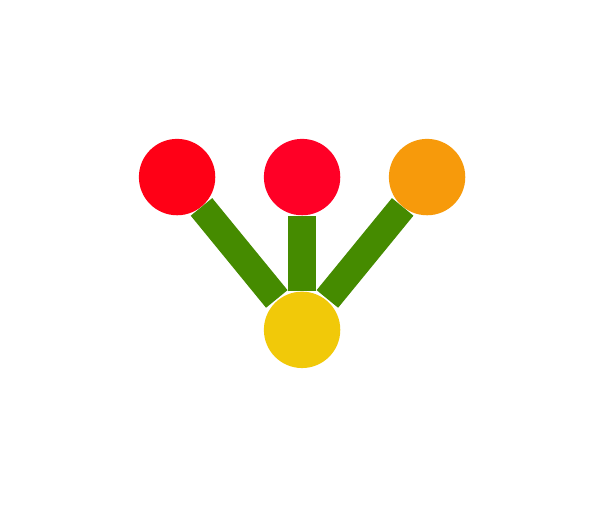}
    \includegraphics[scale=.26]{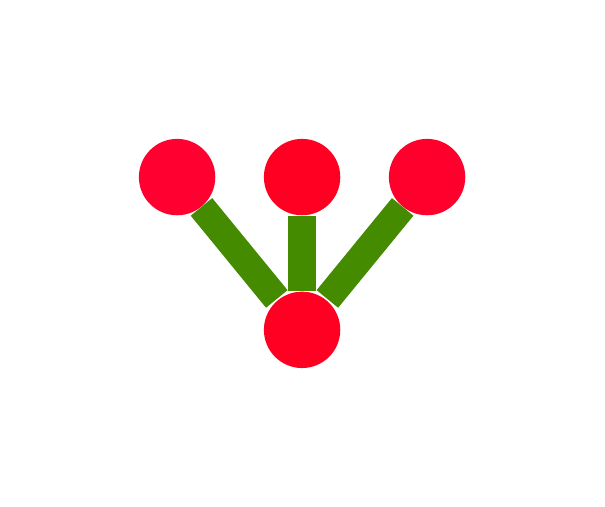}
    \includegraphics[scale=.26]{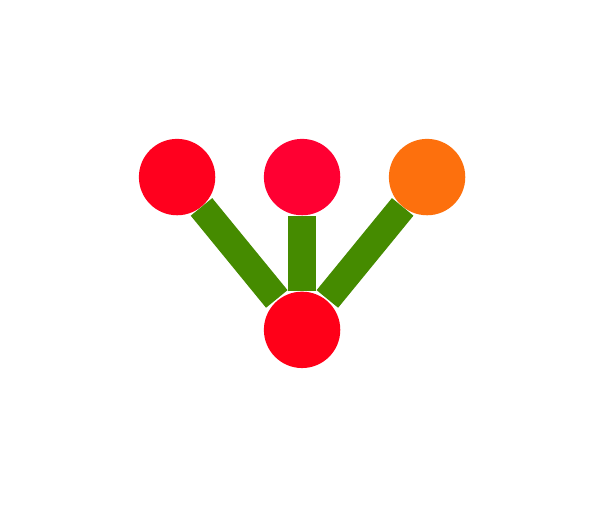}
    \includegraphics[scale=.26]{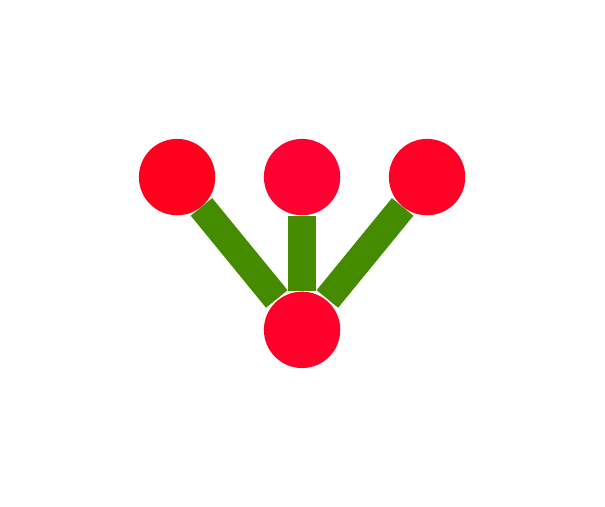}
    \includegraphics[scale=.26]{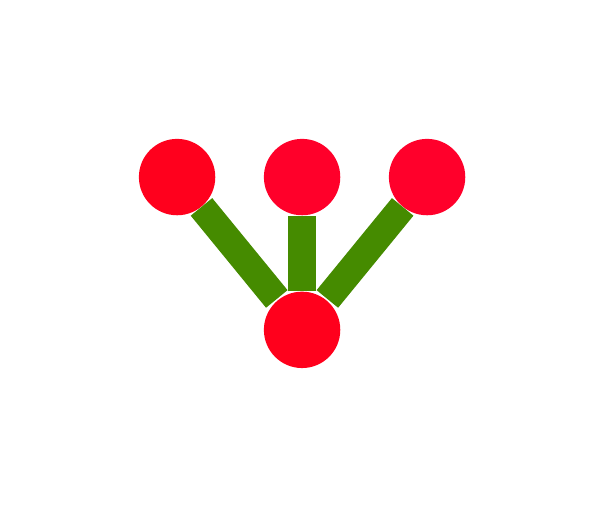}
    \includegraphics[scale=.26]{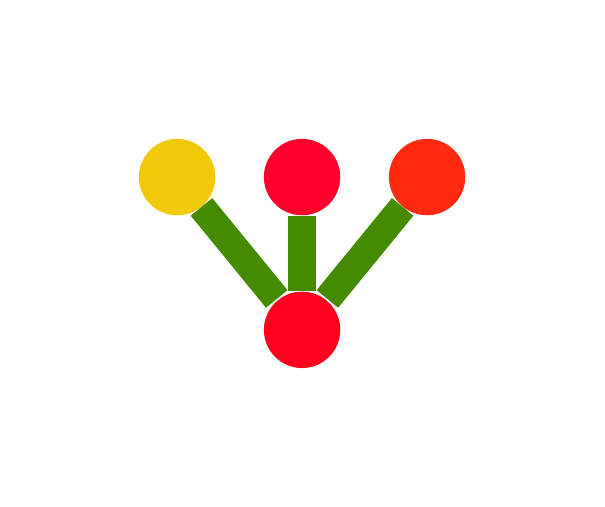}
    \label{fig:merge-bad}        
    }
  \caption{{\bf Merging arguments.} When two arguments take on similar values in all instantiations, merging them into one argument results in samples almost indistinguishable from the original program (b). When instantiations are dissimilar, the samples look unlike the samples of the original program (c).}
\end{figure}

Suppose again that we start with the following program:
\begin{lstlisting}[mathescape=true]
(begin
  (define flower
    ($\lbda$ (V1 V2 V3 V4)
      (node (data (color (gaussian V1 25)) (size 0.3))
            (node (data (color (gaussian V2 25)) (size 0.3)))
            (node (data (color (gaussian V3 25)) (size 0.3)))
            (node (data (color (gaussian V4 25)) (size 0.3))))))
  (uniform-choice
   (flower 200 213 207 211)
   (flower 33 220 224 207))
\end{lstlisting}

If we apply the deargumentation transform with \texttt{same-variable-replacement} to variable \texttt{V2}, we get a program that generates data similar to the original as long as \texttt{V2} is matched to a variable that is a ``petal'', i.e., \texttt{V3} (shown here) or \texttt{V4}:

\begin{lstlisting}[mathescape=true]
(begin
  (define flower
    ($\lbda$ (V1 V3 V4)
      (($\lbda$ (V2)
         (node (data (color (gaussian V1 25)) (size 0.3))
               (node (data (color (gaussian V2 25)) (size 0.3)))
               (node (data (color (gaussian V3 25)) (size 0.3)))
               (node (data (color (gaussian V4 25)) (size 0.3)))))
       V3)))
  (uniform-choice
   (flower 200 207 211)
   (flower 33 224 207)))
\end{lstlisting}

If the variable does not match a petal---which is only the case for \texttt{V1}---we get a program that generates data that is unlike the data generated by the original program:

\begin{lstlisting}[mathescape=true]
(begin
  (define flower
    ($\lbda$ (V2 V3 V4)
      (($\lbda$ (V1)
         (node (data (color (gaussian V1 25)) (size 0.3))
               (node (data (color (gaussian V2 25)) (size 0.3)))
               (node (data (color (gaussian V3 25)) (size 0.3)))
               (node (data (color (gaussian V4 25)) (size 0.3)))))
       V2)))
  (uniform-choice
   (flower 213 207 211)
   (flower 220 224 207)))
\end{lstlisting}

\subsubsection{Inducing recursive functions}

We now illustrate a deargumentation transform that discovers and compactly represents recursive patterns. The \texttt{replacement-function} for this transform, \texttt{recursion-replacement}, redefines a variable as a random choice between a recursive call and a non-recursive call. This redefinition is useful whenever the argument values that are passed to a function $F$ are calls to $F$ itself. The probability of recursion depends on how many argument instances are in fact   calls to the same function. Every program generated in this way will terminate with probability one. While this approach to inducing recursive programs is limited, we view it as an interesting illustration of how recursion can be identified and captured.
\begin{lstlisting}[frame=trbl]
(define (recursion-replacement program abstraction variable variable-instances)
  (let* ([valid-variable-instances (remove has-variable? variable-instances)]
         [recursive-calls (filter (curry abstraction-application? abstraction) valid-variable-instances)]
         [non-recursive-calls (remove (curry abstraction-application? abstraction) valid-variable-instances)]
         [terminates (terminates? program (abstraction->name abstraction) non-recursive-calls)]) 
    (if (or (null? valid-variable-instances) (null? recursive-calls) (not terminates))
        NO-REPLACEMENT
        (let* ([prob-of-recursion (/ (length recursive-calls) (length valid-variable-instances))])
          `(if (flip ,prob-of-recursion) ,(first recursive-calls) (uniform-choice  ,@non-recursive-calls))))))

(define (terminates? program init-abstraction-name non-recursive-calls)
  (define abstraction-statuses (make-hash-table eq?))

  (define (initialize-statuses)
    (let ([abstraction-names (map abstraction->name (program->abstractions program))])
      (begin
        (map (curry hash-table-set! abstraction-statuses) abstraction-names (make-list (length abstraction-names) 'unchecked))
        (hash-table-set! abstraction-statuses init-abstraction-name 'checking))))

  (define (status? name)
    (hash-table-ref abstraction-statuses name))

  (define (set-status! name new-status)
    (hash-table-set! abstraction-statuses name new-status))
  
  (define (terminating-abstraction? abstraction-name)
    (cond [(eq? (status? abstraction-name) 'terminates) #t]
          [(eq? (status? abstraction-name) 'checking) #f]
          [(eq? (status? abstraction-name) 'unchecked)
           (begin
             (set-status! abstraction-name 'checking)
             (if (base-case? (program->abstraction-body program abstraction-name))
                 (begin
                   (set-status! abstraction-name 'terminates)
                   #t)
                 #f))]))
  
  (define (base-case? sexpr)
    (cond [(branching? sexpr) (list-or (map base-case?  (get-branches sexpr)))]
          [(application? sexpr) (if (any-abstraction-application? sexpr)
                                    (and (terminating-abstraction? (operator sexpr)) (all (map base-case? (operands sexpr))))
                                    (all (map base-case? (operands sexpr))))]
          [else #t]))
  (begin
    (initialize-statuses)
    (list-or (map base-case? non-recursive-calls))))
\end{lstlisting}

\begin{figure}
  \subfigure[Samples from the original program.]{
    \includegraphics[scale=.2]{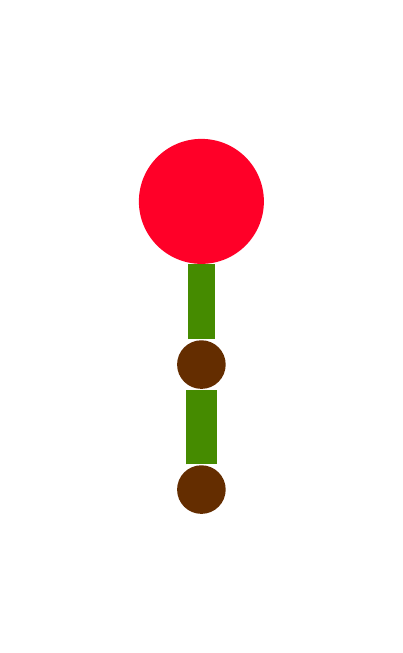}
    \includegraphics[scale=.2]{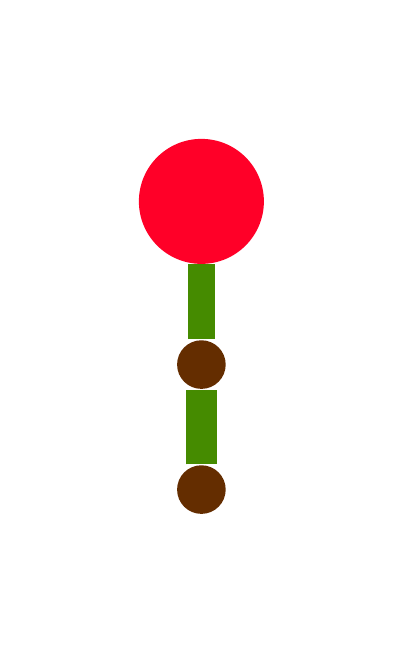}
    \includegraphics[scale=.2]{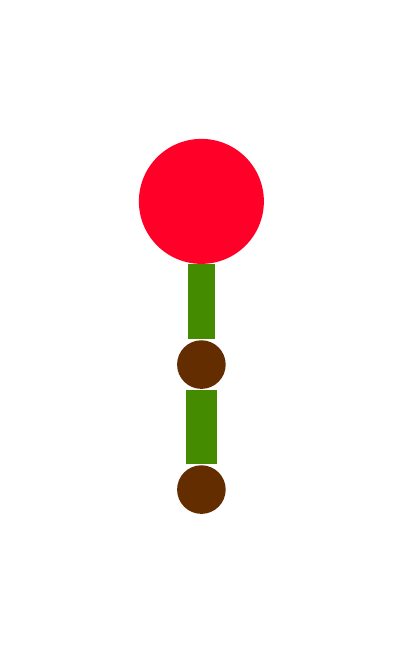}
    \includegraphics[scale=.2]{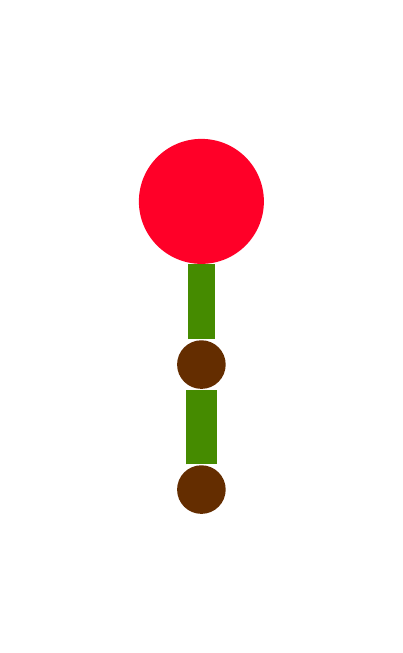}
    \includegraphics[scale=.2]{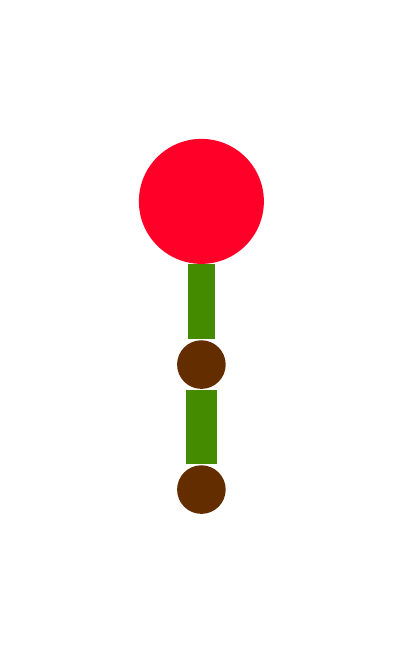}
    \includegraphics[scale=.2]{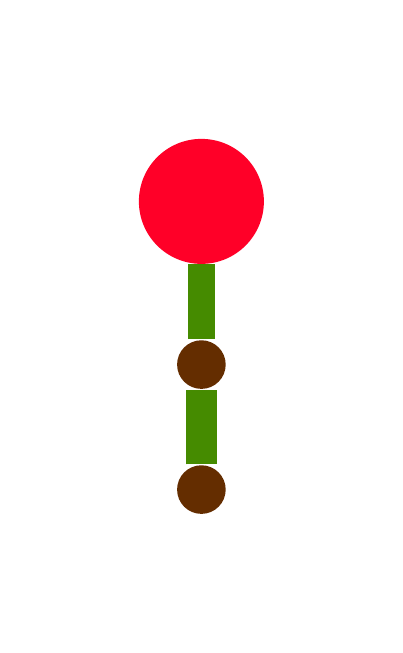}
    \includegraphics[scale=.2]{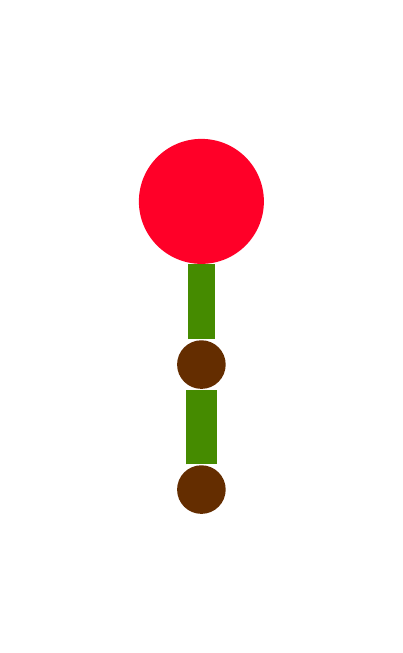}
    \includegraphics[scale=.2]{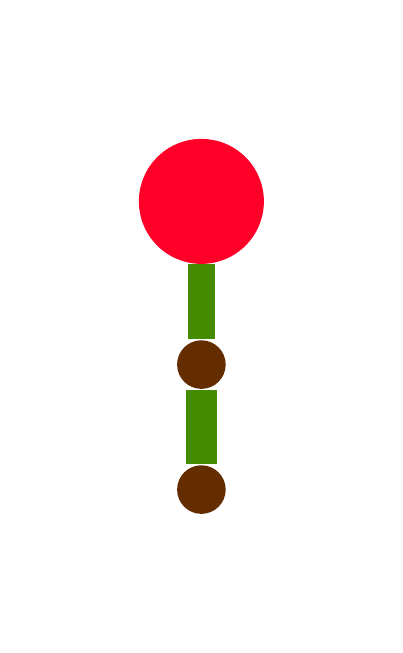}
    \includegraphics[scale=.2]{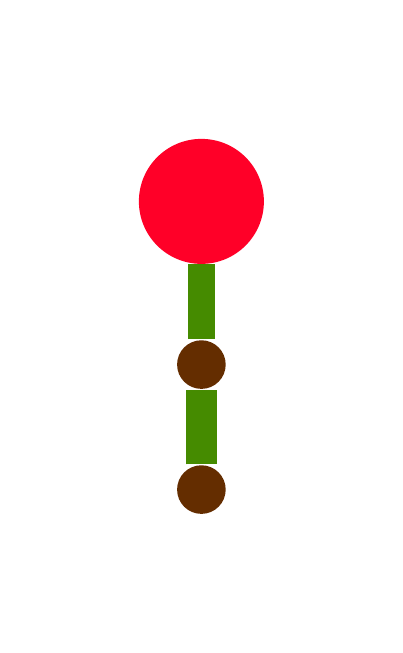}
    \label{fig:rec-orig}
    }
  \subfigure[Samples from the transformed (recursive) program.]{
    \includegraphics[scale=.2]{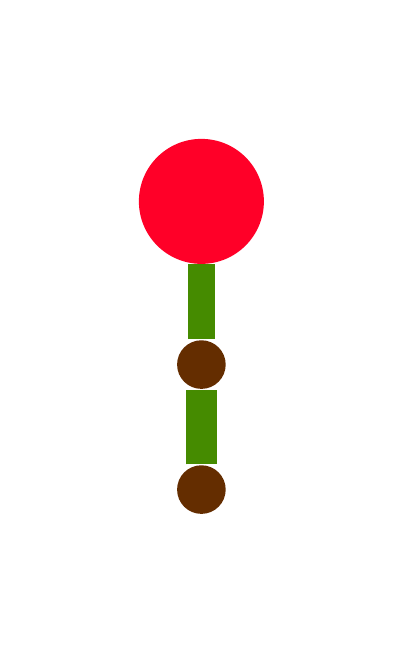}
    \includegraphics[scale=.2]{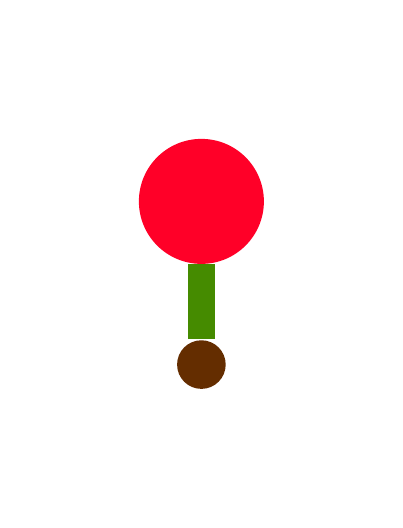}
    \includegraphics[scale=.2]{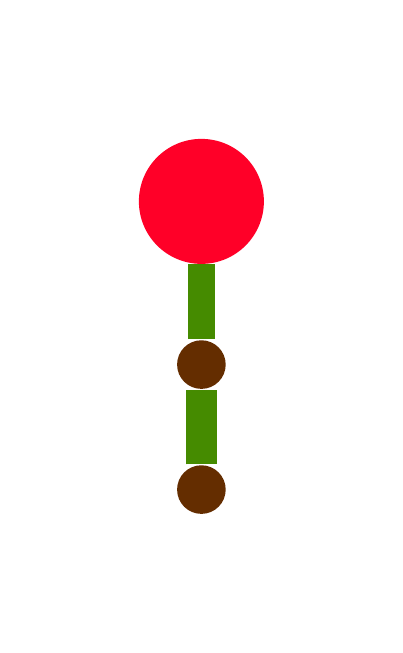}
    \includegraphics[scale=.2]{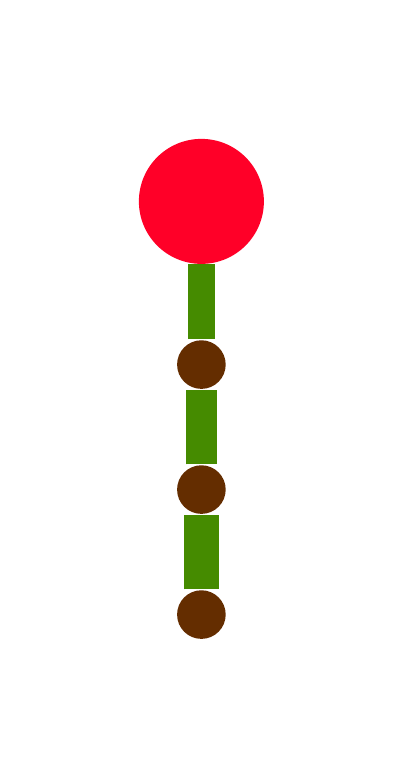}
    \includegraphics[scale=.2]{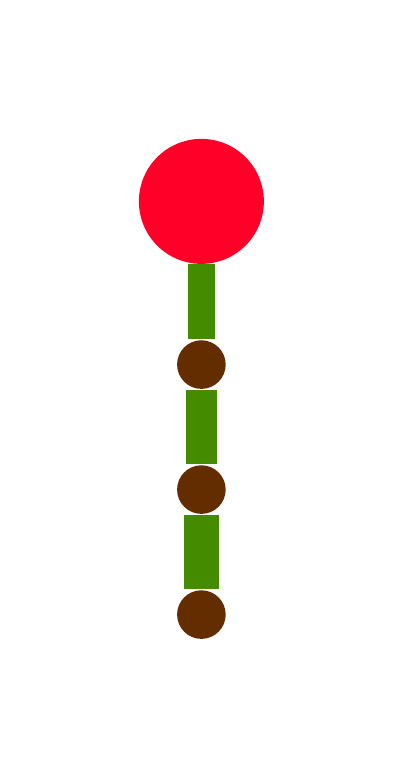}
    \includegraphics[scale=.2]{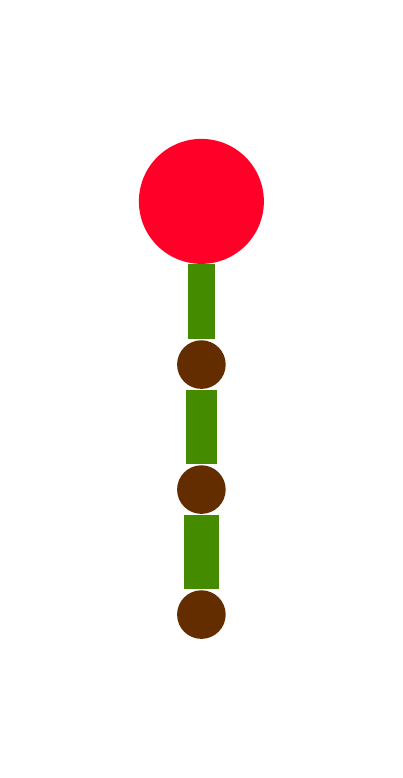}
    \includegraphics[scale=.2]{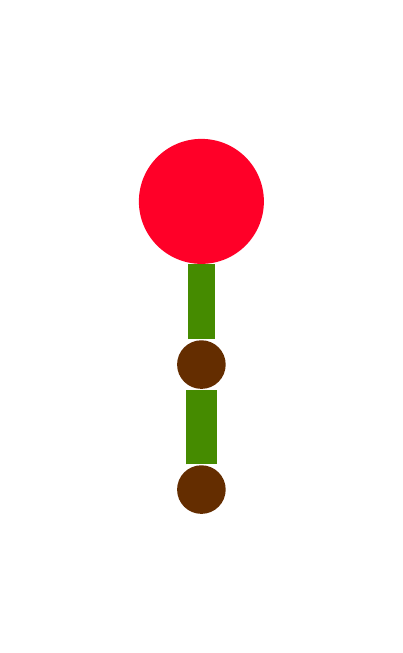}
    \includegraphics[scale=.2]{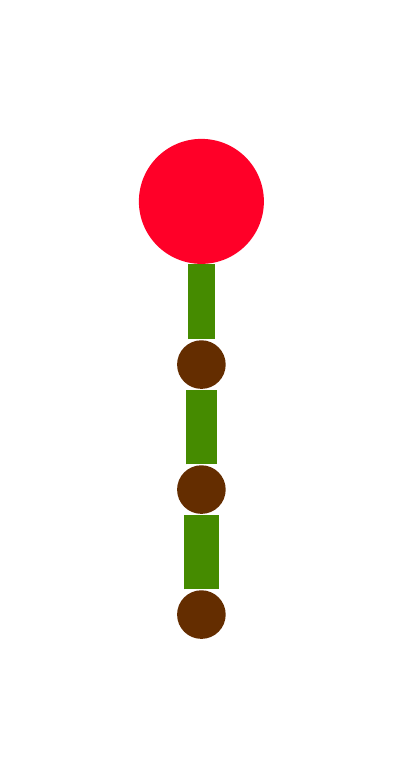}
    \includegraphics[scale=.2]{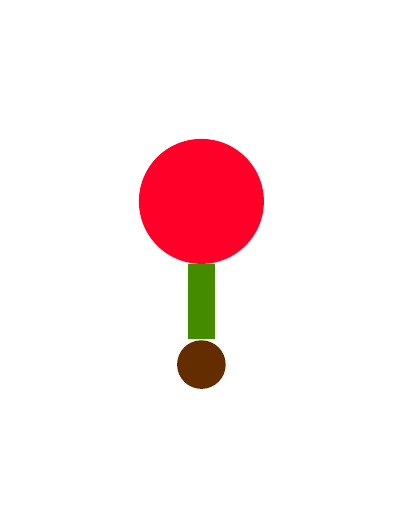}
    \label{fig:rec-transformed}    
    }
  \caption{The stochastic recursion transform allows the program to account for a wider range of data.}
  \label{fig:rec}
\end{figure}

We will illustrate this transform using the program \texttt{(begin (node (node a)))}. Samples from this program are shown in figure \ref{fig:rec-orig}. We apply the abstraction transform and get:
\begin{lstlisting}[mathescape=true]
(begin
  (define (F1 x)
    (node x))
  (F1 (F1 a)))
\end{lstlisting}
We now remove the argument to \texttt{F1} using deargumentation with the \texttt{remove-abstraction-variable} replacement function. This changes the definition of \texttt{F1} such that \texttt{x} is drawn from a uniform distribution on the instances of its former argument:
\begin{lstlisting}
(begin
  (define (F1 x)
    (($\lbda$ (x)
      (node x))
     (if (flip .5)
         (F1 a)
         a)))
  (F1 (F1 a)))
\end{lstlisting}
In this case, the instances of \texttt{x} previously were \texttt{a} and \texttt{(F1 a)}. Note that \texttt{a} is a constant and that the (outer) argument \texttt{x} to \texttt{F1} is not used in the body. We therefore remove the argument from \texttt{F1} and from any applications of \texttt{F1} (using \texttt{remove-application-argument}) and get:
\begin{lstlisting}[mathescape=true]
(begin
  (define (F1)
    (($\lbda$ (x)
      (node x))
     (if (flip .5)
         (F1)
         a)))
  (F1))
\end{lstlisting}
The data generated by this program (figure \ref{fig:rec-transformed}) is much more varied than the data of the program we started out with.

\subsubsection{Inducing noisy data constructors}

In the following, we demonstrate that noisy data constructors make it possible to represent some datasets much more compactly. We describe an instance of the deargumentation transform that can therefore infer when to use such constructors and we show an example of this process.

\begin{figure}[t]
\centering
\includegraphics[scale=.26]{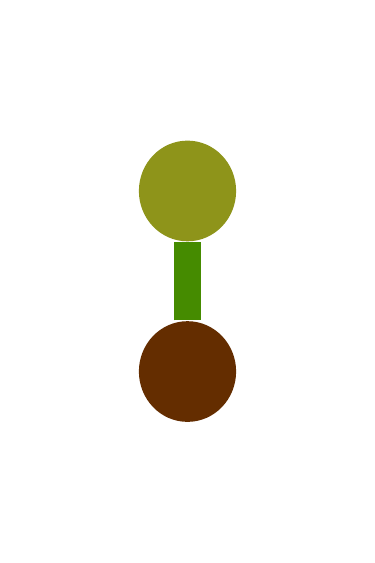}
\includegraphics[scale=.26]{./figures/4-2-4-noisy-constructor-example-b-0.pdf}
\includegraphics[scale=.26]{./figures/4-2-4-noisy-constructor-example-b-0.pdf}
\includegraphics[scale=.26]{./figures/4-2-4-noisy-constructor-example-b-0.pdf}
\includegraphics[scale=.26]{./figures/4-2-4-noisy-constructor-example-b-0.pdf}
\includegraphics[scale=.26]{./figures/4-2-4-noisy-constructor-example-b-0.pdf}
\includegraphics[scale=.26]{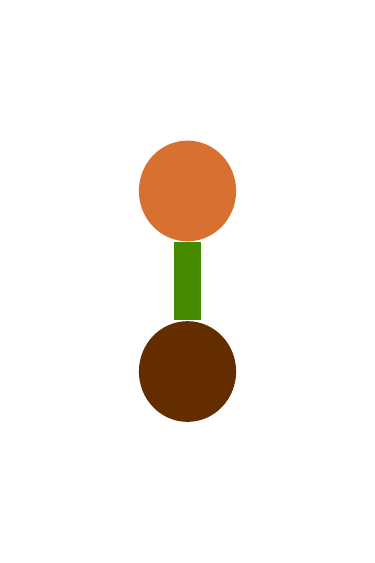}
\includegraphics[scale=.26]{./figures/4-2-4-noisy-constructor-example-b-0.pdf}
\includegraphics[scale=.26]{./figures/4-2-4-noisy-constructor-example-b-0.pdf}
\includegraphics[scale=.26]{./figures/4-2-4-noisy-constructor-example-b-0.pdf}
\caption{A stochastic generative process can represent these observations compactly.}
\label{fig:noiseCons}
\end{figure}

Consider the set of observations shown in figure \ref{fig:noiseCons}. Without noisy data constructors---i.e., if there was no \texttt{gaussian} inside calls to \texttt{color}---we would have to represent these observations using a program such as:
\begin{lstlisting}
(if (flip .9)
    (node (data (color 0) (size .4)) (node (data (color 85) (size .4))))
    (node (data (color 0) (size .4)) (node (data (color 140) (size .4)))))
\end{lstlisting}
If the observations are representative for the generative process we want to model, then we just paid a large penalty in terms of model size for a small gain in explanatory power. Noisy data constructors provide a similar gain at a much lower cost in program size:
\begin{lstlisting}
(node (data (color 20) (size .4))
      (node (data (color (gaussian 85 25) (size .4)))))
\end{lstlisting}

Previously, we have used noisy data constructors as part of \texttt{incorporate-data}. We now show how the gain in compactness due to the use of noisy constructors allows us to infer when to use them. We make two minor adjustments to \texttt{incorporate-data} and \texttt{noisy-number-replacement}. First, we change \texttt{node-data->expression}, a helper function used by \texttt{incorporate-data}, such that it does \textit{not} automatically add a call to \texttt{gaussian}. Instead, it now constructs deterministic nodes:
\begin{lstlisting}[frame=trblsingle]
(define (node-data->expression lst)
  `(data (color ,(first (second lst))) (size ,(first (third lst)))))
\end{lstlisting}
Second, we move the creation of noisy data constructors into the deargumentation replacement function \texttt{noisy-number-replacement}. Instead of returning the sample mean, this function now returns a call to \texttt{gaussian} that uses the sample mean and variance as parameters:
\begin{lstlisting}[frame=trblsingle]
(define (noisy-number-replacement program abstraction variable variable-instances)
  (if (all (map number? variable-instances))
      (let* ([instances-mean (mean variable-instances)]
	     [instances-deviation (sqrt (sample-variance variable-instances))])
	`(gaussian ,instances-mean ,instances-deviation))
      NO-REPLACEMENT))
\end{lstlisting}

\begin{figure}[t]
\centering
\includegraphics[scale=.26]{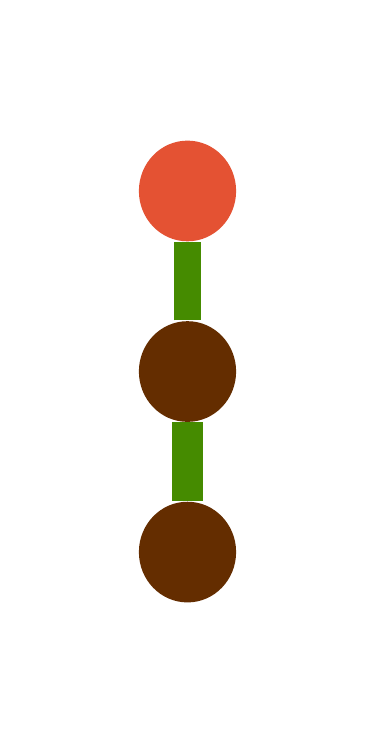}
\includegraphics[scale=.26]{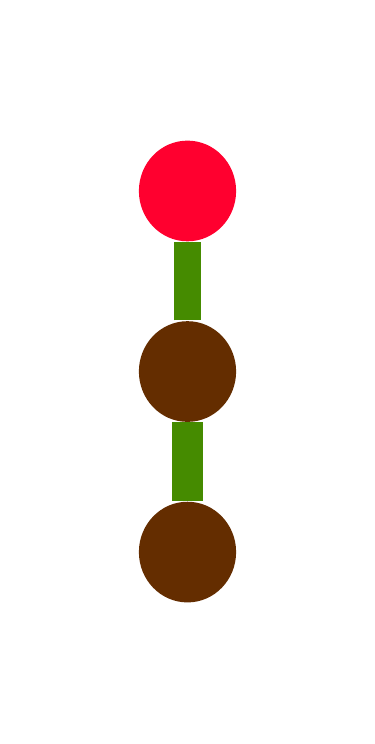}
\includegraphics[scale=.26]{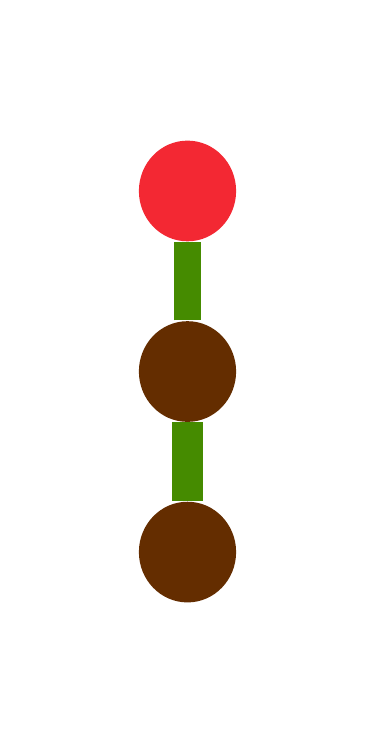}
\includegraphics[scale=.26]{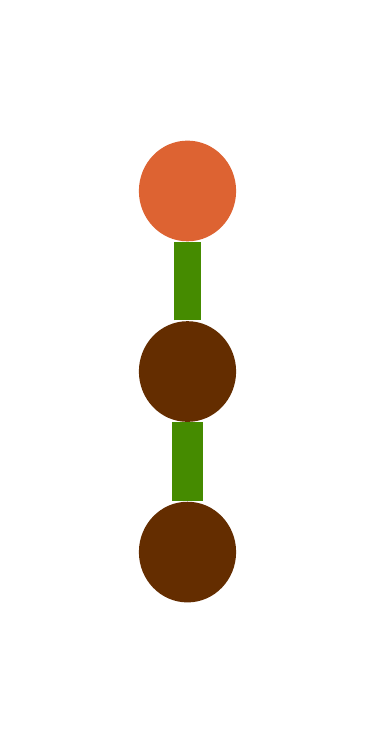}
\includegraphics[scale=.26]{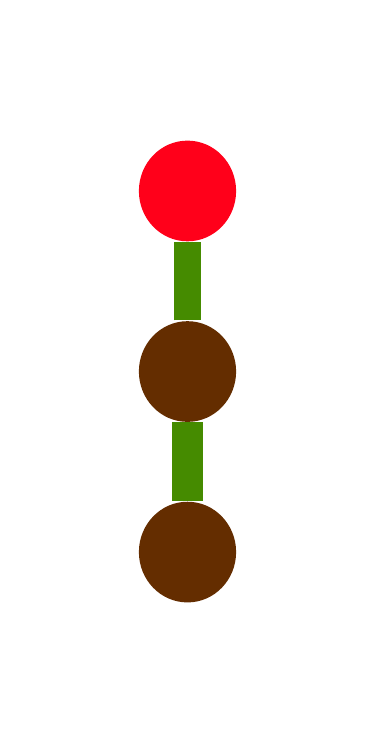}
\includegraphics[scale=.26]{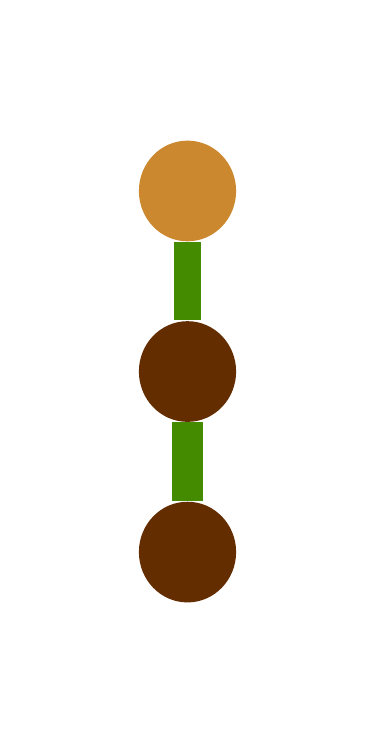}
\includegraphics[scale=.26]{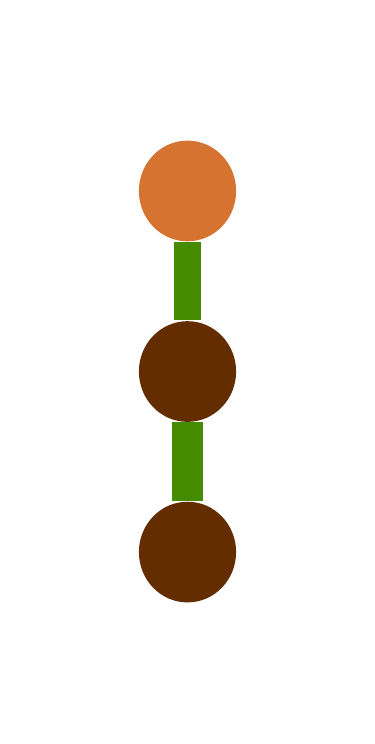}
\includegraphics[scale=.26]{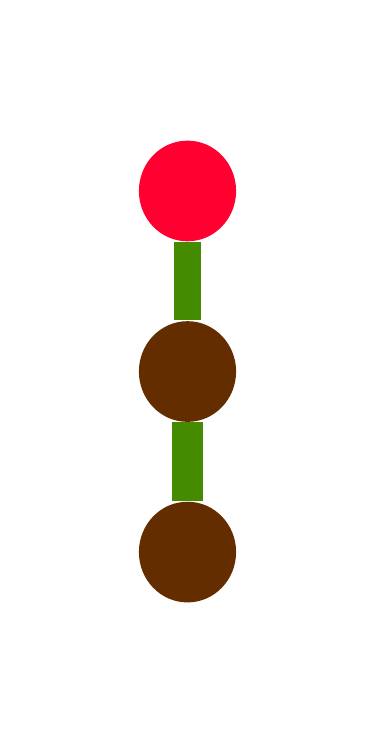}
\includegraphics[scale=.26]{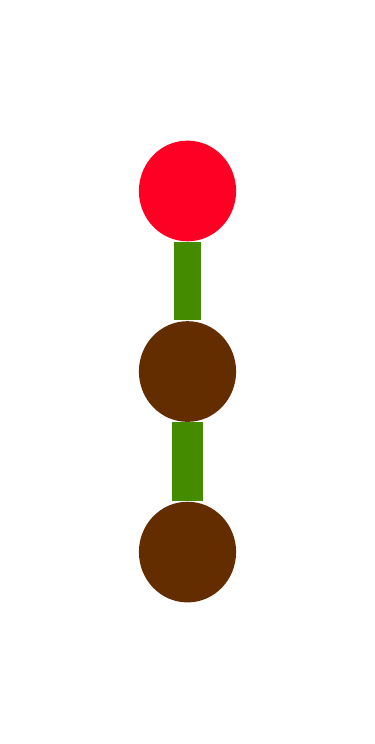}
\includegraphics[scale=.26]{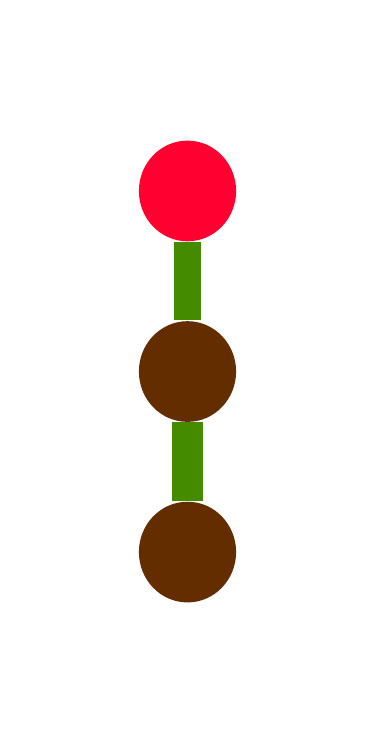}
\caption{Small variance in the data can be accounted for using stochastic data constructors.}
\label{fig:noiseCons2}
\end{figure}

To generate observations for an example of inducing noisy data constructors, we use the following program to sample trees. Each tree consists of three nodes with small variance in the color of the third node. Figure \ref{fig:noiseCons2} shows observations with this property.
\begin{lstlisting}[mathescape=true]
(define (three-node shade)
  (node (data (color 0) (size .4))
        (node (data (color 0) (size .4)) 
              (node (data (color (gaussian shade 10)) (size .4))))))

(list (three-node 200) (three-node 200) (three-node 200) (three-node 200)
      (three-node 200) (three-node 200) (three-node 200) (three-node 200)
      (three-node 200) (three-node 200))
\end{lstlisting}

Running data incorporation on a sample from this generative process results in a program that uniformly chooses between ten generated three-node structures.

\begin{lstlisting}[mathescape=true]
($\lbda$ ()
  (uniform-choice
   (node (data (color 0) (size 0.4))
	 (node (data (color 0) (size 0.4))
	       (node (data (color 209.0) (size 0.4)))))
   (node (data (color 0) (size 0.4))
	 (node (data (color 0) (size 0.4))
	       (node (data (color 196.0) (size 0.4)))))
                $\vdots$
   (node (data (color 0) (size 0.4))
	 (node (data (color 0) (size 0.4))
	       (node (data (color 206.0) (size 0.4)))))))
\end{lstlisting}

Bayesian program merging using the search moves we have described so far compresses this program into a representation that has one function for the three-node structure\footnote{We ran the system with $\alpha=3$, beam width 1, and depth 10 for this example.}. When this function generates the value of the third node, it makes use of a stochastic noise process. 

\begin{lstlisting}[mathescape=true]
(begin
  (define F2 ($\lbda$ (V2) (data (color V2) (size 0.3))))
  (define F1
    ($\lbda$ ()
      (($\lbda$ (V1)
         (node (F2 0) (node (F2 0) (node (F2 V1)))))
       (gaussian 202.3 9.286190463980603))))
  (uniform-choice (F1) (F1) (F1) (F1) (F1) (F1) (F1) (F1) (F1) (F1)))
\end{lstlisting}

While the discussion above hints at the representational benefits of probabilistic data constructors and probabilistic programs, we must postpone a more complete treatment of the implications of this design decision to future work.

\subsubsection{Deargumentation as program induction}
\label{sec:dearg-as-induction}

Deargumentation solves the following problem: Given all values that a function argument takes on in the course of running the overall generative process, we want to determine whether there is a compact stochastic program that can generate all of these values---if this is the case, we can remove the argument and use the definition of this compact program within our function. When phrased in these terms, it is clear that the problem addressed by deargumentation is itself a problem of program induction. We can rephrase the deargumentation moves we have presented as induction steps limited to particular sets of programs:

\begin{enumerate}
\item In the case of noisy-number deargumentation, we induce programs that generate all of the occurring number values, restricting our search to programs corresponding to one-dimensional Gaussian distributions.
\item In the case of recursion deargumentation, we restrict the induced programs to stochastic recursions of the program being deargumented.
\item In the case of the identical variable deargumentation, we assume that the generative process of the values for one variable is identical to the other, hence we ``induce'' the program for the other variable by simply referring to the first one.
\end{enumerate}

This suggests a reformulation of the deargumentation transform in terms of recursive calls to the overall Bayesian program merging procedure, which could make this class of transforms much more general.

\newpage
\section{Search strategy: Beam search}

In our search through program space, we are interested in programs with high posterior probability. Many different search strategies are possible, but in this report, we limit ourselves to beam search using the posterior probability as a search heuristic.

\begin{lstlisting}[frame=trbl]
(define (beam-search data init-program beam-size depth)
  (let* ([all-transformations
          (beam-search-transformations data init-program beam-size depth)]
         [top-transformations (sort-by-posterior data all-transformations)])
    (if (null? top-transformations)
        init-program
        (program+->program (first top-transformations)))))

(define (beam-search-transformations data program beam-size depth)
  (let* ([lp (log-prior program)]
         [ll (log-likelihood data program 10)]
         [init-program+ (make-program+ program 0 ll lp #f)])
    (depth-iter-transformations ($\lbda$ (programs+)
                                  (best-n data programs+ beam-size))
                                init-program+
                                depth)))

(define (best-n data programs+ n)
  (max-take (sort-by-posterior data programs+) n))
\end{lstlisting}
The main part of the search is performed by \texttt{depth-iter-transformations}, which recursively applies program transformations to the best programs at a given search depth and then filters the results to get the best programs for the next depth.
\begin{lstlisting}[frame=trbl]
(define (depth-iter-transformations cfilter program+ depth)
  (let* ([transformed-programs+
          (apply-and-filter-transformations depth cfilter program+)]
         [next-depth ($\lbda$ (prog)
                       (depth-iter-transformations cfilter prog (- depth 1)))])
    (delete '()
            (append transformed-programs+
                    (apply append
                           (map next-depth
                                transformed-programs+))))))
\end{lstlisting}
We reduce the amount of computation required when choosing the best programs at each level of search by separating program transformations that preserve semantics from those that do not.  Marking programs based on their transformation type allows us to reuse previously computed likelihoods for programs that were created by a semantics preserving transformation.
\begin{lstlisting}[frame=trbl]
(define (apply-and-filter-transformations depth cfilter program+)
  (if (= depth 0)
      '()
      (let* ([semantics-preserved-programs+ (apply-transformations program+ semantic-preserving-transformations #t)]
             [semantics-changed-programs+ (apply-transformations program+ semantic-changing-transformations #f)])
        (cfilter (append semantics-preserved-programs+ semantics-changed-programs+)))))

(define (apply-transformations program+ transformations semantics-preserving)
  (let* ([program (program+->program program+)]
         [transformed-programs (delete '() (concatenate (map ($\lbda$ (transform) (transform program #t)) transformations)))]
         [transformed-programs+ (map ($\lbda$ (program) (program+->program-transform semantics-preserving program+ program)) transformed-programs)])
    transformed-programs+))
\end{lstlisting}
We compute an estimate of the posterior probability of each program by combining an estimate of the likelihood with the length prior:
\begin{lstlisting}[frame=trbl]
(define (sort-by-posterior data programs+)
  (let* ([programs (map program+->program programs+)]
         [semantics-flags (map program+->semantics-preserved programs+)]
         [log-priors (map log-prior programs)]
         [log-likelihoods (map ($\lbda$ (prog+ semantics-flag)
                                 (if semantics-flag
                                     (program+->log-likelihood prog+)
                                     (log-likelihood data (program+->program prog+) 10))) programs+ semantics-flags)]
         [posteriors (map + log-priors log-likelihoods)] 
         [new-programs+ (map make-program+ programs posteriors log-likelihoods log-priors semantics-flags)]
         [posteriors> ($\lbda$ (a b) (> (program+->posterior a) (program+->posterior b)))])
    (list-sort posteriors> new-programs+)))
\end{lstlisting}

\newpage
\section{Examples}

We illustrate our program induction method using small examples drawn from the domain of colored trees. All examples have the following shape: We present a probabilistic program, show observations (trees) sampled from this program, highlight the key patterns that are visible both in the original program and in the observed trees, and then show which patterns Bayesian program merging recovers.  The results below are representative of running Bayesian program merging on the examples, but the detailed results of any given run will differ due to the stochastic nature of likelihood estimation.

\subsection{Single color flower}

\begin{figure}
  \subfigure[Samples from the original program.]{
    \includegraphics[scale=.2]{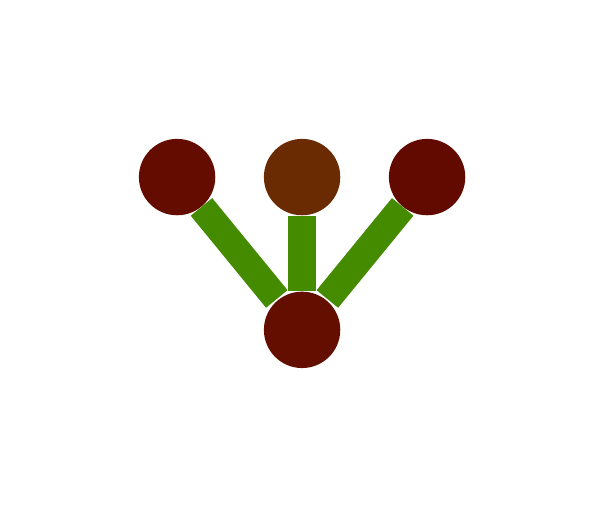}
    \includegraphics[scale=.2]{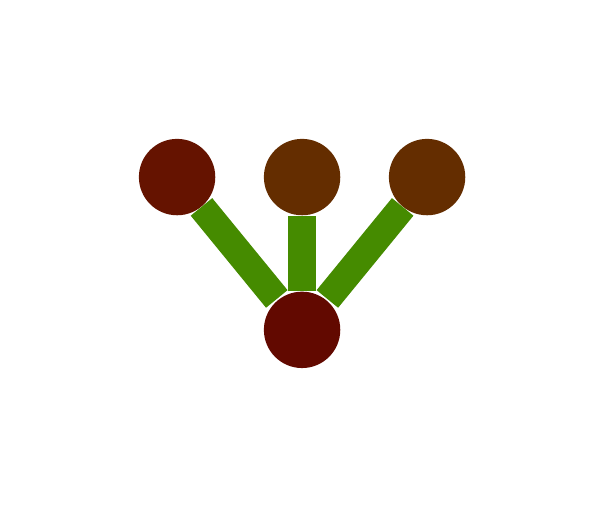}
    \includegraphics[scale=.2]{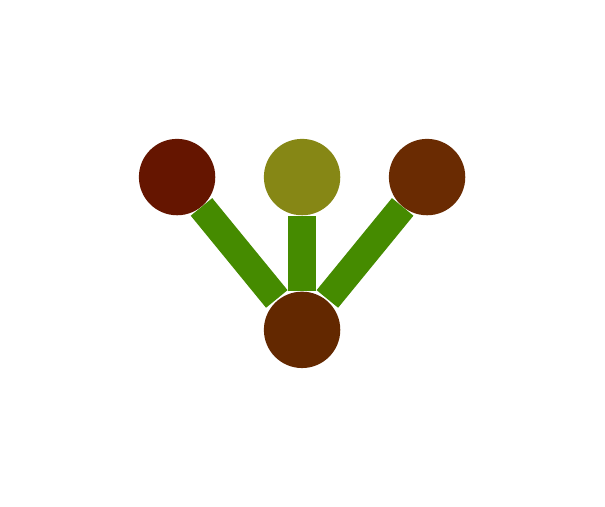}
    \includegraphics[scale=.2]{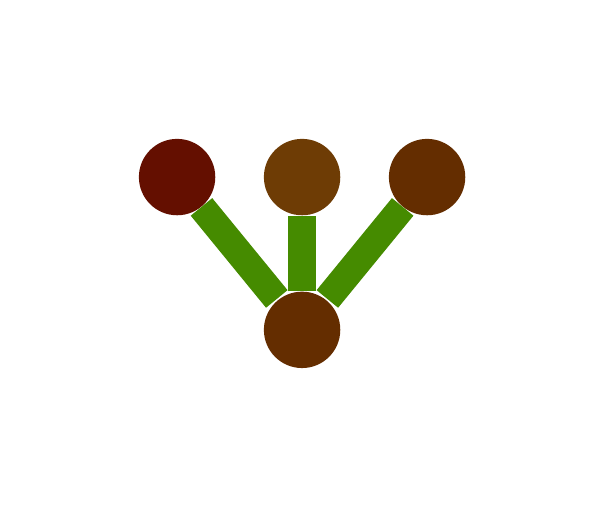}
    \includegraphics[scale=.2]{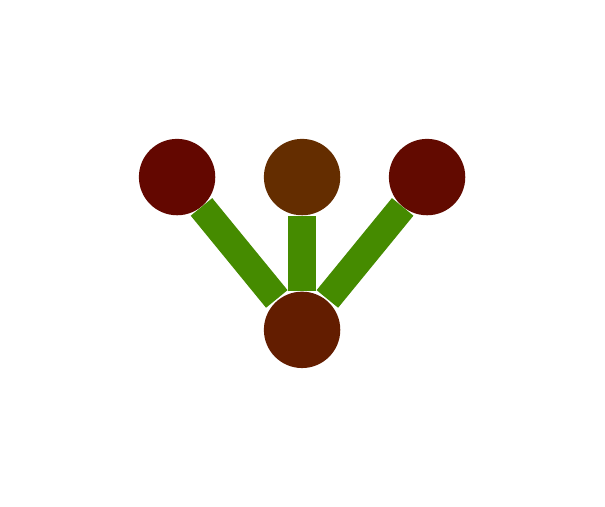}
    \includegraphics[scale=.2]{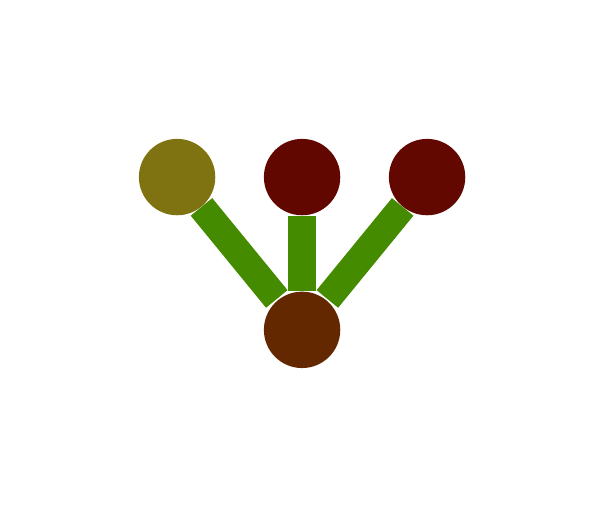}
    \includegraphics[scale=.2]{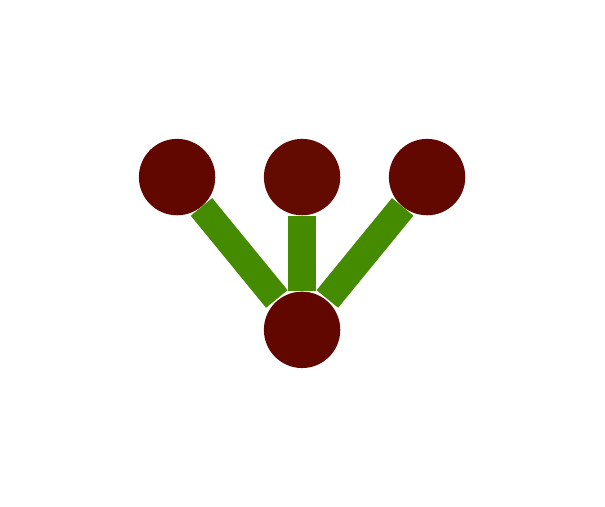}
    \includegraphics[scale=.2]{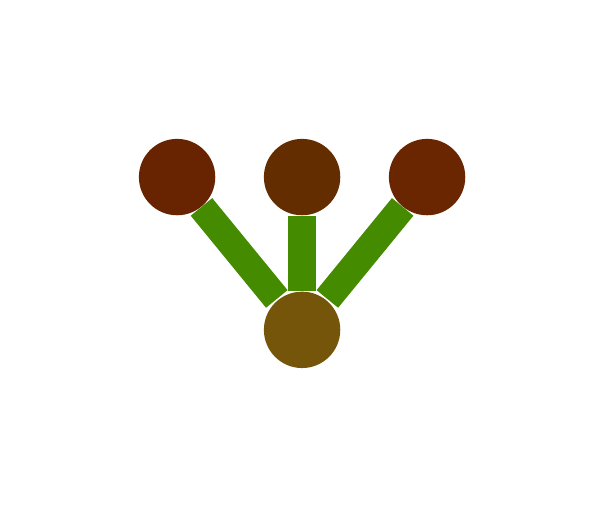}
    \includegraphics[scale=.2]{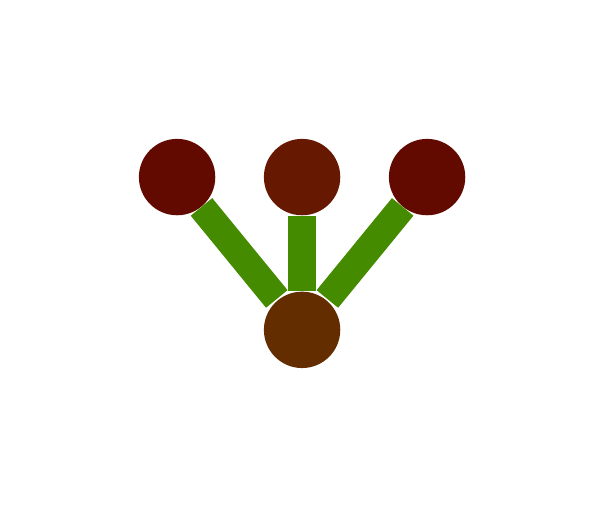}
    \label{fig:exp-single-flower-data}
    }
    \subfigure[Samples from the induced program.]{
    \includegraphics[scale=.2]{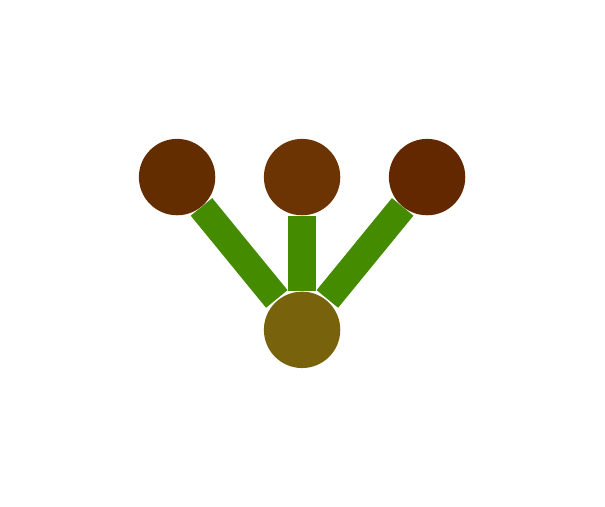}
    \includegraphics[scale=.2]{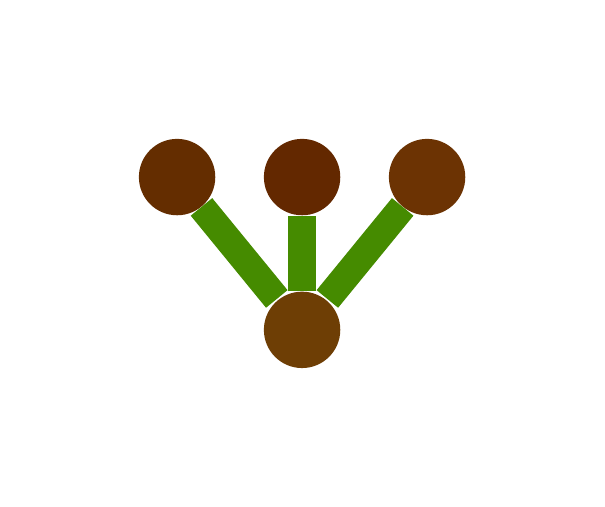}
    \includegraphics[scale=.2]{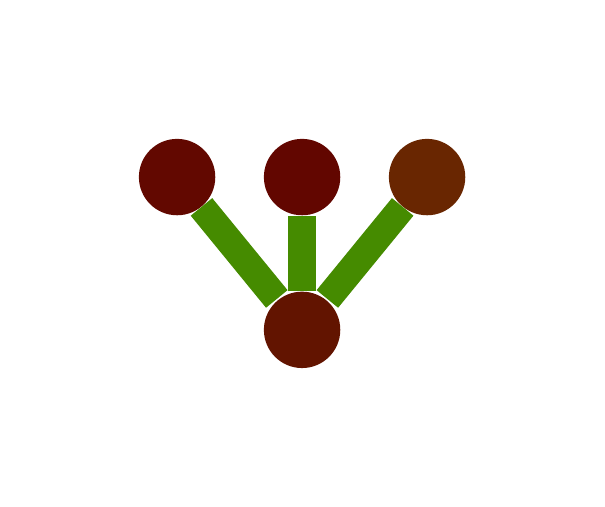}
    \includegraphics[scale=.2]{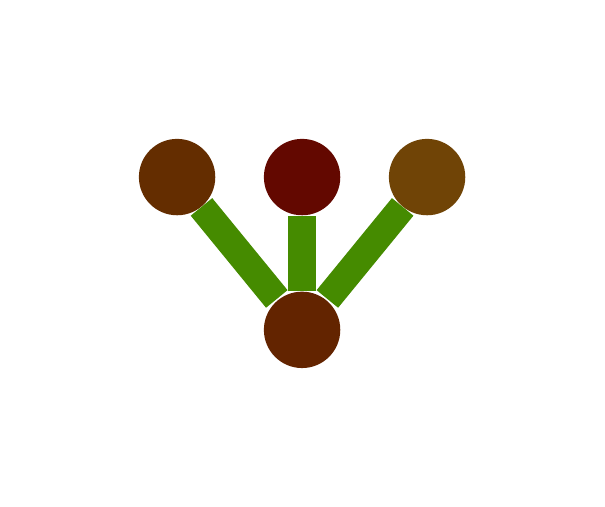}
    \includegraphics[scale=.2]{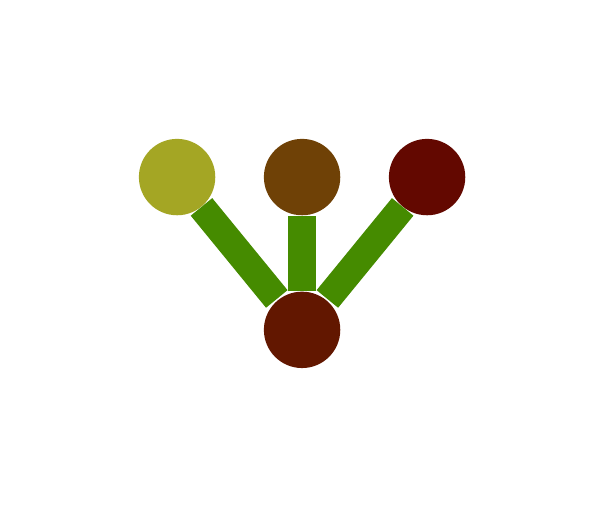}
    \includegraphics[scale=.2]{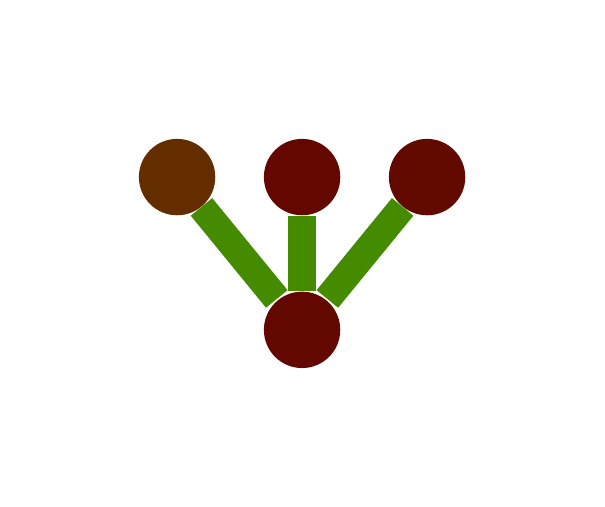}
    \includegraphics[scale=.2]{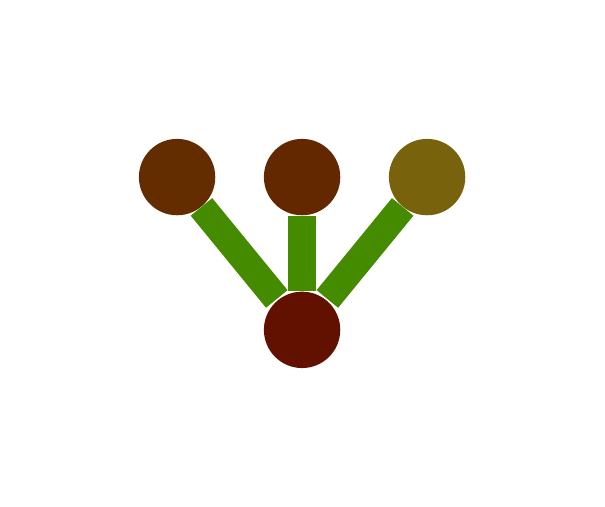}
    \includegraphics[scale=.2]{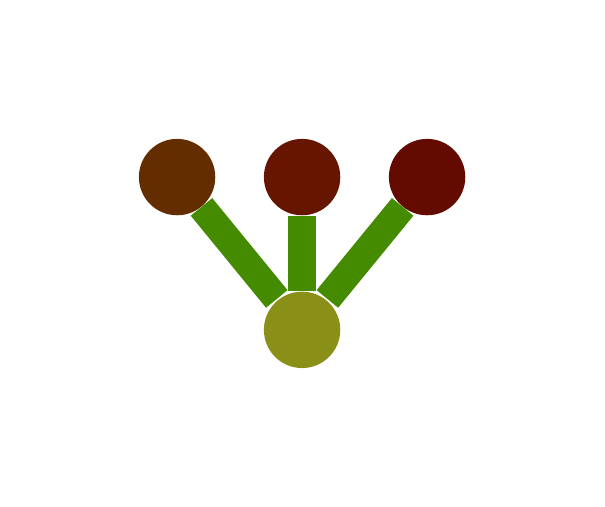}
    \includegraphics[scale=.2]{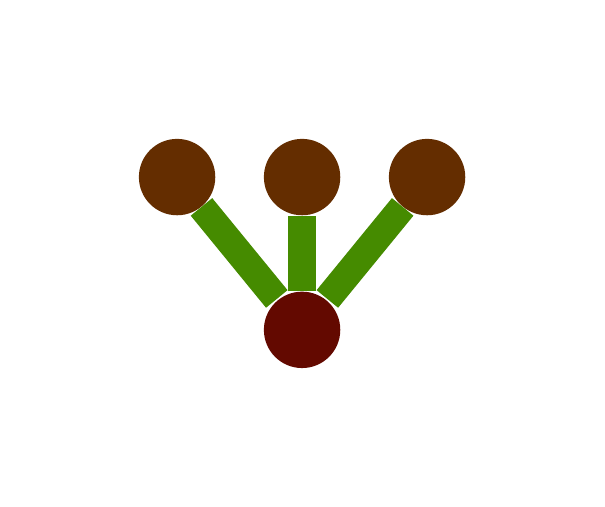}
    \label{fig:exp-single-flower-induced}    
    }
  \caption{Example 1: Single color flower.}
  \label{fig:exp-single-flower}
\end{figure}

The program below generates ten ``flower'' instances that are roughly the same  color.
\begin{lstlisting}[mathescape=true]
(define (flower shade)
  (node (data (color (gaussian shade 25)) (size .3))
        (petal shade)
        (petal shade)
        (petal shade)))

(define (petal shade)
  (node (data (color (gaussian shade 25)) (size .3))))

(repeat 10 ($\lbda$ () (flower 20)))
\end{lstlisting}
We create an initial program that represents these ten flowers using data incorporation and then compress it (result shown below).  We learn a function \texttt{F1} that takes no arguments and creates a flower with petals that are similar in color. The system parameters were $\alpha=1$, beam width $1$, and depth $10$.
\begin{lstlisting}[mathescape=true]
(begin
  (define F2
    ($\lbda$ (V5)
      (data (color (gaussian V5 25)) (size 0.3))))
  (define F1
    ($\lbda$ ()
      (($\lbda$ (V4)
	 (($\lbda$ (V2)
	    (($\lbda$ (V1)
	       (($\lbda$ (V3)
		  (node (F2 V1) (node (F2 V2))
			(node (F2 V3)) (node (F2 V4))))
		17.2))
	     32.9))
	  2.0))
       19.7)))
  ($\lbda$ ()
    (uniform-choice (F1) (F1) (F1) (F1) (F1) (F1) (F1)
		    (F1) (F1) (F1))))
\end{lstlisting}
\subsection{Multiple color flower}
We use a program similar to the previous one to generate ten ``flower'' instances, some of which have mostly red petals, some mostly green.
\begin{lstlisting}[mathescape=true]
(define (flower shade)
  (node (data (color (gaussian 0 25)) (size .3))
        (petal shade)
        (petal shade)
        (petal shade)))

(define (petal shade)
  (node (data (color (gaussian shade 25)) (size .3))))

(repeat 10 ($\lbda$ () (flower (if (flip) 100 220))))
\end{lstlisting}
Bayesian program merging results in the program shown below. The abstraction \texttt{F2} corresponds to \texttt{petal} and \texttt{F1} corresponds to \texttt{flower}.  System parameters: $\alpha=1$, beam width $1$, and depth $10$.
\begin{lstlisting}[mathescape=true]
(begin
  (define F2
    ($\lbda$ (V5)
      (data (color (gaussian V5 25)) (size 0.3))))
  (define F1
    ($\lbda$ (V2)
      (($\lbda$ (V1)
         (($\lbda$ (V3)
            (($\lbda$ (V4)
               (node (F2 V1) (node (F2 V2)) (node (F2 V3))
                     (node (F2 V4))))
             V2))
          V1))
       V2)))
  ($\lbda$ ()
    (uniform-choice (F1 91.0) (F1 85.0) (F1 254.0) (F1 234.0) (F1 82.0)
                    (F1 243.0) (F1 104.0))))
\end{lstlisting}

\begin{figure}
  \subfigure[Samples from the original program.]{
    \includegraphics[scale=.2]{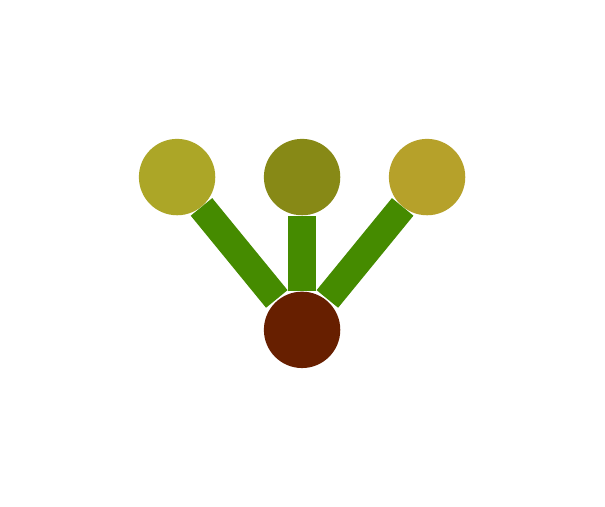}
    \includegraphics[scale=.2]{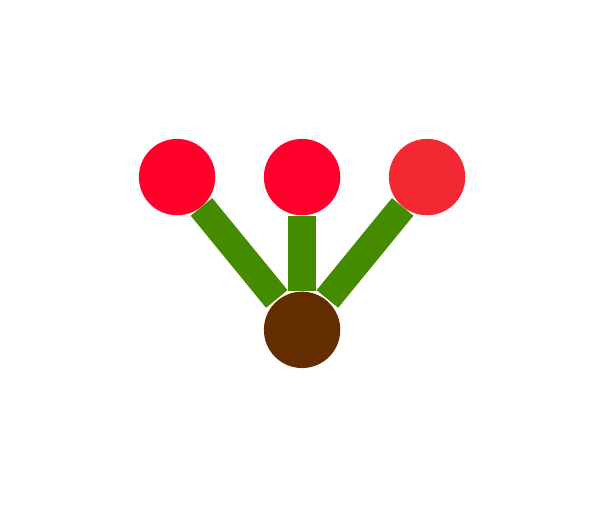}
    \includegraphics[scale=.2]{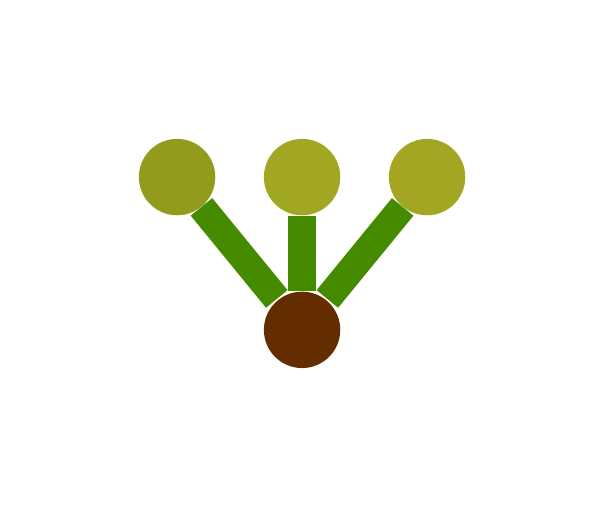}
    \includegraphics[scale=.2]{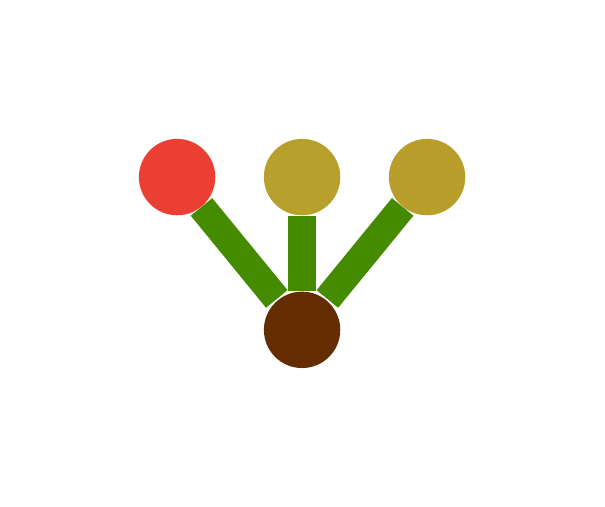}
    \includegraphics[scale=.2]{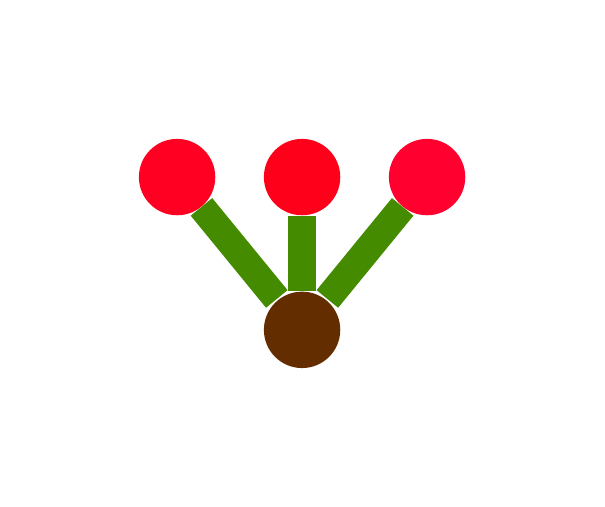}
    \includegraphics[scale=.2]{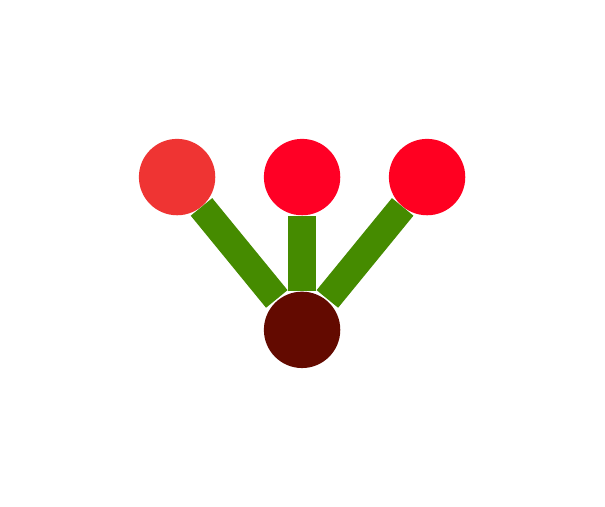}
    \includegraphics[scale=.2]{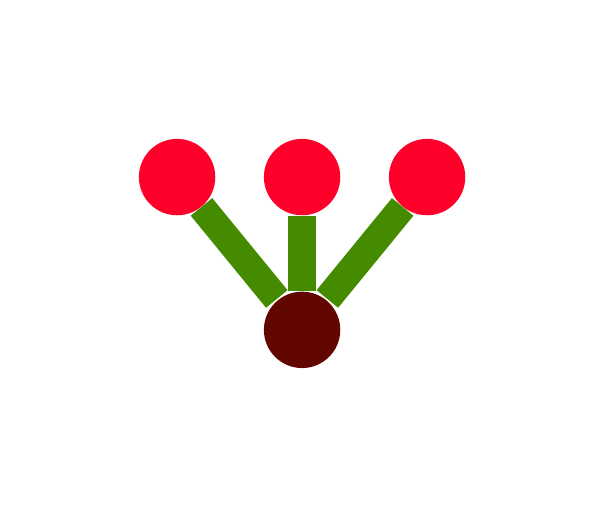}
    \includegraphics[scale=.2]{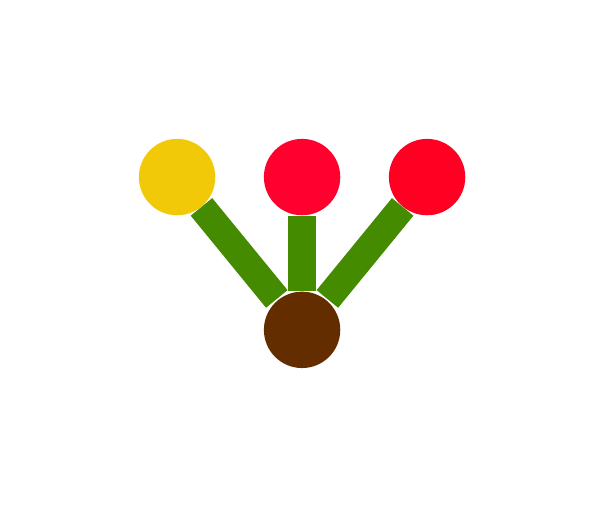}
    \includegraphics[scale=.2]{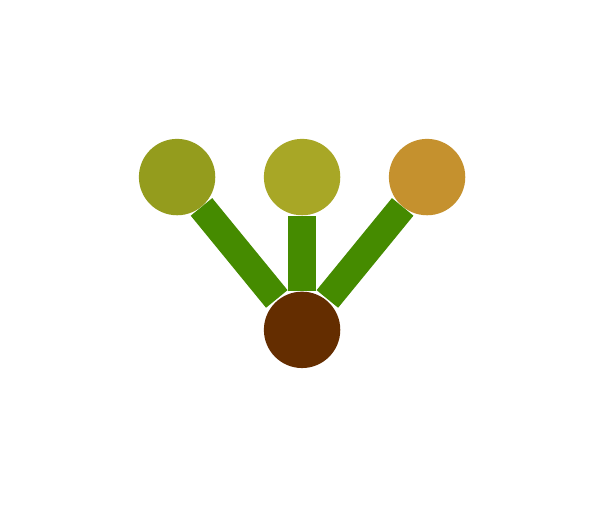}
    \label{fig:exp-multiple-flower-data}
    }
    \subfigure[Samples from the induced program.]{
    \includegraphics[scale=.2]{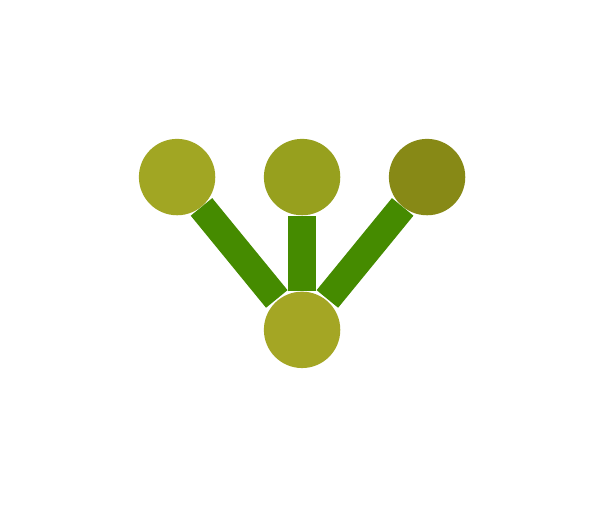}
    \includegraphics[scale=.2]{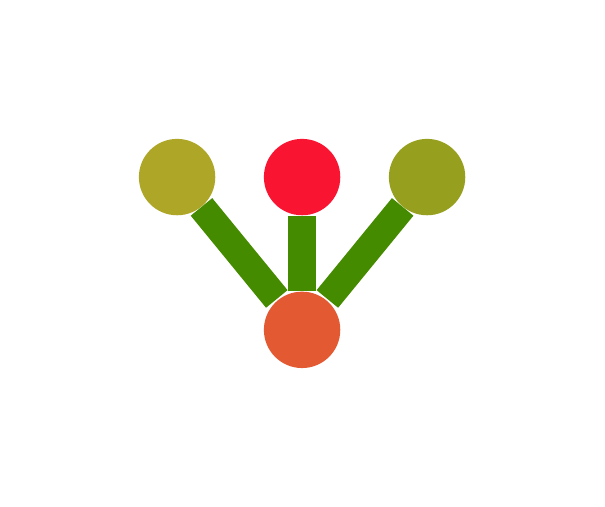}
    \includegraphics[scale=.2]{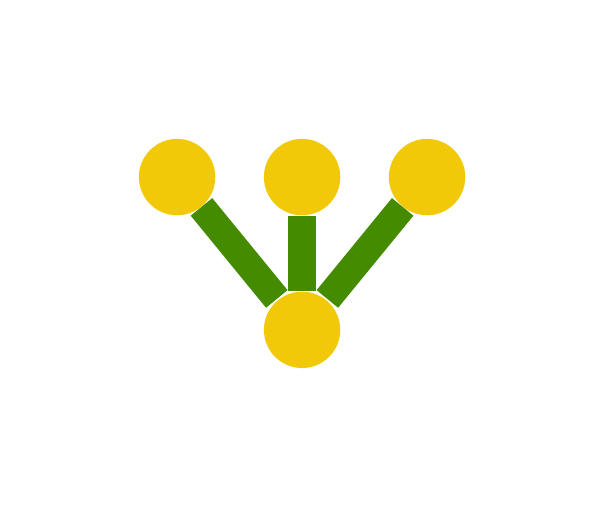}
    \includegraphics[scale=.2]{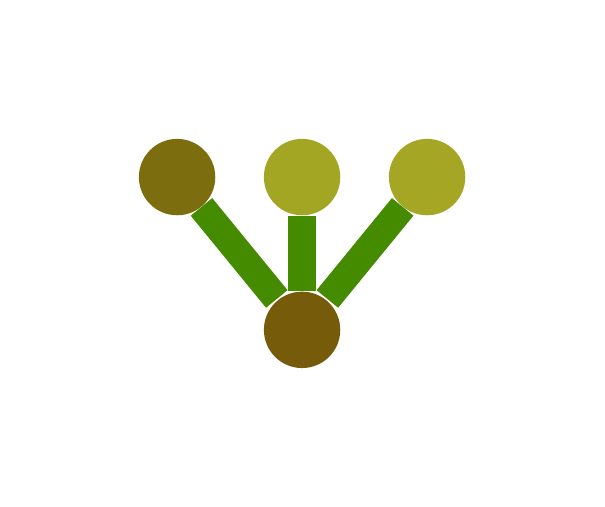}
    \includegraphics[scale=.2]{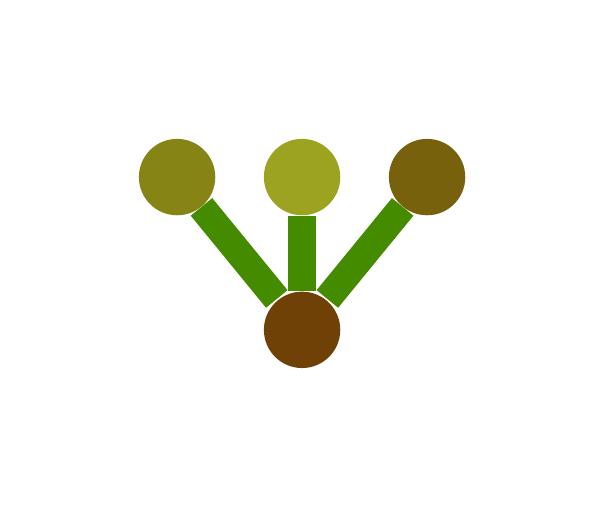}
    \includegraphics[scale=.2]{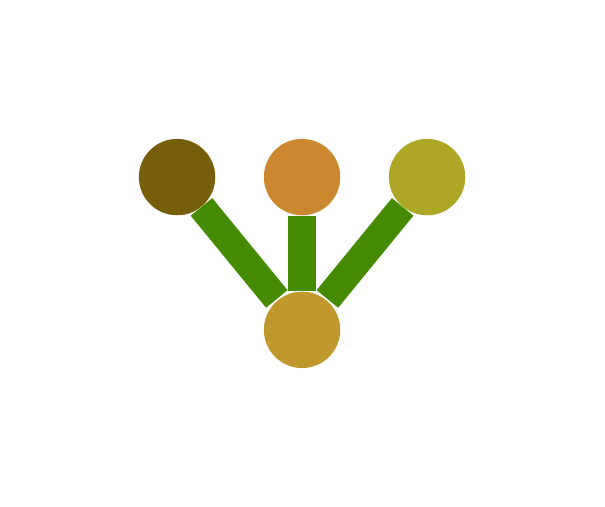}
    \includegraphics[scale=.2]{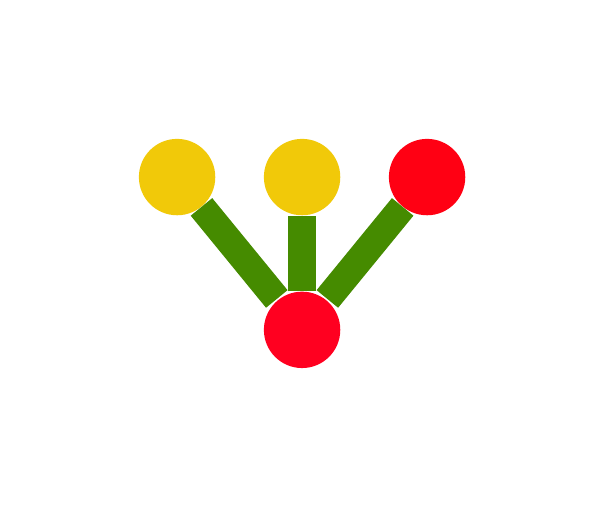}
    \includegraphics[scale=.2]{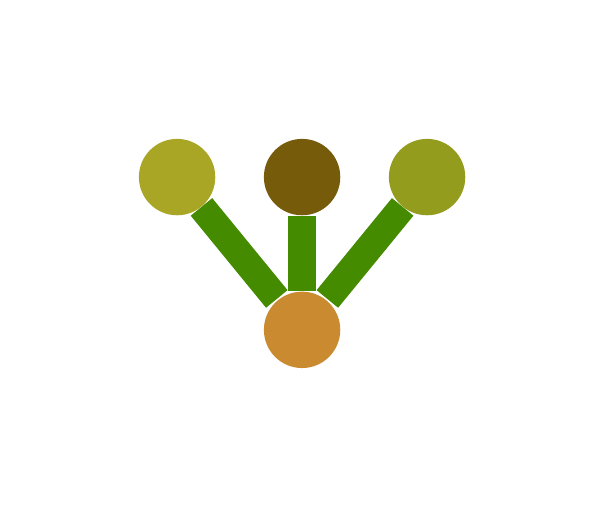}
    \includegraphics[scale=.2]{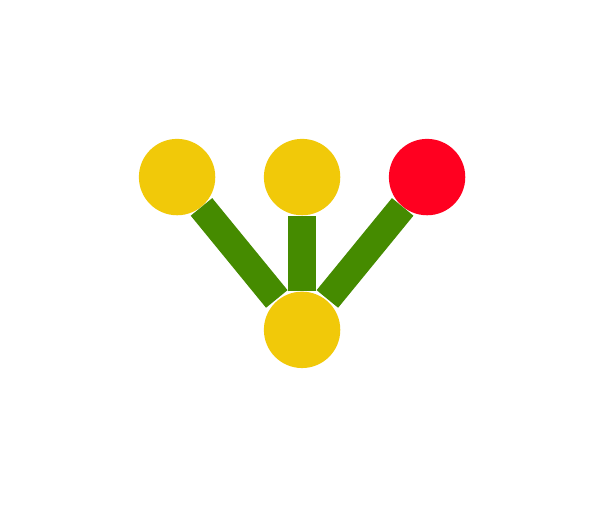}
    \label{fig:exp-multiple-flower-induced}    
    }
  \caption{Example 2: Multiple color flower.}
  \label{fig:exp-multiple-flower}
\end{figure}

\subsection{Simple recursion}
The next program generates a tree made of a single stem of nodes.
\begin{lstlisting}[mathescape=true]
(define (stem)
  (if (flip .1)
      (node (data (color 200) (size .5)))
      (node (data (color 200) (size .5)) (stem))))

(repeat 5 stem)
\end{lstlisting}
With parameters $\alpha=1$, beam width 1, and depth 10, Bayesian program merging finds the following recursion: \texttt{F1} recursively calls itself or stops at \texttt{F2}, thus building up a line of nodes.
\begin{lstlisting}[mathescape=true]
(begin
  (define F3 ($\lbda$ () (node (F2))))
  (define F2
    ($\lbda$ () (data (color (gaussian 200 25)) (size 0.5))))
  (define F1
    ($\lbda$ ()
      (($\lbda$ (V1) (node (F2) V1))
       (if (flip 29/33)
	   (F1)
	   (uniform-choice (F3) (F3) (F3) (F3))))))
  ($\lbda$ () (uniform-choice (F1) (F3) (F1) (F1) (F1))))
\end{lstlisting}
Once the size of the observations generated by the original recursive program reaches a length threshold, the trade-off between prior and likelihood favors recursive programs due to their small size. We can adjust where this threshold occurs using the size constant, $\alpha$, in the prior. 

\begin{figure}
  \subfigure[Samples from the original program.]{
    \includegraphics[scale=.14]{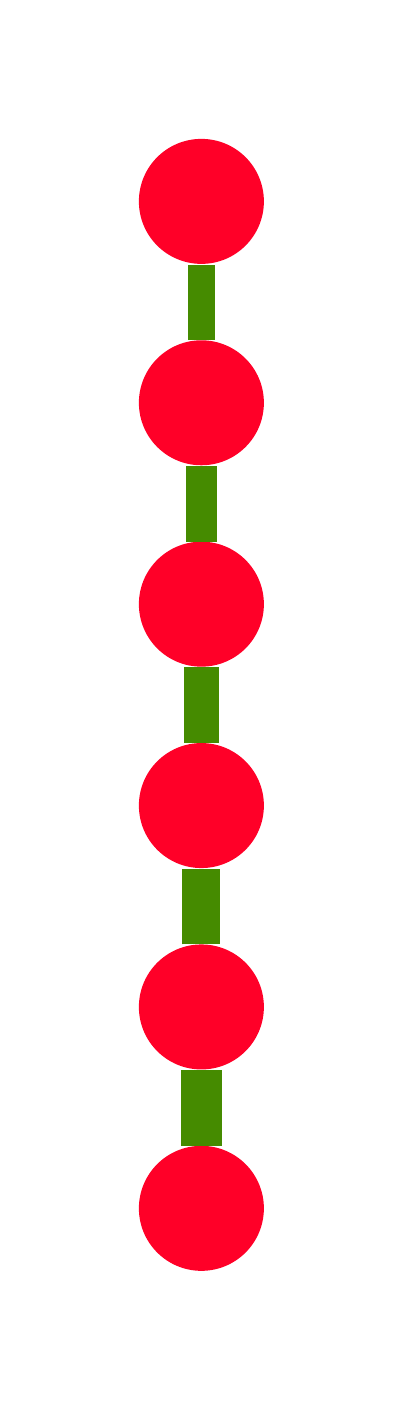}
    \includegraphics[scale=.14]{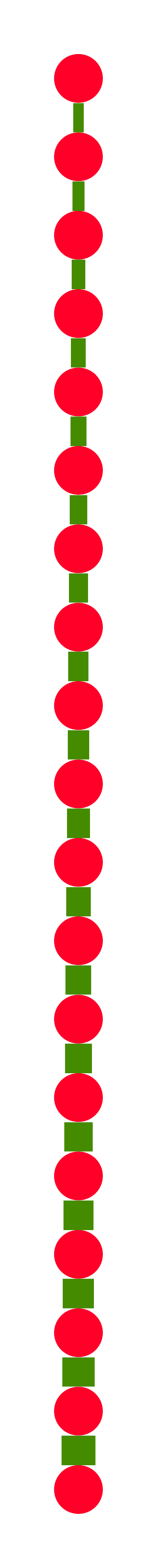}
    \includegraphics[scale=.14]{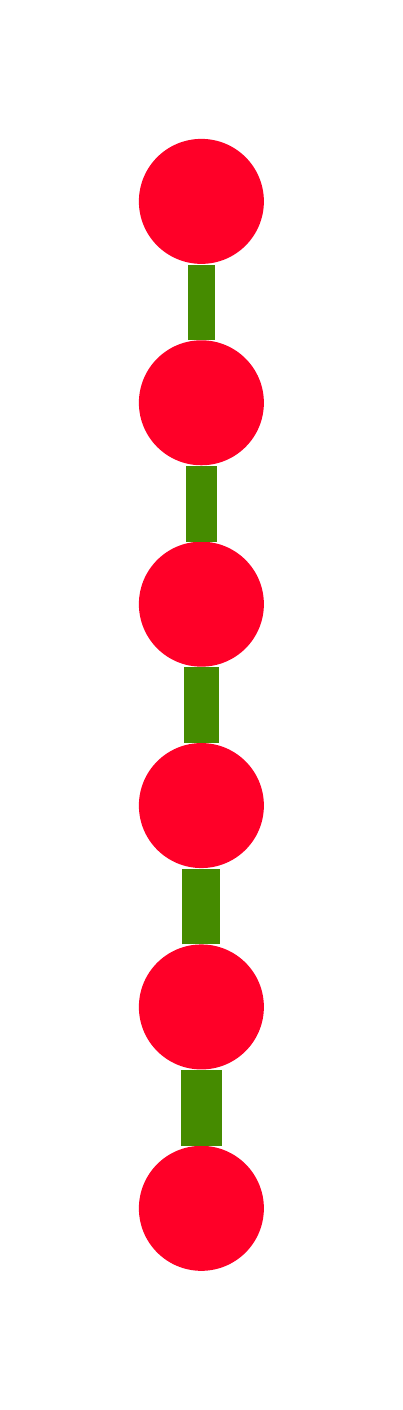}
    \includegraphics[scale=.14]{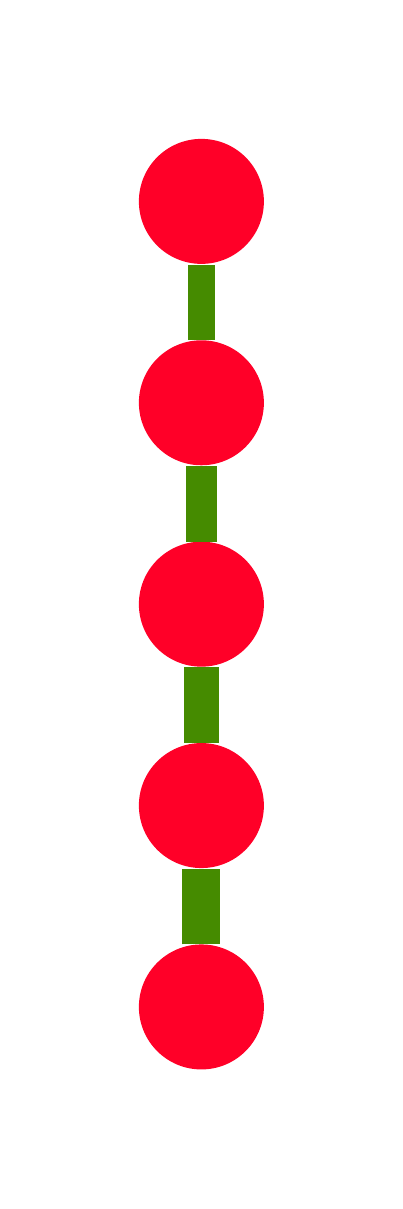}
    \includegraphics[scale=.14]{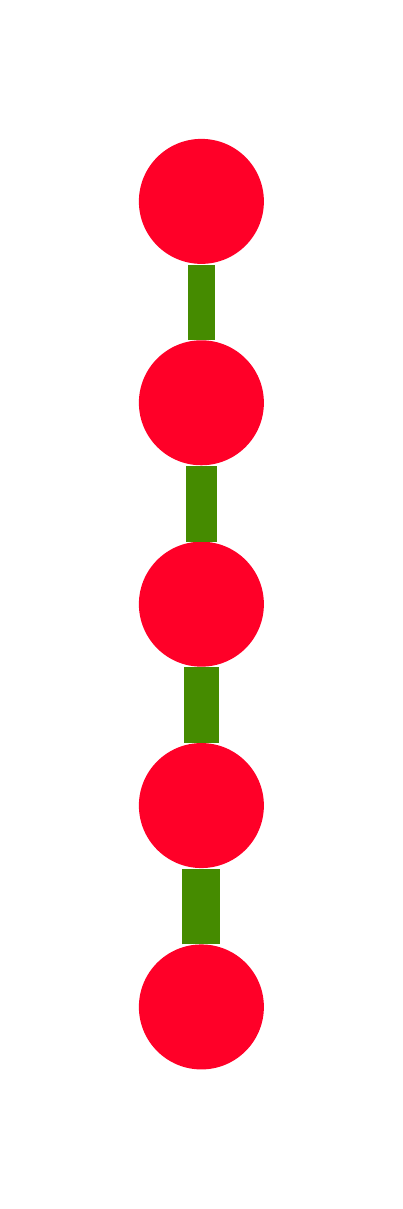}
    \includegraphics[scale=.14]{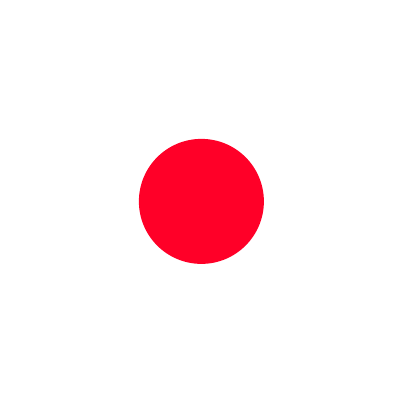}
    \includegraphics[scale=.14]{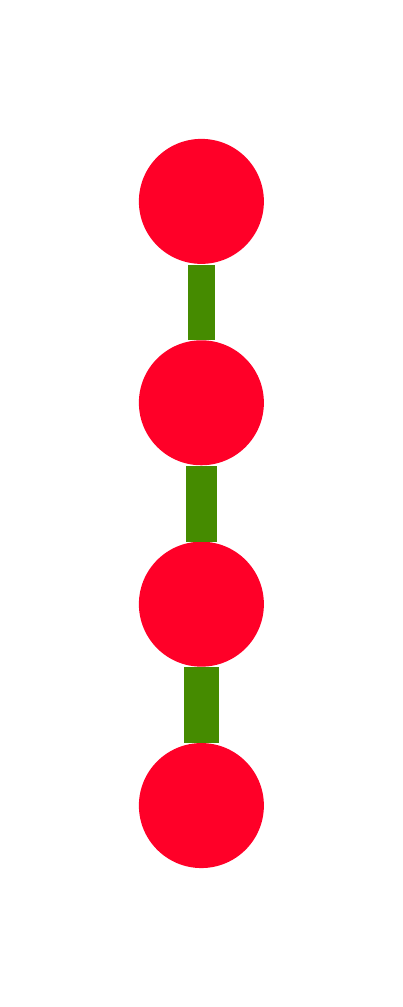}
    \includegraphics[scale=.14]{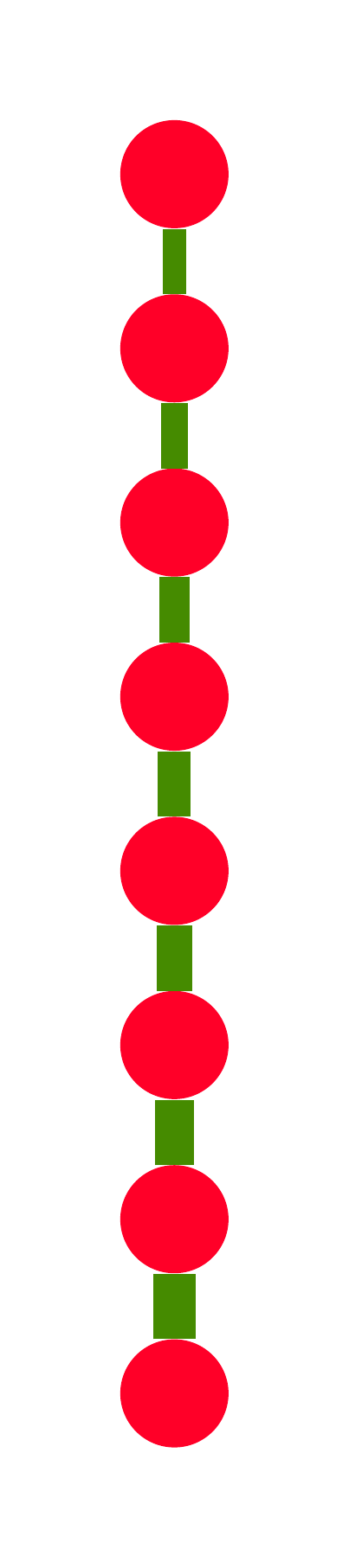}
    \label{fig:exp-simple-recursion-data}
    }
    \qquad
    \subfigure[Samples from the induced program.]{
    \includegraphics[scale=.14]{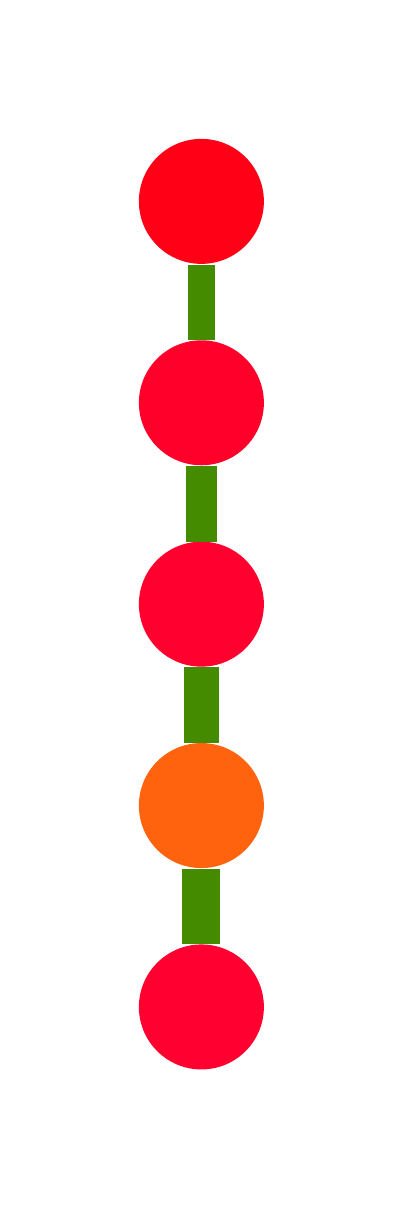}
    \includegraphics[scale=.14]{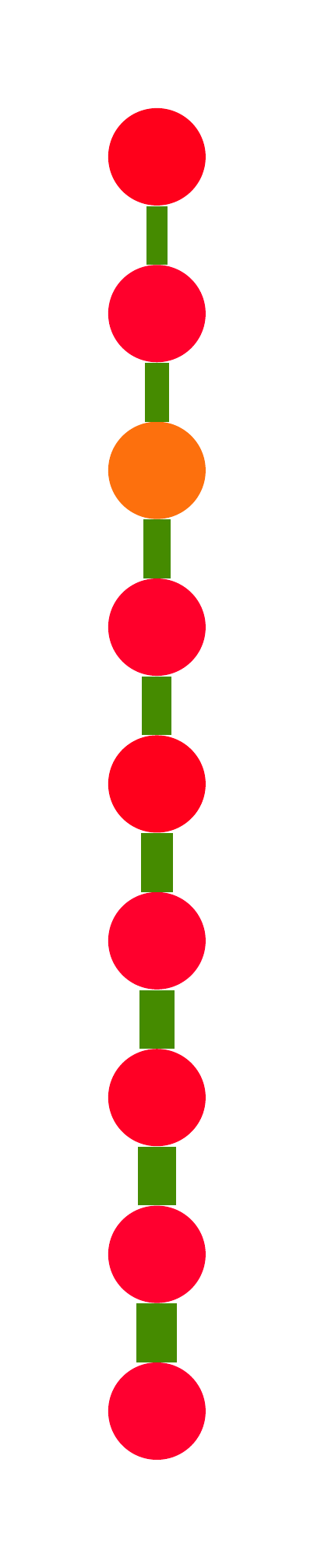}
    \includegraphics[scale=.14]{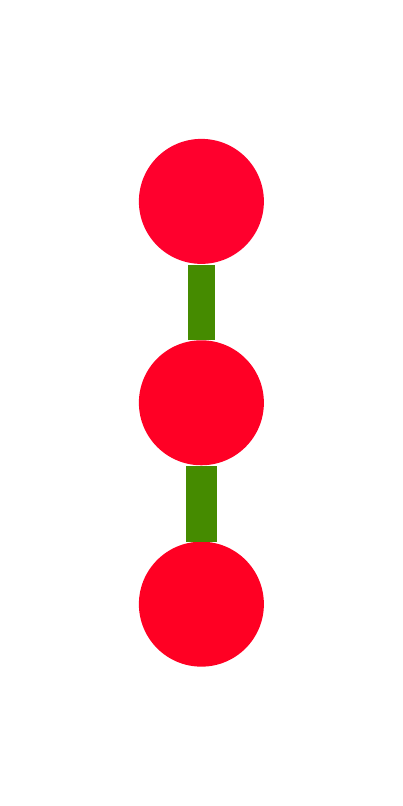}
    \includegraphics[scale=.14]{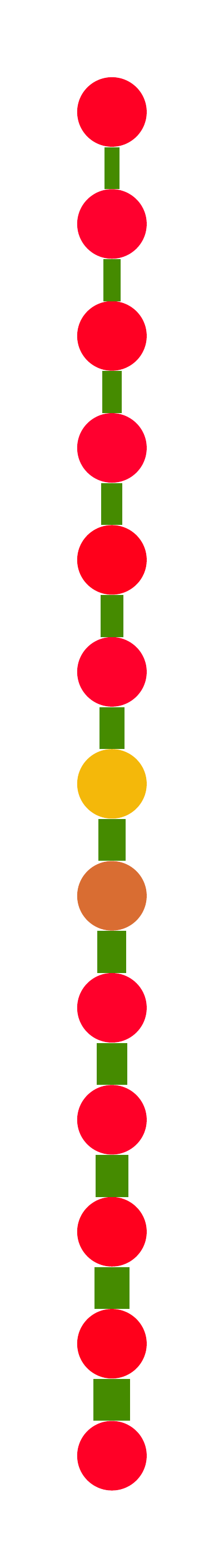}
    \includegraphics[scale=.14]{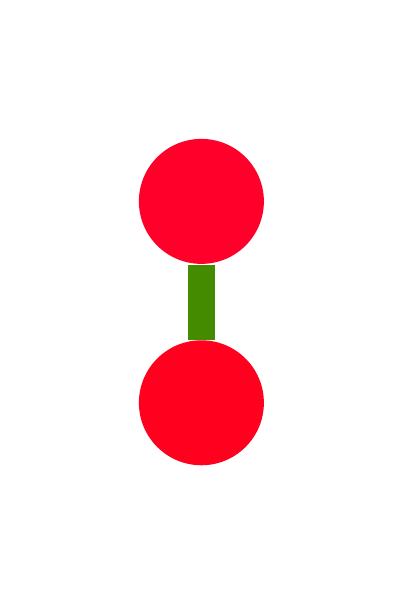}
    \includegraphics[scale=.14]{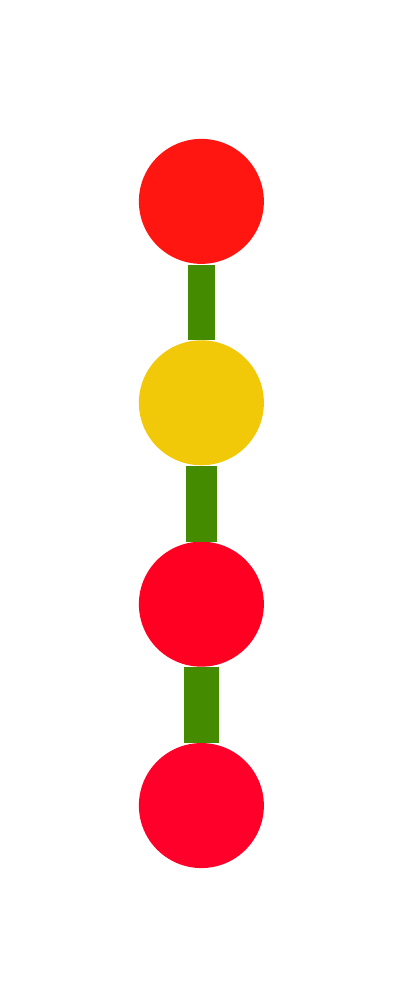}
    \includegraphics[scale=.14]{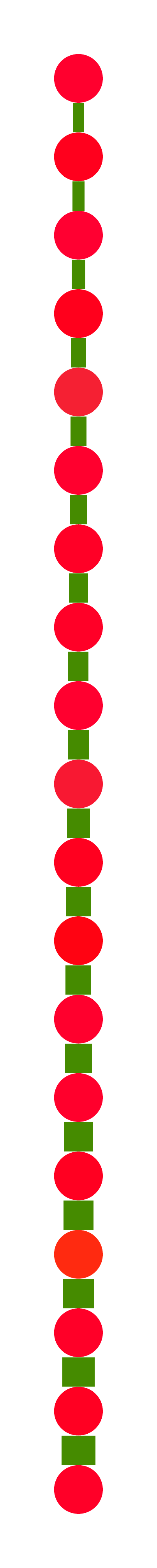}
    \includegraphics[scale=.14]{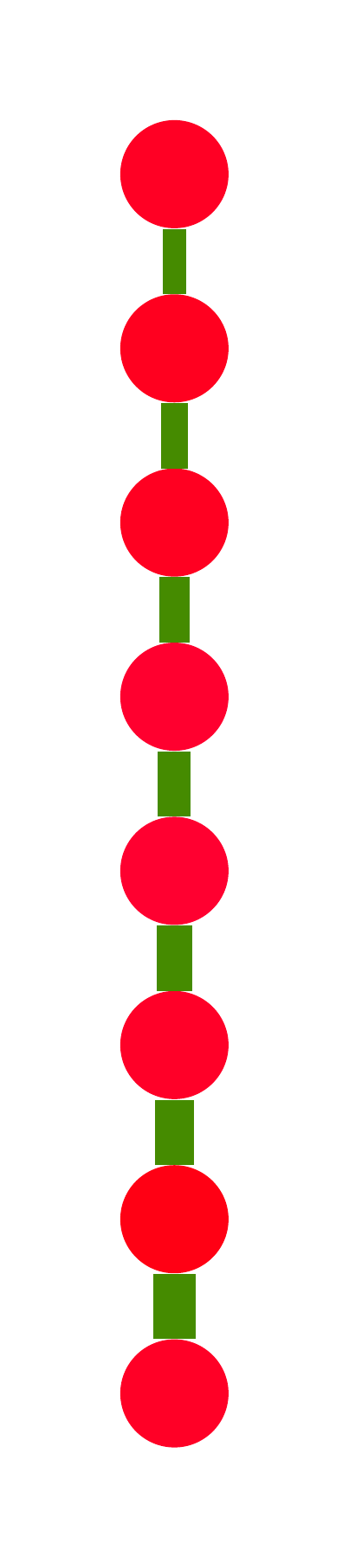}
    \label{fig:exp-simple-recursion-induced}    
    }
  \caption{Example 3: Simple recursion.}
  \label{fig:exp-simple-recursion}
\end{figure}

\subsection{Vine}
In this example, we use Bayesian program merging to induce a model of a ``vine with identical flowers''. This demonstrates multiple types of program transformations, all applying to the same data. System parameters: $\alpha=1$, beam width 1, depth 10.
\begin{lstlisting}[mathescape=true]
(define (vine)
  (if (flip .1)
      (node (data (color 100) (size .1)))
      (node (data (color 100) (size .1)) (vine) (flower))))

(define (flower)
  (node (data (color (gaussian 20 25)) (size .3))
        (petal 20)))

(define (petal shade)
  (node (data (color (gaussian shade 25)) (size .3))))
\end{lstlisting}
In the induced code below, \texttt{F1} roughly corresponds to \texttt{vine}. 
\begin{lstlisting}[mathescape=true]
(begin
  (define F3 ($\lbda$ () (F2 100 0.1)))
  (define F2
    ($\lbda$ (V4 V5)
      (data (color (gaussian V4 25)) (size V5))))
  (define F1
    ($\lbda$ (V2)
      (($\lbda$ (V1)
	 (($\lbda$ (V3)
	    (node (F3) V1
		  (node (F2 V2 0.3) (node (F2 V3 0.3)))))
	  12.538461538461538))
       (if (flip 12/13)
	   (F1 28.0)
	   (uniform-choice (node (F3)))))))
  ($\lbda$ () (uniform-choice (F1 42.0))))
\end{lstlisting}

\begin{figure}
  \subfigure[A sample from the original program.]{
    \includegraphics[scale=.14]{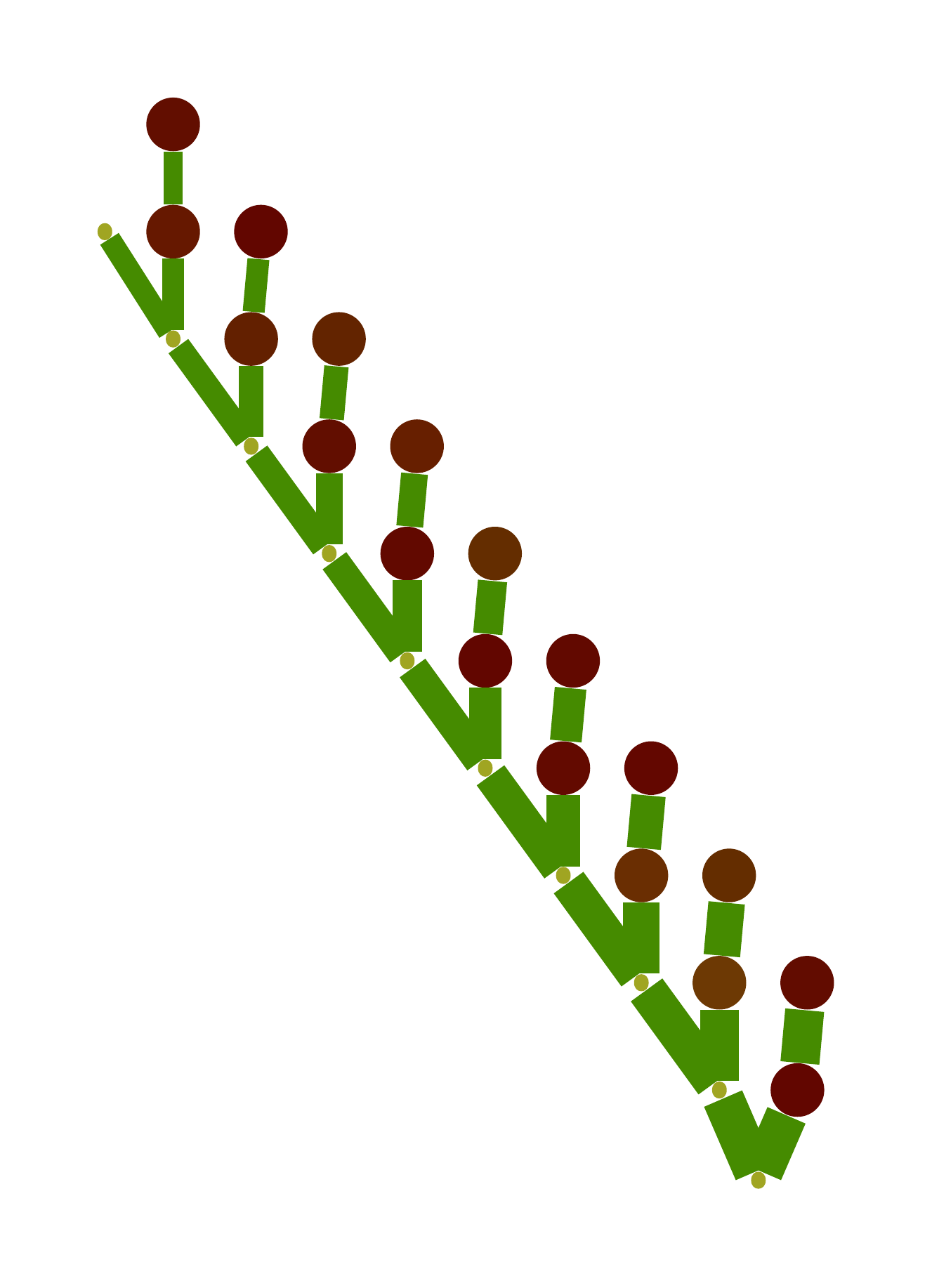}
    \label{fig:exp-vine-data}
    }
    \qquad \quad
    \subfigure[Samples from the induced program. A single observation is sufficient for the induction of a recursive program.]{
    \includegraphics[scale=.14]{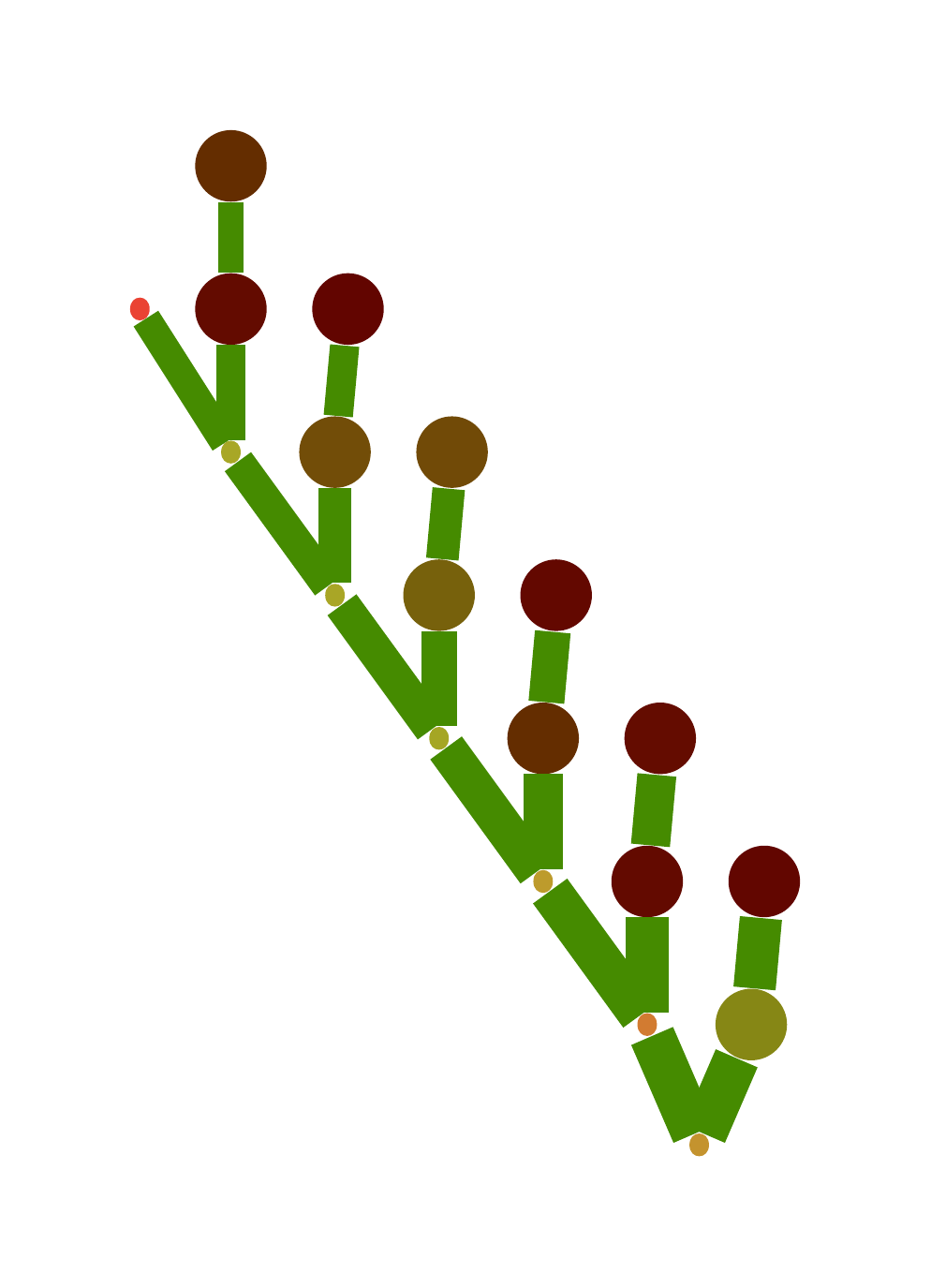}
    \includegraphics[scale=.14]{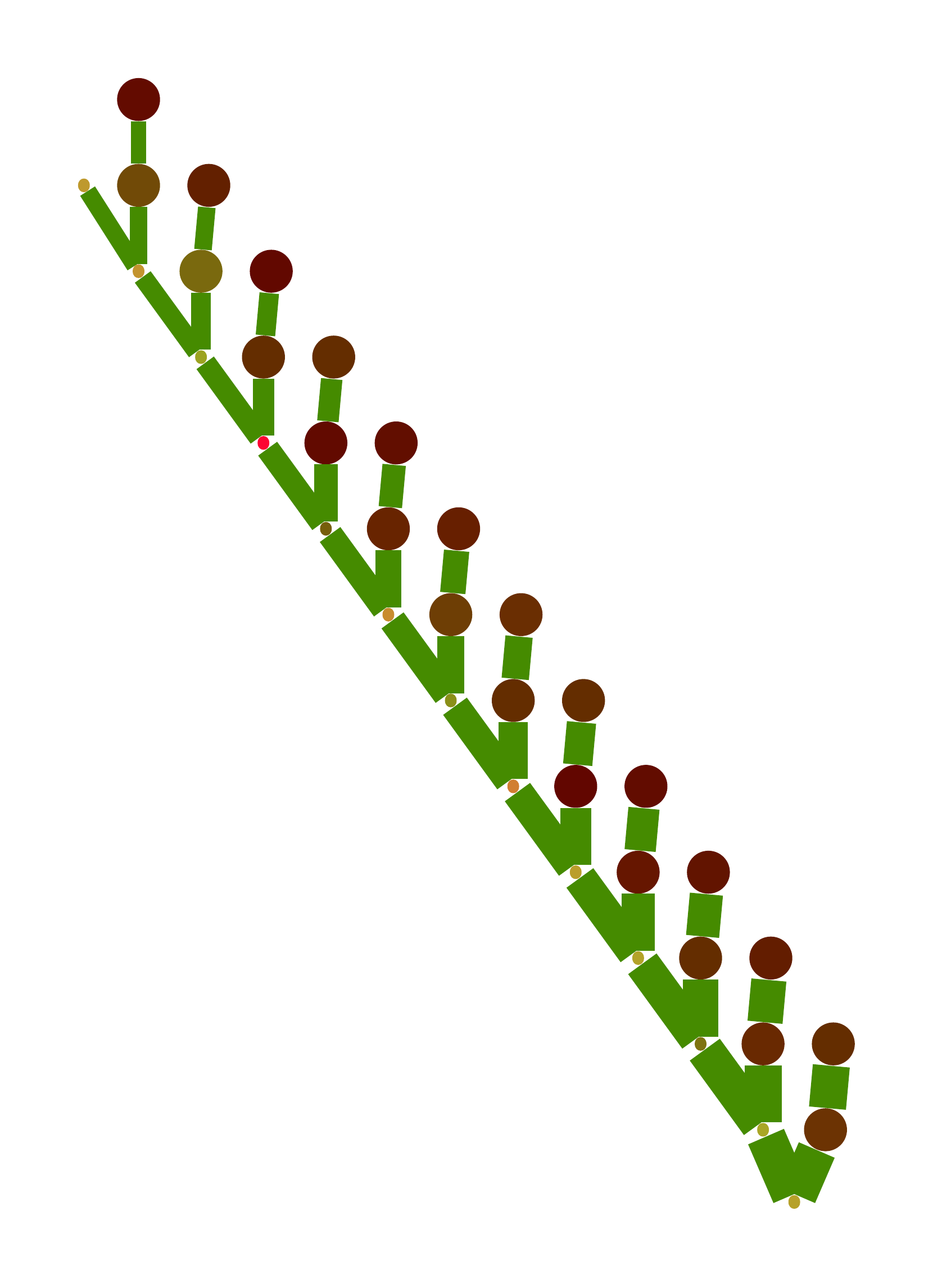}
    \includegraphics[scale=.14]{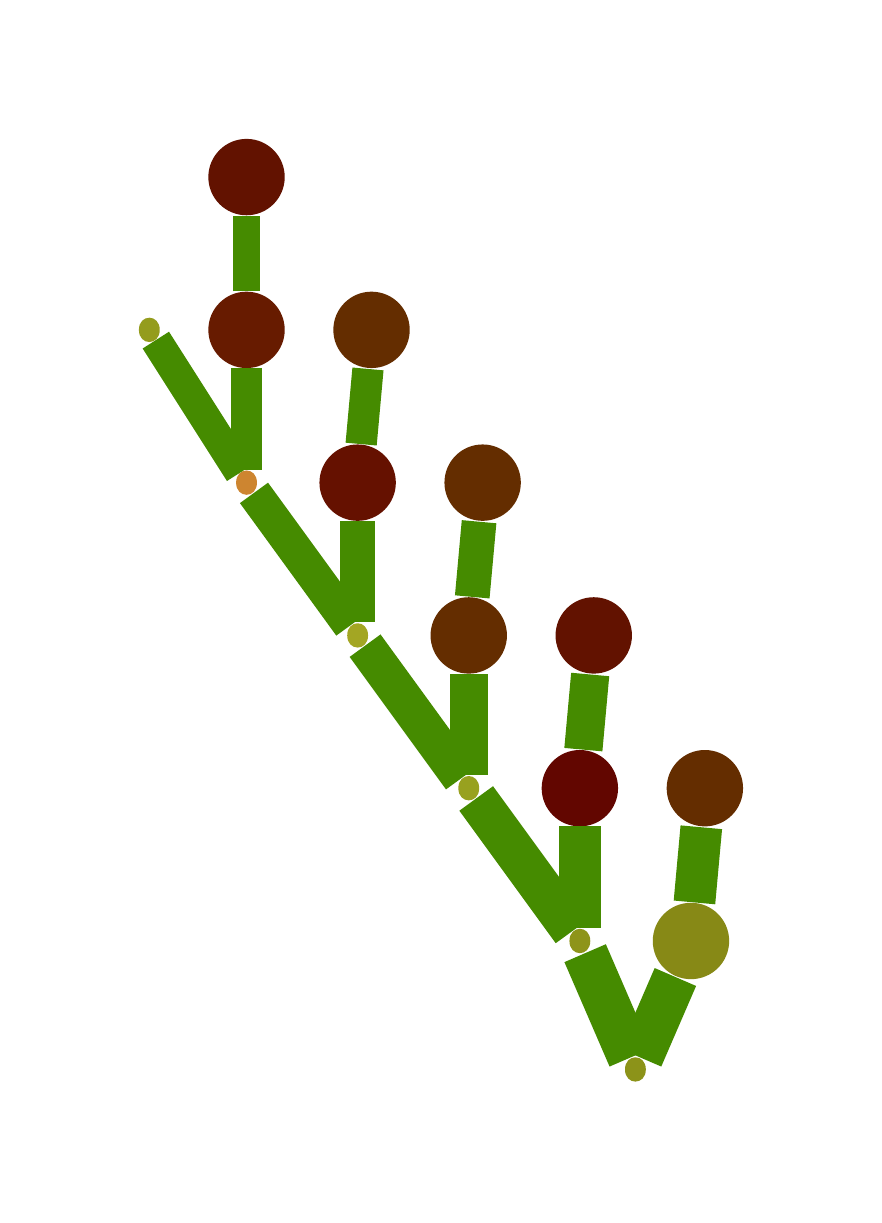}
    \includegraphics[scale=.14]{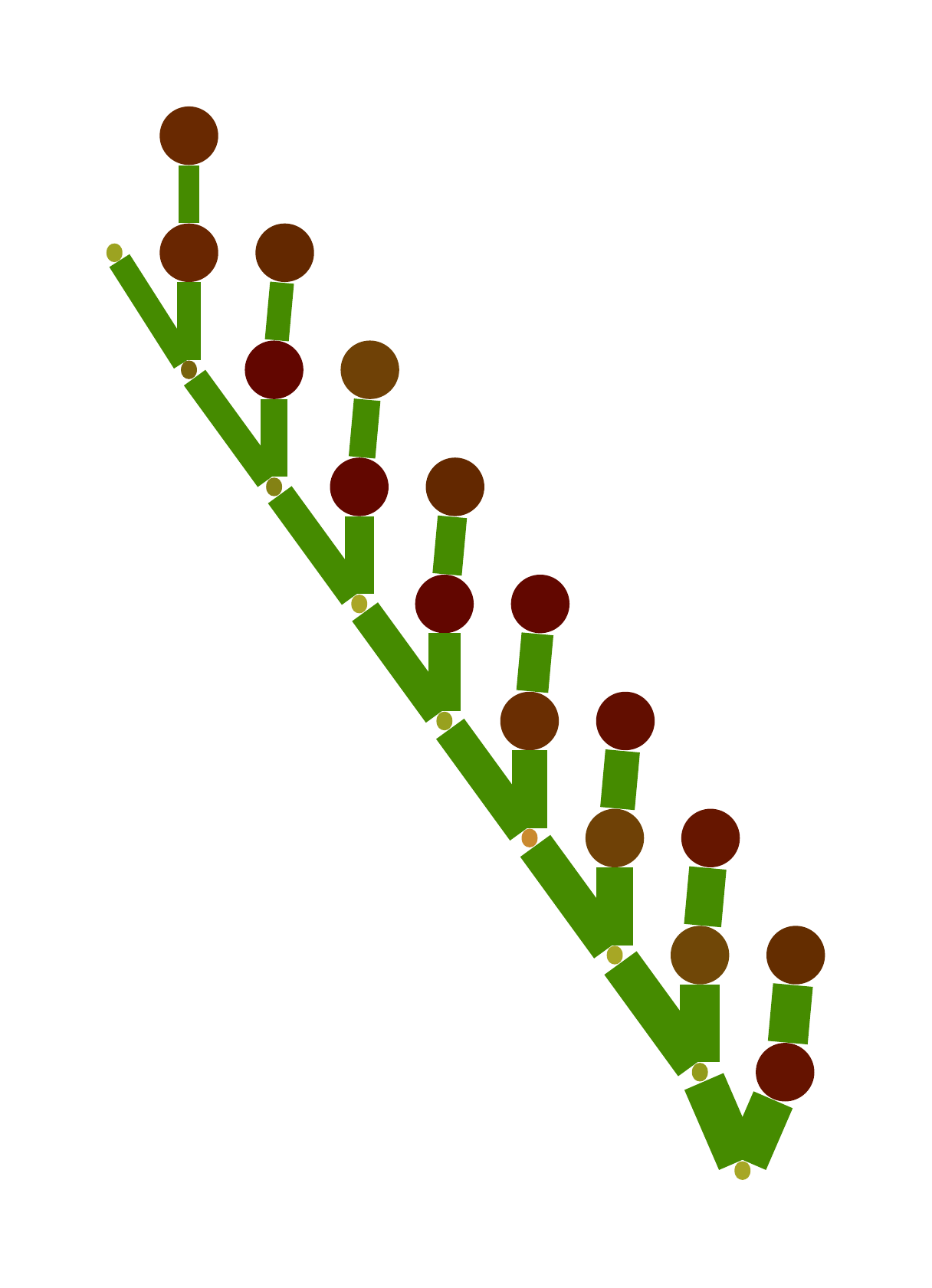}
    \includegraphics[scale=.14]{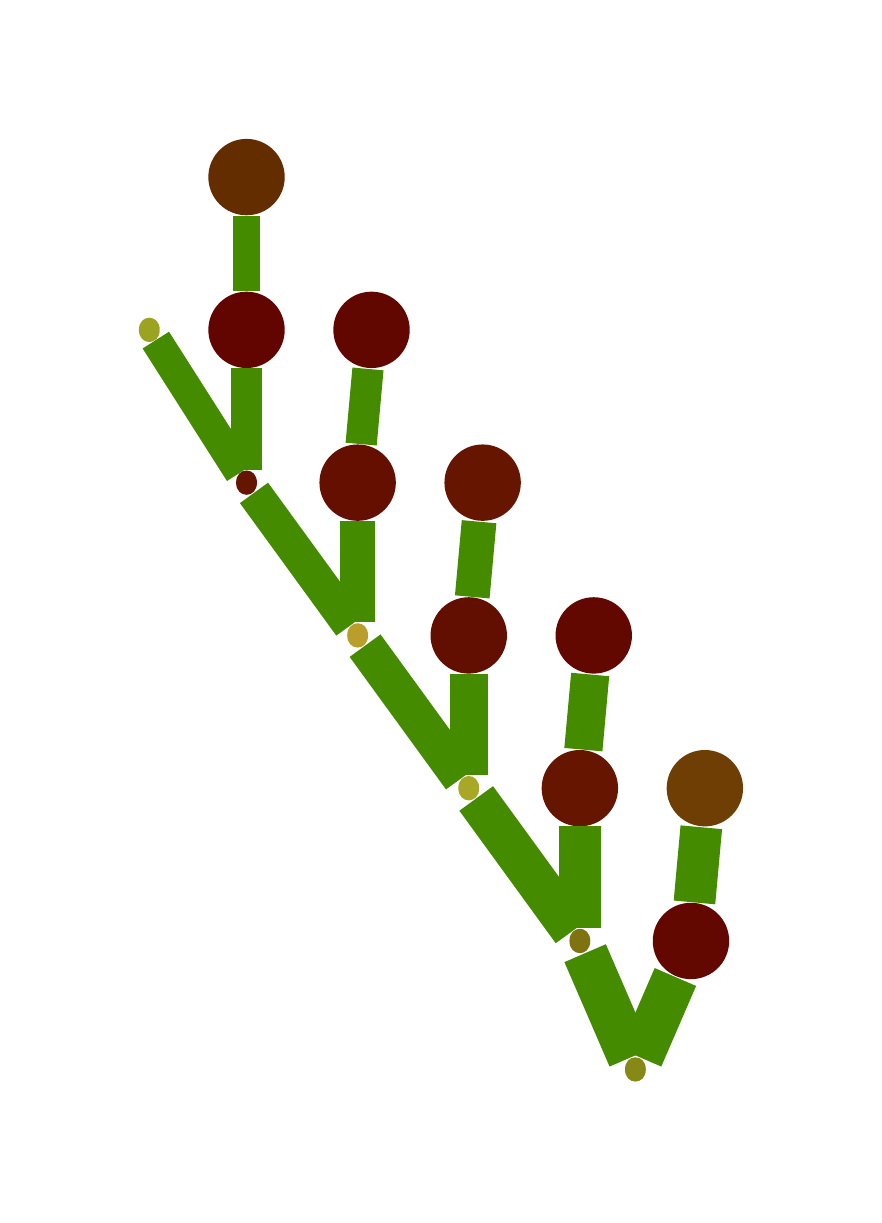}
    \includegraphics[scale=.14]{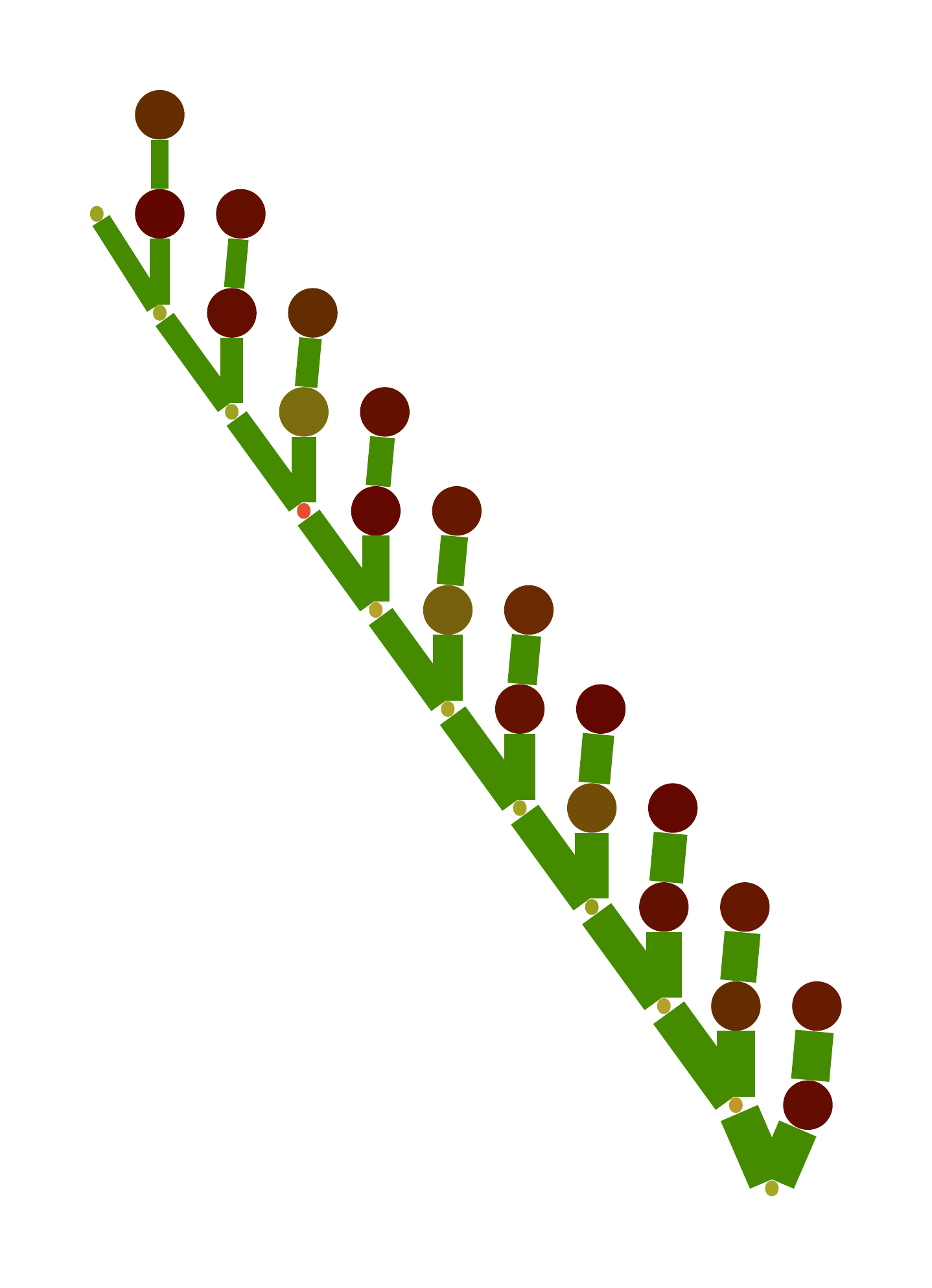}
    \label{fig:exp-vine-induced}    
    }
  \caption{Example 4: Vine.}
  \label{fig:exp-vine}
\end{figure}

\subsection{Tree}
This example demonstrates learning both a parameterized function and a recursion and applying these functions in multiple places within the same program. We use the functions shown below to generate flower-like patterns and a tree that consists of two branches, each of which ends in a flower of different color. We ran the system with $\alpha=3$, beam width $1$, and depth $10$.  Setting $\alpha=1$ results in a trade-off between prior and likelihood that does not make introducing recursion worthwhile. We expect that the system could learn a similar program with $\alpha=1$ if we provide training examples with more branch instances, since this would increase the amount of program compression.
\begin{lstlisting}[mathescape=true]
(define tree
  ($\lbda$ ()
    (uniform-choice
     (node (body) (branch) (branch)))))

(define (body) 
  (data (color (gaussian 50 25)) (size 1)))

(define (branch)
  (if (flip .1)
      (uniform-choice (flower 20) (flower 220))
      (node (branch-info) (branch))))

(define (branch-info)
  (data (color (gaussian 100 25)) (size .1)))

(define (flower shade)
  (node (data (color (gaussian shade 25)) (size .3))
        (petal shade)
        (petal shade)))

(define (petal shade)
  (node (data (color (gaussian shade 25)) (size .3))))
\end{lstlisting}
Bayesian model merging produces a program that has structure similar to the original generating program.  \texttt{F3} plays a similar role to the \texttt{flower} function by taking a single color as argument and creating three nodes of size $.3$ with the passed in color.  \texttt{F2} is a function that creates a branch that ends in a flower with either red or brown petals.
\begin{lstlisting}[mathescape=true]
(begin
  (define F4 ($\lbda$ (V9 V10) (node (F1 V9 0.1) V10)))
  (define F3
    ($\lbda$ (V6)
      (($\lbda$ (V7)
	 (($\lbda$ (V8)
	    (node (F1 V6 0.3) (node (F1 V7 0.3)) (node (F1 V8 0.3))))
	  V6))
       V6)))
  (define F2
    ($\lbda$ (V3 V4)
      (($\lbda$ (V5) (F4 V3 (F4 V4 V5)))
       (if (flip 6/7)
	   (F2 110.0 67.0)
	   (uniform-choice (F3 225.0) (F4 82.0 (F3 43.0)))))))
  (define F1
    ($\lbda$ (V1 V2)
      (data (color (gaussian V1 25)) (size V2))))
  ($\lbda$ ()
    (uniform-choice (F3 -4.0) (F3 192.0)
		    (node (F1 48.0 1) (F2 163.0 85.0) (F2 108.0 84.0))
		    (F3 -6.0))))
\end{lstlisting}

\begin{figure}
    \subfigure[Observations used for program induction.]{
    \includegraphics[scale=.16]{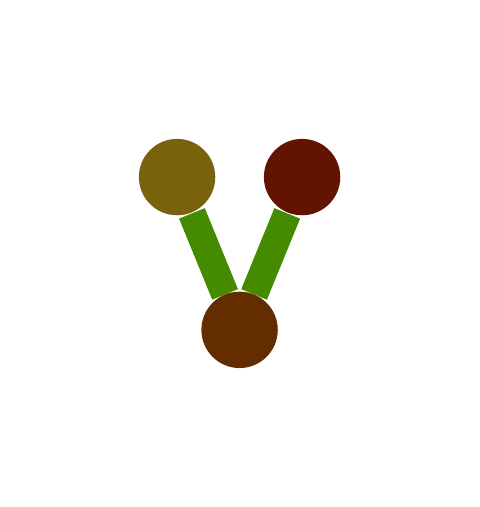}
    \includegraphics[scale=.16]{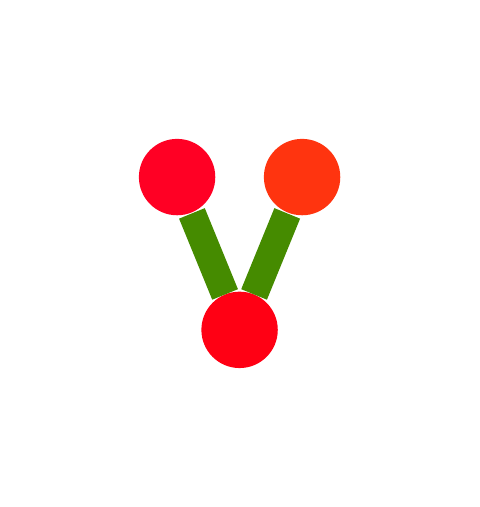}
    \includegraphics[scale=.16]{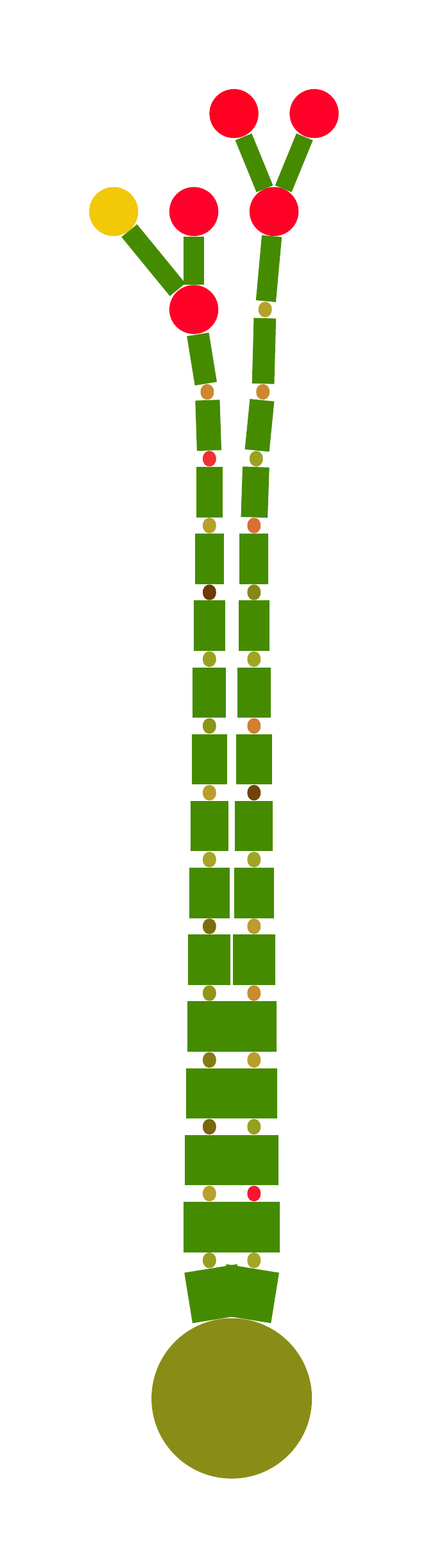}
    \includegraphics[scale=.16]{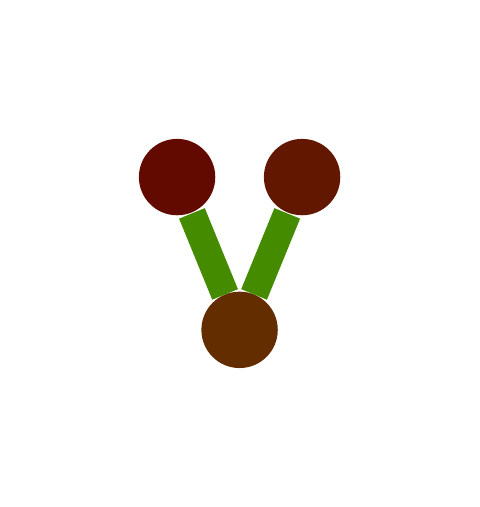}
    \label{fig:exp-tree-data}        
    }
    \quad \quad
    \subfigure[Samples from the induced program.]{
    \includegraphics[scale=.16]{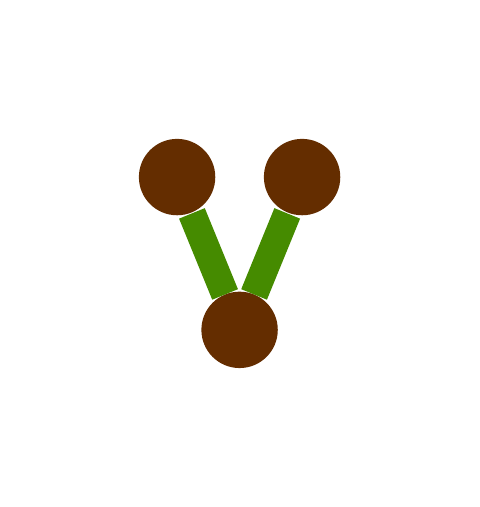}
    \includegraphics[scale=.16]{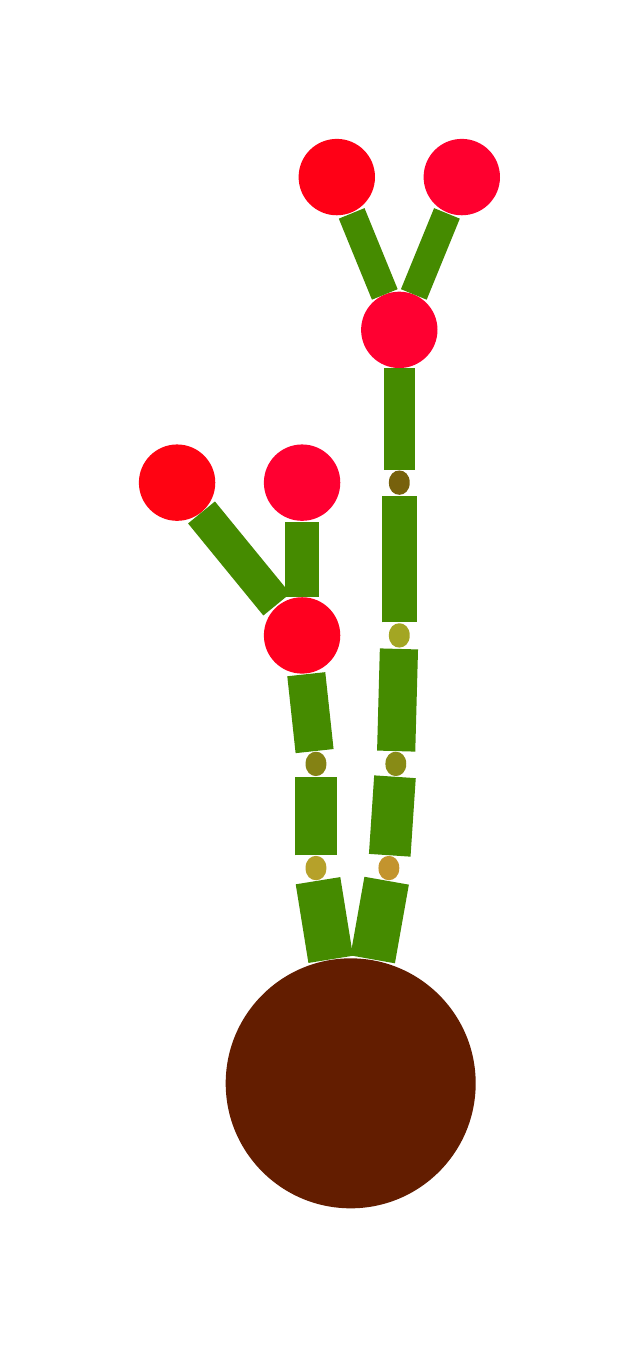}
    \includegraphics[scale=.16]{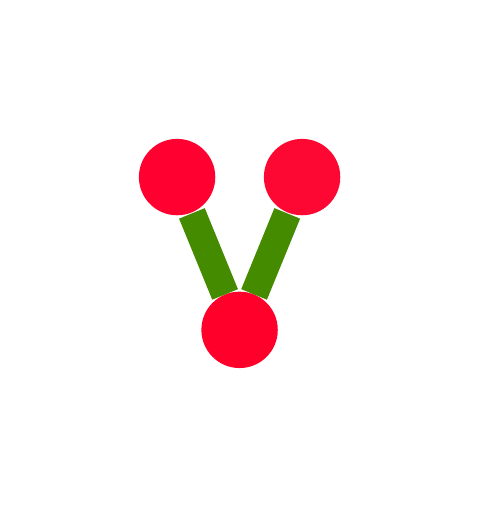}
    \includegraphics[scale=.16]{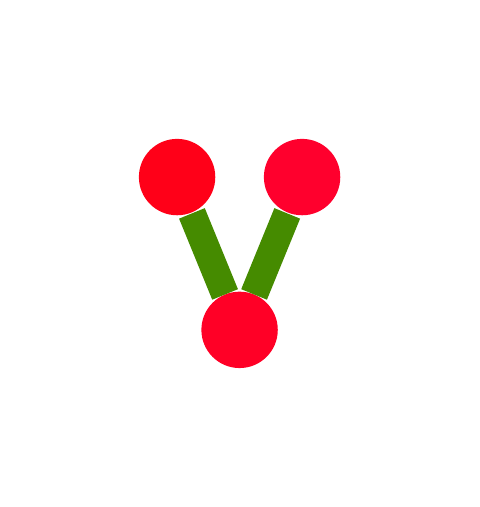}
    \includegraphics[scale=.16]{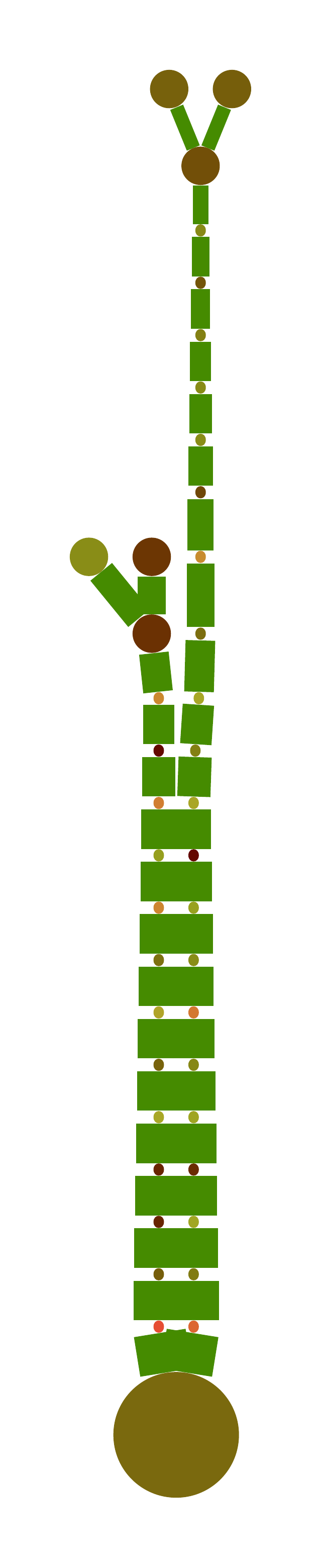}
    \includegraphics[scale=.16]{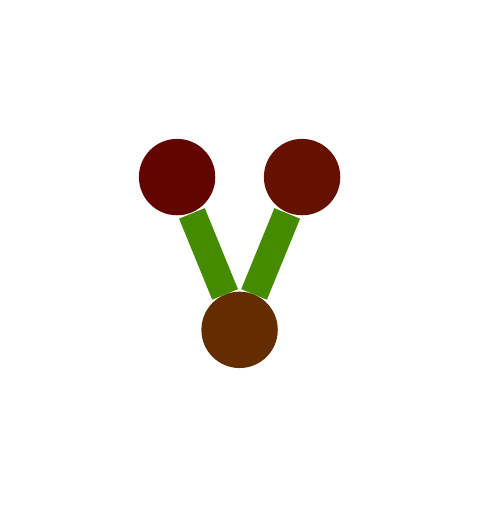}
    \includegraphics[scale=.16]{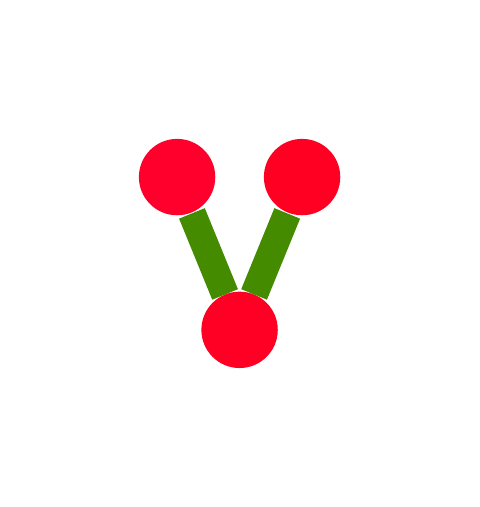}
    \includegraphics[scale=.16]{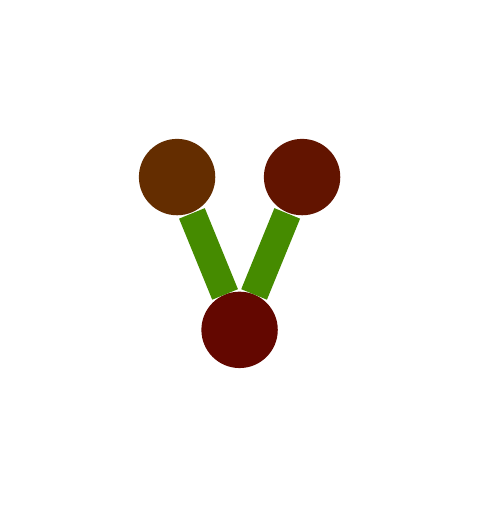}
    \includegraphics[scale=.16]{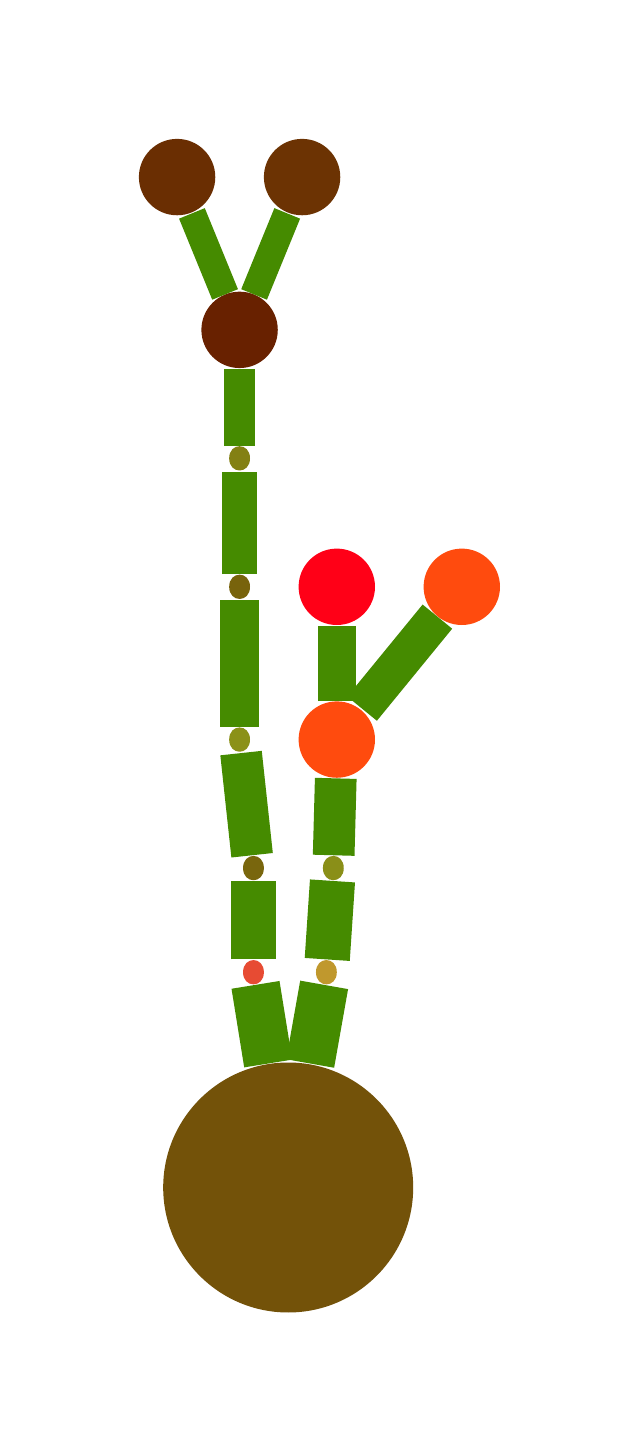}
    \includegraphics[scale=.16]{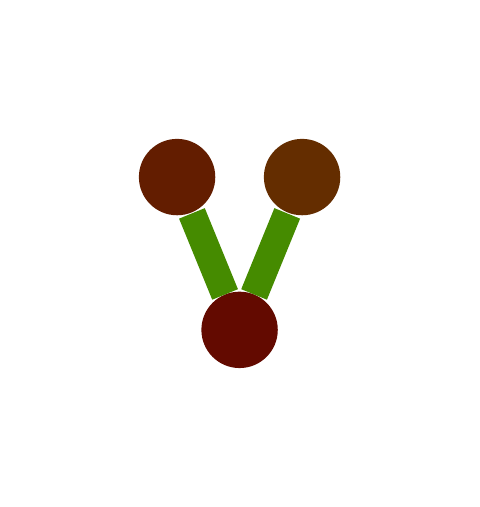}
    \includegraphics[scale=.16]{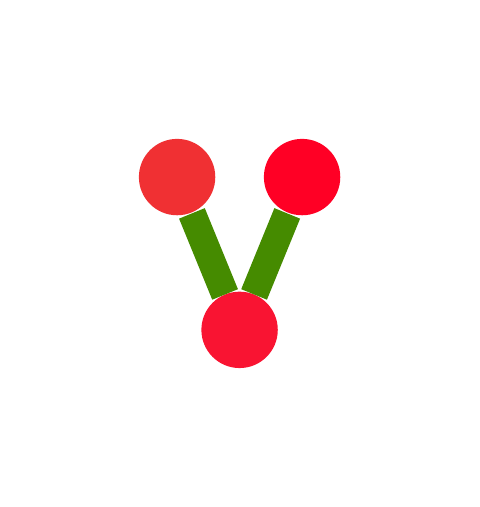}    
    \label{fig:exp-tree-induced}    
    }
  \caption{Example 5: Tree.}
  \label{fig:exp-tree}
\end{figure}

\newpage
\section{Conclusion}
We have presented Bayesian program merging, an approach to inducing generative models from data. The central idea of this approach is to first translate the data into a program without abstractions and to then compress this program by identifying repeated computations. We perform this compression using program transformations, with a goal of maximizing the posterior probability of the program given the data.

Many possible improvements of the system described in this paper present themselves. These include more sophisticated search strategies, more efficient ways of computing the likelihood, more sophisticated and robust methods of program transformation.
Furthermore, considerations such as those in section \ref{sec:dearg-as-induction} suggest that a more systematic development of the general framework is possible, uniting many of the individual search moves that we have proposed.

There are many barriers to overcome before probabilistic program induction can compete with state-of-the-art machine learning algorithms on real world problems. The potential for capturing rich patterns with only limited dependence on human engineering makes this a worthy pursuit.

\newpage
\bibliographystyle{plain}
\bibliography{bpm-report}
\end{document}